%% file: main_arxiv.tex
\pdfoutput=1

\documentclass[11pt]{article}

\usepackage{acl}

\usepackage{times}
\usepackage{latexsym}
\usepackage[T1]{fontenc}
\usepackage[utf8]{inputenc}
\usepackage{microtype}
\usepackage{inconsolata}

\usepackage{url}
\usepackage{graphicx}
\usepackage{booktabs}
\usepackage{amsfonts}
\usepackage{amsmath}
\usepackage{enumitem}
\usepackage{multirow}
\usepackage{makecell}
\usepackage{xcolor}
\usepackage[most]{tcolorbox}
\usepackage{longtable}
\usepackage{placeins}
\usepackage{float}

\newenvironment{promptbox}[1][]{%
  \par\vspace{4pt}%
  \noindent\colorbox{gray!12}{\begin{minipage}{\dimexpr\linewidth-2\fboxsep\relax}\small\textbf{#1}\end{minipage}}\par
  \nopagebreak\vspace{2pt}\nopagebreak
  \noindent\begin{minipage}{\dimexpr\linewidth-12pt\relax}\small\leftskip=4pt%
}{%
  \end{minipage}\par\vspace{4pt}%
}

\graphicspath{{figures/}}

\title{Cultural Awareness is Represented but Not Decoded:\\Tracing Mythological Knowledge across 18 Open-Source LLMs}

\author{
  Iaroslav Chelombitko\textsuperscript{\,1,2,3} \and
  Ekaterina Chelombitko\textsuperscript{\,1} \and
  Mika H\"am\"al\"ainen\textsuperscript{\,2}
  \\[2pt]
  \textsuperscript{1}DataSpike \quad
  \textsuperscript{2}Metropolia University of Applied Sciences, Helsinki, Finland \\
  \textsuperscript{3}Neapolis University Pafos, Paphos, Cyprus \\
  \texttt{i.chelombitko@nup.ac.cy}
}

\begin{document}
\maketitle

\begin{abstract}
Open-source LLMs reliably name \textit{Zeus}, \textit{Jupiter}, and \textit{Thor}, but recover their counterparts in less-represented traditions like Finnish, Slavic, Egyptian, or Chinese mythology far less consistently. We ask where inside the model this cultural default is produced. On a parallel cross-cultural substrate of Thompson-motif entities, we instrument 18 open-source LLMs from 8 architecture families with linear probing, logit lens, activation patching, and output extraction. The residual stream cleanly distinguishes cultures, well above a name-string baseline, yet the decoder collapses culturally-specific tokens onto dominant-tradition ones. \emph{The failure is at readout, not at representation.} Asking the same question in the target culture's native language versus English produces failures that cluster within language but decouple across language: the decoder is gated on prompt language. We release a per-entity (probe, output) decomposition framework, a citation-anchored cross-cultural ground truth, a within- versus cross-mode correlation test for language-conditioned readout, and per-entity predictions for all 18 models.\footnote{Code, data, and per-entity predictions for every model: \url{https://github.com/AragonerUA/folkmotif}; dataset card: \url{https://huggingface.co/datasets/Aragoner/folkmotif}.}
\end{abstract}


\section{Introduction}\label{sec:intro}

Large language models are deployed across the world, yet their pretraining corpora are dominated by English-language web data and a narrow band of culturally hegemonic sources. The resulting cultural skew shows up in tasks as different as moral reasoning, food and religious practice, named-entity recognition, and even temperature-scale defaults, with LLMs defaulting to the Anglo-American or broader Western frame across all of them \citep{naous2024beerprayer, palta2023fork, atari2023humans, adilazuarda2024culturesurvey}. Folk-narrative content sits squarely inside that bias surface, but it has a property the surveyed tasks lack: a long tradition of parallel structural cataloguing across cultures (the Aarne-Thompson-Uther type index and the Thompson Motif-Index of Folk-Literature \citep{thompson1955motifindex}) lets one ask the same question of the same model across many traditions with an unambiguous gold answer. We use this property to push the cultural-bias diagnosis from \emph{behavioral} (``what does the model say?'') to \emph{mechanistic} (``where inside the model is the bias produced?''), and ask what mitigations it suggests.

When asked to name the supreme sky god, open-source LLMs reliably produce \textit{Zeus} for ``Greek'', \textit{Jupiter} for ``Roman'', and \textit{Thor} for ``Norse'',\footnote{The chief sky god of Norse tradition is Odin; the Thompson-index canonical filler is Thor, the storm-sky figure.} but for less-represented traditions (Finnish, Ukrainian, Mesopotamian, etc.) the canonical filler is recovered far less consistently, a familiar behavioral diagnosis \citep{naous2024beerprayer, palta2023fork, ramezani2023cultural}; wrong answers tend to be refusals, hedges, or a culturally plausible but incorrect entity from within the target tradition. The follow-up is mechanistic: \emph{is the model unaware that the entities are culturally distinct, or is it aware but unable to retrieve the right name?} The two answers split cultural bias into two regimes that respond to opposite interventions, \textbf{representational flattening} (the residual stream does not separate cultures, so the fix needs new information) and \textbf{decoding flattening} (the residual separates cultures but the readout collapses culturally-specific tokens onto dominant-tradition ones). Disentangling these regimes under scaling and prompt-language conditioning motivates our five research questions:
\vspace{-0.4em}
\begin{itemize}\setlength{\itemsep}{0pt}\setlength{\parskip}{0pt}\setlength{\topsep}{0pt}
\item \textbf{RQ1} (representational preservation): \emph{To what extent} does the residual stream encode culture above a name-surface baseline when the output is wrong?
\item \textbf{RQ2} (layer-wise emergence): \emph{At what depth} does the gold-token continuation enter the top-$k$, relative to the probe peak?
\item \textbf{RQ3} (causal localization): \emph{Where} along the network does swapping residual streams between cross-cultural prompts begin to flip the model's continuation?
\item \textbf{RQ4} (scaling): \emph{How} does within-family scaling reshape the gap between probe and output accuracy: does it close, shrink, or persist?
\item \textbf{RQ5} (cross-lingual querying): \emph{How} do paraphrase failures decompose under prompt-language conditioning, and \emph{how much} cell recovery does a bilingual ensemble buy over either single mode?
\end{itemize}
\vspace{-0.4em}

To answer these we instrument the $18$ open-source LLMs listed in Table~\ref{tab:models}, spanning $1.2$B--$34$B across eight architecture families \citep{dubey2024llama3, gemma3technical, abdin2024phi4, yang2024qwen25, yang2025qwen3, young2024yi, olmo2025, ustun2024aya23, aryabumi2024aya101}, with four mechanistic measurements:
layer-wise linear probing of the residual stream for the $10$-way culture label (\textbf{E1}), a logit lens tracking the depth at which the gold entity's first sub-token enters the top-$k$ continuations (\textbf{E2}), cross-cultural activation patching that swaps source-culture residuals into the target-culture prompt (\textbf{E3}), and output extraction by greedy chat-template generation in two prompt modes, English (EN) and target-language (NL) query (\textbf{E4}); \S\ref{sec:instruments} specifies each. Each entity's E1 outcome (does the probe read the culture?) and EN outcome (does the model emit the right name?) form a $2\times2$ decomposition (\S\ref{sec:decomp}) with four cells: \textbf{Preserved} (recognized and emitted), \textbf{DecodingSuppressed} (recognized but emits a wrong name), \textbf{SurfaceLuck} (right name without internal disambiguation), and \textbf{RepresentationallyFlat} (neither). The substrate is a parallel cross-cultural entity set of $27$ Thompson-index motifs evaluated across $10$ cultures.

\textbf{Our main empirical finding is that DecodingSuppressed (probe right, output wrong) is the largest cell in every one of the $18$ models;} the asymmetry is at the decoder, not the encoder. Within-family scaling shrinks the probe$-$output gap but does not close it in any family (\S\ref{sec:scaling-v3}). The unembedding is also language-conditioned: cross-language paraphrases decouple while within-language ones correlate ($\overline{r}_{\text{within}} = 0.57$ vs $\overline{r}_{\text{cross}} = 0.29$), and a bilingual ensemble lifts cell recovery at zero training cost (\S\ref{sec:independence}).

\section{Related Work}\label{sec:related}

A growing behavioral literature documents that LLMs default to Western and Anglo-American cultural reference frames \citep{naous2024beerprayer, palta2023fork, ramezani2023cultural, atari2023humans}, with recent surveys consolidating the programme \citep{hershcovich2022challenges, adilazuarda2024culturesurvey}; cross-language and cross-script asymmetries underpinning these gaps are documented at the data and tokenization levels \citep{kreutzer2022quality, chelombitko2024qtok, chelombitko2024uralic, chelombitko2026subword242}, and the disconnect between formal ``low-resource'' coverage and actual cultural representation is articulated by \citet{hamalainen2021endangered}. What this strand cannot tell us is \emph{where} inside the model the failure happens.

The natural instruments are the linear probe \citep{alain2017probe, hewitt2019designing, belinkov2022probing}, which since LAMA \citep{petroni2019lama} has become the standard tool for asking what knowledge is linearly readable from a residual stream and has been applied to factual, geographic, and temporal information \citep{gurnee2023world, marks2024geometry}; the logit lens \citep{belrose2023tunedlens}, used in multilingual form to argue that LLMs reason in a dominant language and translate at the last layer \citep{wendler2024llamas}; and activation patching \citep{vig2020causal, meng2022rome, geiger2024finding, wang2023ioi}, the central causal instrument of the mechanistic-interpretability programme \citep{olah2020zoom, elhage2021mathframework}. For these instruments to mean anything cross-culturally we need a substrate parallel by construction, which we draw from Thompson's Motif-Index of Folk-Literature \citep{thompson1955motifindex}, previously applied in computational folkloristics \citep{karsdorp2015a}, in the spirit of treating LLM evaluation as a humanities-informed enterprise \citep{hamalainen2024humanities}. Our contribution is to localize the behavioral gap mechanistically: where the probe sees a strong signal but the lens and the output do not, the failure is the unembedding's, not the residual stream's.

Closest to us is CultureScope \citep{yu2025entangled}, which also brings mechanistic interpretability to cultural bias: it patches internal representations to extract cultural knowledge on everyday cultural-knowledge prompts and scores how far less-documented cultures are \emph{entangled} with dominant ones inside the representation, reporting that low-resource cultures are less susceptible because the model holds less parametric knowledge of them. Our question is complementary and our substrate is different: we ask whether the failure is representational or a readout failure, on a parallel citation-anchored mythology grid with one unambiguous gold entity per (motif, culture) cell, which lets us pair a per-cell probe outcome with a per-cell generation outcome. The two studies consequently localize cultural bias at different loci, entanglement within the representation vs suppression at decoding; the substrate (open-ended cultural knowledge vs a closed parallel entity set) and the readout target (a flattening score over representations vs the gold token at the output) are the plausible reasons the pictures differ, and both mechanisms can coexist in one model.

\section{Method}\label{sec:method}

\subsection{Parallel cultural entity set}\label{sec:data}

Our base unit is the (motif, culture) pair. A motif is a Thompson-index entry with a specifiable role, e.g.\ \textsc{A1141.2} ``the supreme sky god'' or \textsc{A220} ``the sun god''. We instantiate the role of 27 such motifs across ten cultures (\textbf{Greek}, \textbf{Roman}, \textbf{Norse}, \textbf{Finnish}, \textbf{Ukrainian}, \textbf{Indian}, \textbf{Egyptian}, \textbf{Chinese}, \textbf{Japanese}, and \textbf{Mesopotamian}) for a total of $270$ entities. The $10$ cultures span six language families (Indo-European, Uralic, Afro-Asiatic, Sino-Tibetan, Japonic, and Akkadian-Sumerian), so that within-family scaling, cross-lingual querying, and output decoding are not evaluated on a corpus dominated by Greco-Roman default-fitting.

For each (motif, culture) we record the canonical proper name in romanized English, the native-script form where applicable, the standardized motif description, and a per-row \texttt{source} field pointing to either a primary text (Eddas, Kalevala, Rig Veda, Pyramid Texts, Kojiki, Enuma Elish, the \emph{Primary Chronicle}, etc.) or a modern academic monograph.\footnote{Native-script forms are resolved via Wikidata. The complete dataset, including all native-script names and the NL prompt templates, is released as a Hugging Face dataset card (\url{https://huggingface.co/datasets/Aragoner/folkmotif}); see Appendix~\ref{app:datasetcard} for the release schema and Appendix~\ref{app:dataset-table} for the canonical name and citation anchor of every entity.} Four cells are structural absences (Japanese flood A1010, Finnish flood A1010, Mesopotamian theft-of-fire A1415, Egyptian forest-spirit F460), excluded from output-accuracy scoring; a further handful use a contested filler (e.g.\ Khors vs Dažbog as sun-disk vs sun-deity in Slavic, Mielikki vs Rauni for Finnish T110), where we follow modern scholarly consensus \citep{luczynski2020bogowie, gieysztor2006mitologia}. Per-tradition source repertoire and contested/absence rows are documented in Appendix~\ref{app:datasetcard}; per-row citation anchors in Appendix~\ref{app:dataset-table}.

For each entity we build two prompts:

\begin{itemize}[noitemsep]
\item \textbf{Contextual} (used for E1 hidden-state extraction): \emph{``\$\{name\} embodies the role of \$\{description\}.''} The entity comes first, the culture word is never present, and the entity-token activations cannot trivially copy a culture token via causal attention. The probe must derive culture from the name and role alone.
\item \textbf{Chat} (used for EN generation): \emph{``In \$\{culture\} mythology, what is the name of the \$\{description\}? Reply with only the proper name, no explanation, no extra words.''} Wrapped in the model's chat template where applicable.
\end{itemize}

\subsection{Models}

Table~\ref{tab:models} lists the 18 LLMs probed. Selection criteria: (i) open-source; (ii) decoder-only causal LM; (iii) at most $35$B parameters at fp16 (single $4 \times$ A100 $40$\,GB tensor-parallel slot). All experiments run at fp16 with no further quantization to avoid confounding cultural recall with quantization-induced degradation, which we treat as a separate study.

\input{tab_models}

\subsection{Mechanistic instruments}\label{sec:instruments}\label{sec:probe}\label{sec:e6}\label{sec:decomp}

We apply four complementary measurements to every (entity, culture) pair in every model.

\paragraph{Linear probing (E1).} For each model and each layer we average-pool residual-stream activations over the entity span of the contextual prompt and fit a 5-fold-stratified ridge classifier ($\alpha=1.0$) to predict the 10-way culture label \citep{alain2017probe, hewitt2019designing}. We report layer-wise accuracy, the layer at which accuracy peaks, and a label-shuffled chance ceiling.

\paragraph{Logit lens (E2).} Following the standard logit-lens recipe \citep{belrose2023tunedlens}, we apply the model's final norm and unembedding to every intermediate hidden state and ask at which depth the gold entity's first sub-token enters the top-$k$ continuations. We report top-$k$ rather than top-1 because gold-token boundaries are sensitive to leading-space tokenization and to morphological variants (\emph{Mokoš}/\emph{Mokosha} differ by one edit), a known instability of subword segmentation for morphologically rich languages \citep{chelombitko2025samponlp}, and our depth-of-readout claim is robust to whether one accepts the canonical or alternative filler in the contested rows of Appendix~\ref{app:dataset-table}. We use the raw lens for cross-model comparability; the tuned variant of \citet{belrose2023tunedlens} is discussed in the Limitations section.

\paragraph{Activation patching (E3).} For each motif we form pairs of cross-cultural prompts (Greek $\to$ Finnish, etc.); at each layer $\ell$ we replace the residual stream of the source prompt with that of the target prompt and measure the change in the gold-target log-probability \citep{vig2020causal, meng2022rome, geiger2024finding, wang2023ioi}.

\paragraph{Output extraction (E4).} We greedily decode up to $256$ tokens of the chat-template form. Reasoning-style chat templates that emit a \texttt{<think>...</think>} trace by default (Qwen 3.6 family) are run with the think trace disabled and any leftover block stripped before scoring. The cleaned generation is scored by exact match, substring match, and length-normalized Levenshtein similarity (threshold 0.8); a positive on any of the three counts as \emph{correct}. The instrument is applied to two prompt modes per cell: EN (English query, 5 paraphrases) and NL (target-culture native language, 5 paraphrases); see \S\ref{sec:h6e6} and Appendix~\ref{app:prompts}. Combining per-cell E1 and E4 outcomes gives the $2\times2$ decomposition introduced in \S\ref{sec:intro}; per-model cell shares are in Appendix~\ref{app:decomp-stacked}.

\section{Results}\label{sec:results}

\paragraph{How to read the four instruments.} Each instrument answers one question and, equally important, does not answer the others; the argument of this section is the \emph{conjunction}. The \textbf{linear probe} (E1) shows that the culture label is linearly decodable from the residual stream; it does not show that the model uses that information downstream, which is why we pair it per cell with generation. The \textbf{logit lens} (E2) shows at what depth the gold token enters the vocabulary projection; because intermediate states are not calibrated for direct decoding it gives a relative ordering, not a calibrated probability (see Limitations). \textbf{Activation patching} (E3) is the causal instrument: it shows \emph{where} intervening on the residual changes which culture's name is preferred, though not by what circuit. \textbf{Output extraction} (E4) shows what the user actually receives, and the \textbf{MCQ control} (\S\ref{sec:mcq}) separates that from the format of the task. Read together: culture is encoded (E1), encoded before it is decoded (E2), causally bound to the output late (E3), and still not emitted (E4) even when the task format is matched (\S\ref{sec:mcq}).

\subsection{RQ1: Decoding flattening is universal}\label{sec:decoding-universal}

Across all 18 LLMs in the sweep, the cultural identity of a mythological entity is recoverable from the residual stream at a rate the model never reproduces in its own generation (Table~\ref{tab:headline}). Seventeen of the $18$ probes clear the $0.60$ char-$n$-gram surface baseline (\S\ref{sec:eval-probe-control}), peaking at $0.61$--$0.88$; a one-sided paired bootstrap and an exact McNemar test confirm significance for all $17$ ($p<0.05$, App.~\ref{sec:probe-stability}), the sole exception being Gemma-4-E2B ($+0.011$, n.s.). The same models, asked to emit the name, land at $0.09$--$0.43$. \textbf{The smallest gap in the sweep is $0.26$} (Gemma-4-31B), the largest is $0.70$ (Llama-3.2-1B), and the gap shrinks but never closes within any family. The geometric form of this readout is the same across the sweep: at each model's probe-peak layer, the residual stream separates cultures into visibly distinct clusters under a linear projection (Figure~\ref{fig:latent-showcase} shows four representative panels; the full $18$-model grid is in Appendix~\ref{app:latent-all}).

Cell-wise, the loss has a single locus. \emph{DecodingSuppressed is the largest cell in every single model}, with shares $51$--$76\%$ (mean $0.65$): the probe reads the culture but the generation emits a wrong name. \emph{RepresentationallyFlat}, the cell predicted by a strong ``LLMs are translation machines'' reading, is consistently smaller ($0.10$--$0.33$, mean $0.18$): the cultural distinction is rarely missing from the residual stream, it is just rarely produced. Per-model trajectories, stacked decomposition shares, per-(model, culture) heatmaps, the direction-of-flattening confusion matrix, per-motif Preserved share, and full per-model results are in Appendices~\ref{app:probe}--\ref{app:fullresults}.

\paragraph{Is the measurement itself robust?}\label{sec:evaluation} Three checks say the gap is not an artefact of how we score. Re-scoring all $48{,}568$ generations under three stricter criteria (substring-only, first-word exact match, strict exact match) preserves the per-model ranking at Pearson $\geq 0.87$ (Table~\ref{tab:eval-scoring}), so no model's standing depends on our leniency. Per-(model, mode) $95\%$ bootstrap CIs have median half-width $0.046$, well inside every gap we report. And when a cell is wrong across all five paraphrases of a mode, the five wrong answers agree on their first word only $1.1\%$ of the time: the models are exploring, not converging on one confident substitute.

\input{tab_headline}

\begin{figure*}[!t]
    \centering
    \includegraphics[width=0.97\linewidth]{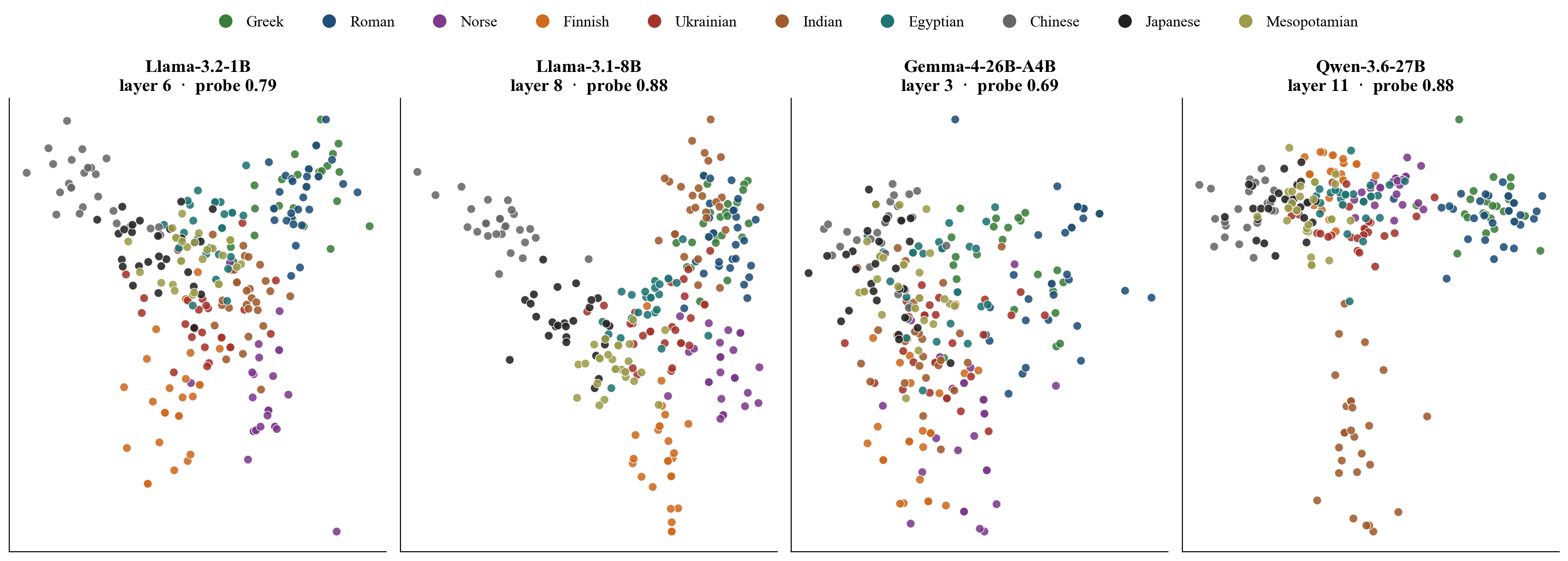}
    \caption{Latent-space projection at each model's probe-peak layer (50-dim PCA $\to$ LDA on culture labels) for four representative models from the $18$-model sweep; the $270$ (motif, culture) entities are colored by culture (legend on top). Cultures form visibly distinguishable clusters in the residual stream of every model, regardless of family or scale, even when the corresponding output accuracy collapses to a Greco-Roman default. The full $18$-model grid is in Appendix~\ref{app:latent-all} (Figure~\ref{fig:latent-all}).}
    \label{fig:latent-showcase}
\end{figure*}

\subsection{Output-format control: selection vs generation}\label{sec:mcq}

The probe$-$output gap could in part be a task-format artefact: the probe is a $10$-way classification while output extraction is open-ended generation. To match the output task to the probe, for each (motif, culture) cell we pose a multiple-choice question whose options are the parallel fillers of the same Thompson role across all $10$ cultures (per-cell chance $\approx 0.10$, matching the probe). Because every option instantiates the same role, topical matching cannot solve the task; only the culture$-$entity association the probe measures can. We score by restricted first-token log-probability (a single forward pass, no generation), which is also immune to the chat-template artefacts of \S\ref{sec:lens-patch} and \S\ref{sec:patching}. Table~\ref{tab:mcq} reports the result over $5$ paraphrases $\times$ $3$ option orders per cell.

Format does explain part of the raw gap: at identical chance, mean selection accuracy ($0.67$) sits far above free generation ($0.26$). But the chain \emph{representation ($0.79$) $\to$ selection ($0.67$) $\to$ generation ($0.26$)} localizes the residual loss at the generation step: with the output format matched to the probe, generation still loses $\sim$$41$ points, and scale does not close it (Phi-4 recovers in selection everything its probe reads, $0.86/0.86$, yet generates only $0.31$). The per-culture asymmetry persists under selection (Roman $0.60$, Finnish $0.58$ vs Japanese $0.78$; App.~\ref{sec:mcq-perculture}), and on Roman cells $49\%$ of errors land on the same-motif Greek counterpart (vs $\sim$$11\%$ under a uniform error model): the collapse onto the dominant tradition is visible inside a pure selection task. The dominant failure is therefore \emph{generation-time decoding suppression}, not a classification-vs-generation format effect.

\subsection{RQ2: Layer-wise emergence}\label{sec:lens-patch}

\begin{figure*}[!t]
    \centering
    \includegraphics[width=0.77\linewidth]{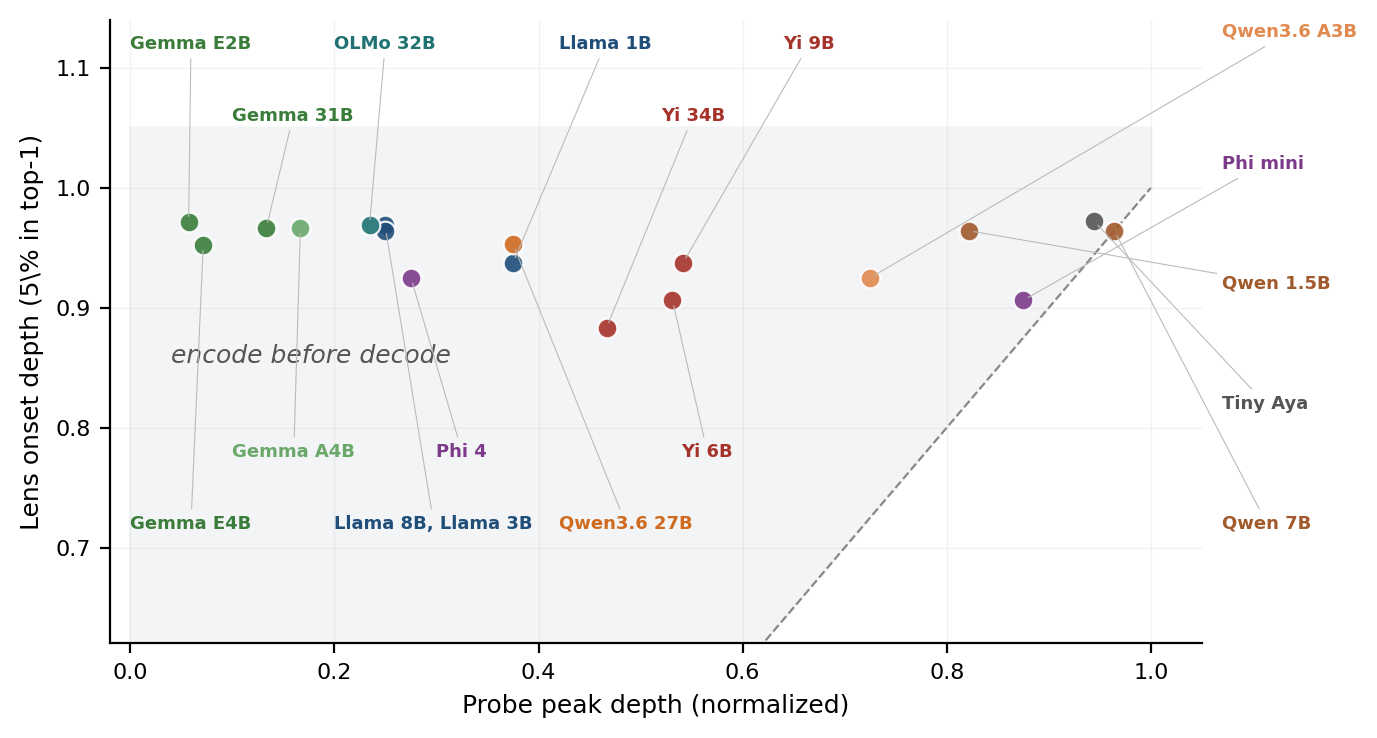}
    \caption{Encoded depth (linear-probe peak) vs decoded depth (logit-lens $5\%$-in-top-$1$ onset) per model. Every model is on or above the diagonal: culture is read out of the residual stream before the model commits to emitting the right token. Encoding depth spreads across families (Gemma early, Yi mid, Phi-mini / Tiny-Aya / Qwen-7B late); decoding depth is uniformly in the last $12\%$ of layers. Lens-onset bar chart and per-culture emergence curves in Appendix~\ref{app:lens}.}
    \label{fig:probe-vs-lens}
\end{figure*}

Decoding depth is uniformly late, encoding depth varies by family. Across all $18$ models, the logit lens (E2) crosses its $5\%$-in-top-$1$ onset in the last $12\%$ of layers on every single model (onset range $0.88$--$0.97$, median $0.96$; for Qwen-3.6-27B the bare-prompt lens is used because its chat template emits a generic ``Here'' token at every layer, App.~\ref{app:lens}). The linear probe peak depth is more spread across families (Figure~\ref{fig:probe-vs-lens}). Every model sits on or above the diagonal of the encoded-vs-decoded scatter: the residual stream becomes culturally legible before the network commits to the output token, and the late-decoding band is robust across $1.2$B$\to 32$B, dense vs MoE, and chat-tuned vs base architectures.

The per-culture lens schedule is not uniform: Greek and Roman cross the $5\%$-in-top-$5$ threshold first, while the other eight cultures cross later by $\sim 1.3$ layers on a $32$-layer model on average (App.~\ref{app:lens}).

\subsection{RQ3: Causal localization}\label{sec:patching}

The causal locus of the cultural readout coincides with the lens decoding band, not the probe encoding band. Activation patching (E3, \S\ref{sec:instruments}) measures, for each cross-cultural prompt pair and each layer, whether swapping the source-culture residual into the target-culture prompt causes the source-culture gold name to receive a higher next-token logit than the target-culture one. Across all $18$ models, this preference-flip rate sits at the baseline level through the early third of the network, rises through the middle, and peaks in the last quarter (peak depth $0.75$--$1.00$, median $0.89$; peak rate $0.40$--$0.95$, median $0.75$; median lift over baseline $+0.55$). The same depth band shows the lens onset (\S\ref{sec:lens-patch}); the culture direction is therefore not just decoded late but also \emph{causally bound to the output} late, at roughly $7\times$ the early-network rate. Because patching intervenes rather than observes, this makes the encode-vs-decode separation mechanistic rather than merely correlational.

\textbf{Per-family pattern.} Peak depth is concentrated near the network end across families (Llama 3.x $0.86$--$1.00$, Gemma 4 $0.88$--$1.00$, Phi 4 $0.78$--$0.86$, Qwen 1.x $0.85$--$1.00$, Qwen 3.6 $0.89$--$0.95$, Yi 1.5 $0.86$--$1.00$, OLMo $0.87$). \textbf{Qwen-3.6-27B} is reported on a bare-prompt estimate. Chat-template patching scores it as flat $\approx 0\%$ because the swap position decodes to the generic token \texttt{Here} in $320/320$ rows (same artefact as its lens, App.~\ref{app:lens}); switching to bare-prompt the lens recovers and the layerwise culture-preference rate peaks at $0.95$ at depth $0.95$ (App.~\ref{app:patching}), matching the other $17$ models. Explicit bare-prompt patching for this model is left to a follow-up.

\subsection{RQ4: Within-family scaling shrinks but does not close the gap}\label{sec:scaling-v3}

Scaling helps but does not close the gap. Table~\ref{tab:scaling-summary} reports smallest-to-largest probe peak, output, and gap for the five families with more than one completed size: in every family both probe and output rise with parameters and the probe$-$output gap shrinks; none close it. Llama 3.x gives the cleanest signal across three sizes ($1.2$B$\to 8$B): probe peak $0.79 \to 0.88$, output $0.09 \to 0.25$, gap $0.70 \to 0.63$ monotonically. \emph{Even at $8$B the gap is still $0.63$}, and a naive log-parameter extrapolation predicts only $\approx +0.10$ output gain per decade of parameters, insufficient to close the gap within an order of magnitude. Per-family scaling curves with both NL and EN modes in Appendix~\ref{app:scaling}.

\begin{table}[h]
\centering\small
\begin{tabular}{lccc}
\toprule
\textbf{Family} & \textbf{Probe} & \textbf{Output} & \textbf{Gap} \\
\midrule
\makecell[tl]{Llama 3.x\\($1$B$\to 8$B)}      & $.79 \to .88$ & $.09 \to .25$ & $.70 \to .63$ \\
\makecell[tl]{Phi 4\\($3.8$B$\to 14$B)}        & $.83 \to .86$ & $.20 \to .31$ & $.63 \to .55$ \\
\makecell[tl]{Gemma 4\\(E2B$\to 31$B)}         & $.61 \to .69$ & $.27 \to .36$ & $.34 \to .33$ \\
\makecell[tl]{Yi 1.5\\($6$B$\to 34$B)}         & $.77 \to .83$ & $.13 \to .26$ & $.64 \to .57$ \\
\makecell[tl]{Qwen 1.x\\($1.5$B$\to 7$B)}      & $.75 \to .80$ & $.12 \to .24$ & $.63 \to .57$ \\
\bottomrule
\end{tabular}
\caption{Within-family scaling endpoints: probe peak, best-of-(NL, EN) majority-correct output, and gap, smallest$\to$largest size in each family. The gap shrinks with scale in every family; no family closes it.}
\label{tab:scaling-summary}
\end{table}

\subsection{RQ5: Cross-lingual querying conditions the readout}\label{sec:h6e6}\label{sec:percult-h6}\label{sec:independence}\label{sec:bilingual-honest}

The cultural readout from the residual stream is language-conditioned: under English query and under in-culture query the model gates partially disjoint subsets of the same representation. For each (motif, culture) we pose five paraphrases in English (EN) and five in the target culture's native language (NL): Modern Greek, Italian, Norwegian Bokm\aa{}l, Finnish, Ukrainian, Hindi, Modern Standard Arabic, Mandarin, Japanese, and, for Mesopotamian, English with an explicit ``(Sumerian-Akkadian)'' parenthetical (the only fallback case; both substrate languages extinct without continuous spoken descendant). In every case the NL language is the \emph{living} language of present-day reception of the canon, not the historical language of original composition (Modern Greek over Ancient Greek, Italian over Latin, Hindi over Sanskrit, etc.); the per-culture rationale is in Appendix~\ref{app:prompts}. The prompt protocol is in the same appendix.

\begin{figure*}[!t]
    \centering
    \includegraphics[width=0.81\linewidth]{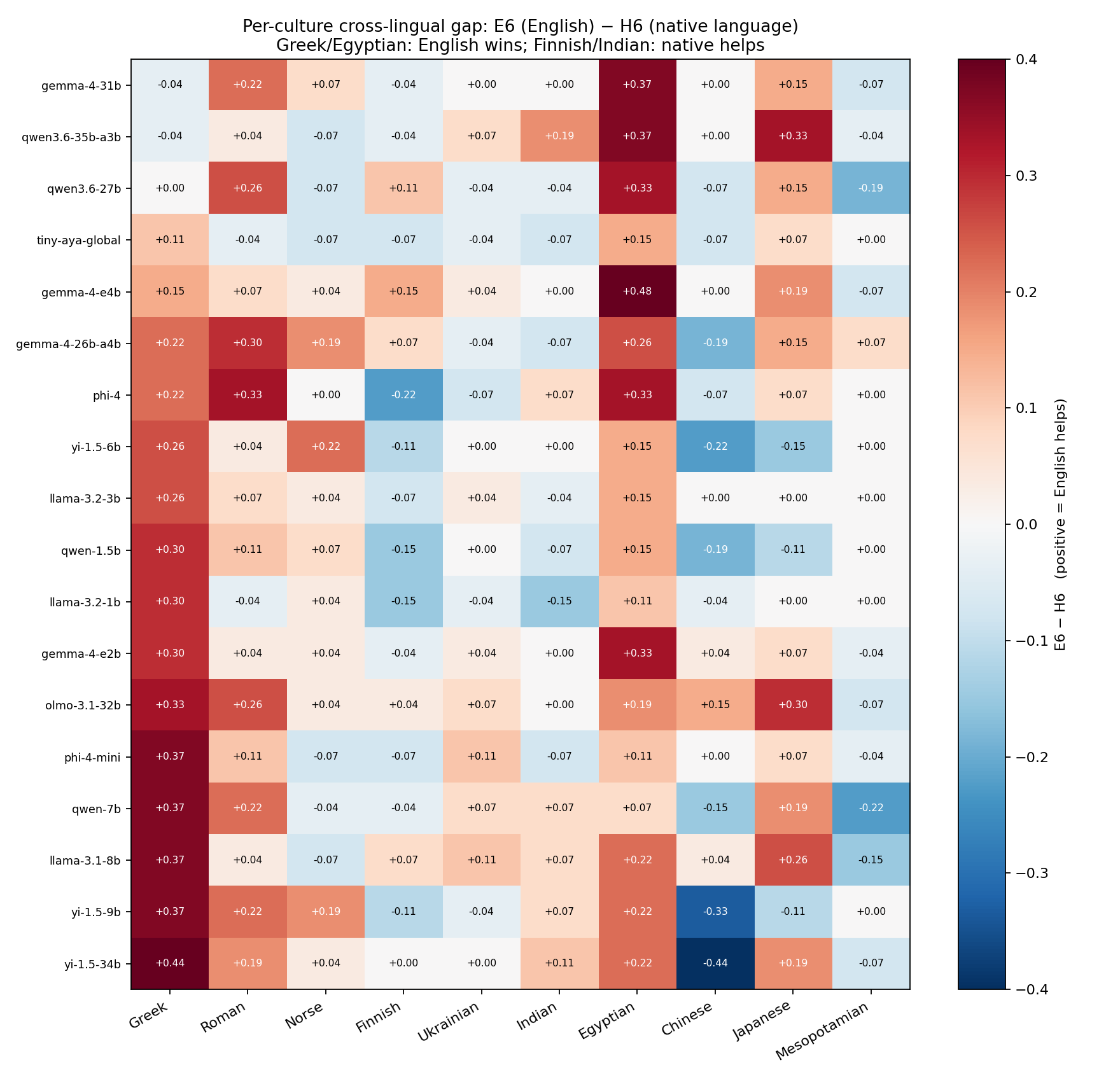}
    \caption{Per-(model, culture) EN$-$NL majority-accuracy delta. Red: English query wins; blue: native query wins; white: tie. The pattern tracks the \emph{language in which each mythological canon is documented}, not the script of the NL language nor the speaker base of the target language. Per-model aggregate NL vs EN and the NL leaderboard in Appendix~\ref{app:h6e6-extra}.}
    \label{fig:percult-delta}
\end{figure*}

\paragraph{Per-culture asymmetry.} Aggregated across the chat-template models with both modes, EN wins on average by $+0.04$ majority-accuracy (English mean $0.23$, native mean $0.19$); three exceptions (Gemma-4-31B, Qwen-3.6-27B, Tiny-Aya-Global) prefer NL. That average, however, is the least interesting number here, because the per-culture deltas run in both directions and track \emph{the language in which each canon is documented} (Table~\ref{tab:percult-delta}): Greek ($+0.24$) and Egyptian ($+0.18$) are strongly English-favouring, both traditions being read today mostly through English-language classical and Egyptological scholarship, while Chinese ($-0.07$) and Finnish ($-0.06$) favour the native query, their canons being curated in-language. The six cultures in between sit within $\pm 0.09$ of zero, close to the Mesopotamian control ($-0.04$), which poses both modes in English and therefore measures pure paraphrase noise. Figure~\ref{fig:percult-delta} shows the same pattern per model.

That the direction flips by culture already argues against a simple ``models cannot read the target language'' account, and three further checks close it off. Tokenizer fertility, the standard proxy for how well a tokenizer serves a language \citep{chelombitko2024qtok}, does not predict the per-cell delta ($r = -0.004$, $n = 120$, on Belebele passages \citep{bandarkar2024belebele, costajussa2022flores}). Tiny-Aya-Global \citep{ustun2024aya23, aryabumi2024aya101}, which covers all eight of our non-CJK non-fallback NL languages by design, shows the \emph{smallest} gap in the sweep ($0.004$) where a capability account predicts the largest. And recomputing the contrast on only the $14$ models with documented multilingual pretraining leaves the per-culture ordering unchanged (Spearman $\rho = 0.96$; App.~\ref{sec:tier-a}). A missing-capability confound can depress NL accuracy but cannot manufacture the NL advantages we see on Finnish, Norse and Chinese.

\begin{table}[h]
\centering\small
\setlength{\tabcolsep}{4pt}
\begin{tabular}{lrl}
\toprule
\textbf{Culture} & \textbf{$\overline{\Delta}_{\text{EN-NL}}$} & \textbf{Reading} \\
\midrule
Greek         & $+0.241$ & EN canon \\
Egyptian      & $+0.177$ & EN canon \\
Roman         & $+0.085$ & mild EN \\
Japanese      & $+0.040$ & mild EN \\
Norse         & $+0.026$ & mild EN \\
Ukrainian     & $+0.016$ & mild EN \\
Indian        & $-0.008$ & tie \\
Mesopotamian  & $-0.037$ & within-EN control \\
Finnish       & $-0.058$ & native canon \\
Chinese       & $-0.071$ & native canon \\
\bottomrule
\end{tabular}
\caption{Per-culture EN$-$NL majority-accuracy delta, averaged across chat-template models with both modes. Positive: English query wins; negative: native query wins.}
\label{tab:percult-delta}
\end{table}

\paragraph{Within-language failures correlate, cross-language failures decouple.} A sharper question sits below the aggregate deltas: at the (motif, culture) cell, comparing within-mode correctness correlation (5 paraphrases of one language) against cross-mode correlation (5 NL $\times$ 5 EN) isolates language-conditioning from paraphrase-format sensitivity. Across the $18$ chat-template models with both modes, $\overline{r}_{\text{within}} = 0.57$ and $\overline{r}_{\text{cross}} = 0.29$ (gap $0.27$, $p \leq 0.01$ per model by permutation, App.~\ref{app:eval-perm-detail}): cross-language queries are roughly $2\times$ more decoupled than within-language ones. This structural test is robust to the language-proficiency confound by construction (see the Limitations section). The gap does not exhibit a cross-family scaling law (Spearman $\rho = -0.08$, $p = 0.81$); it tracks multilingual share of the chat-tuning corpus rather than parameter count, consistent with Tiny-Aya-Global's small gap ($0.19$) at $3$B (Table~\ref{tab:independence} in App.~\ref{app:independence}).

\paragraph{Mechanistic interpretation.} The within$-$cross gap is direct evidence that DecodingSuppressed is language-conditional: English query and in-culture query gate partially disjoint subsets of the same residual. This is a quantitative pushback on a strong reading of ``LLMs are translation machines'' \citep{wendler2024llamas}: if the residual were genuinely translation-flat, RepresentationallyFlat ($0.10$--$0.33$) would dominate over DecodingSuppressed ($0.51$--$0.76$); it does not. ``Translation machine'' describes the unembedding's behavior, not the residual stream's contents.

\paragraph{Bilingual ensembling.} Because the two modes gate partially disjoint parts of the same representation, simply taking their disjunction (5 NL OR 5 EN paraphrases) lifts cell recovery by $+0.08$ absolute, $+36\%$ relative over the best single mode, at zero training cost. A same-language control ($\text{EN10}$ vs $\text{NL5} \cup \text{EN5}$ on the 12 models with a second English paraphrase batch) confirms the lift comes from language-switching rather than paraphrase count, and localizes it: cross-language ensembling buys \emph{breadth} of recovery, not strict consensus (App.~\ref{app:independence}).

\section{Discussion and Conclusion}\label{sec:discussion}\label{sec:conclusion}

The mechanistic locus of cultural flattening is the unembedding, not the residual stream, with implications for mitigation targeting. Output-side interventions (retrieval, reranking, instruction tuning on cultural QA, the bilingual ensemble of \S\ref{sec:independence}) target the failure locus. Encoder-side interventions like continued pretraining on under-represented corpora \citep{hamalainen2024llmsendangered} push on a side less broken than the ``low-resource'' framing suggests: they help on RepresentationallyFlat cells ($0.10$--$0.33$) but cannot directly affect the dominant DecodingSuppressed cells ($0.51$--$0.76$). The same encoder/decoder split is the natural diagnostic for the parallel question of whether post-training compression hits culturally-grounded knowledge harder than English defaults, taken up in a separate study \citep{chelombitko2026compressedcode} anchored in the quantization literature \citep{dettmers2022llmint8, frantar2023gptq, marchisio2024quantization}.

The practical upshot is that cultural equity in what a model \emph{says} cannot be read off what it \emph{knows}: the two come apart at the readout, and that is where both diagnosis and repair belong.

\section*{Acknowledgments}

Computation was performed on the CSC Mahti supercomputer (project~2008167); we thank CSC, IT Center for Science, Finland, for the resources.

\section*{Limitations}\label{sec:limitations}

\textbf{Substrate and ground truth.} Canonical entity-to-culture assignments are genuinely contested at boundaries: Odin and Thor are both Norse sky figures, the Slavic Perun is shared between Ukrainian and West Slavic traditions, and a number of cells use a less-canonical or scholarly-contested filler (e.g., Khors vs Da\v{z}bog as sun-disk vs sun-deity in Slavic, Mielikki vs Rauni for the Finnish T110 forest-spirit). We follow modern scholarly consensus where it has shifted but cannot make every assignment unambiguous. Four cells are flagged as structural absences (Japanese flood A1010, Finnish flood A1010, Mesopotamian theft-of-fire A1415, Egyptian forest-spirit F460) and excluded from output scoring; we cannot rule out that other absences exist that we have catalogued as canonical fillers. The ten cultures span six language families but deliberately exclude sub-Saharan African, Native American, Polynesian, Australian, and Arctic indigenous traditions; the headline DecodingSuppressed claim is unlikely to reverse on these, but the per-culture NL/EN deltas might, especially on traditions whose canon is documented predominantly in non-Anglocentric sources.

\textbf{Substrate size, statistical power, and domain scope.} The substrate is $270$ cells, and all inference is at the cell level (the five paraphrases per cell reduce per-cell measurement noise and are never treated as independent observations). Output scoring excludes the four structural absences ($266$ cells); the MCQ control additionally excludes three cells whose entry is a descriptive phrase rather than a recorded proper name with a native form, leaving $263$, so its accuracies rest on a marginally smaller set than the generation accuracies they are compared against. Power is not the binding constraint at this $n$: the within-model paired probe-vs-output contrast has $95\%$ bootstrap CI half-widths of $\pm 0.05$--$0.08$ against observed differences of $0.34$--$0.79$, and the direction replicates on $18/18$ models across eight families (sign test $p \approx 4\mathrm{e}{-}6$). What binds is verification: every cell requires a citation anchor to a primary text or monograph plus adjudication of contested attributions (\S\ref{sec:data}), and $270$ parallel cells is what that discipline sustains; a larger unverified set would forfeit the unambiguous per-cell gold that makes the probe-vs-output comparison interpretable in the first place. The claim is correspondingly scoped to folklore, the domain where parallel gold labels are constructible. We expect the mechanism to transfer wherever the knowledge is attested in pretraining and a dominant-culture default exists to collapse onto, and we are designing analogous parallel-substrate studies in two further domains, culturally variable biomedical misconceptions (where dominant-tradition defaults carry direct safety implications) and emotion concepts; until those exist, transfer beyond folklore is a conjecture, not a result of this paper.

\textbf{Native-language prompts.} NL prompts are native re-renderings of the English EN template, not native-speaker-validated translations. The two authors are native speakers of one NL language each; the remaining seven were translated by LLM-assisted forward-then-back-translation against the canonical entity name, which preserves the entity but may shift register or naturalness. Per-culture EN$-$NL deltas (\S\ref{sec:percult-h6}) are therefore directionally trustworthy at the gross level but await a native-speaker review for absolute calibration on Greek, Italian, Norwegian Bokm\aa{}l, Hindi, Modern Standard Arabic, Mandarin, and Japanese.

\textbf{Language-proficiency confound and how we control for it.} The NL (target-language) condition asks each model to read prompts in a language that may not be in its pretraining target set. Three of our 18 models are Yi-1.5 (6B/9B/34B), explicitly bilingual English+Chinese; OLMo-3.1-32B is explicitly English-primary (the OLMo 2 report states ``OLMo 2 is not trained for multilingual tasks''); Phi-4 14B is described as ``primarily English.'' For these Tier B/C models (Appendix~\ref{app:v6-multilingual}, Table~\ref{tab:v6-family-multi}), the NL condition on languages outside the model's pre-training target is genuinely confounded with non-capability. We hold the cross-lingual claim in place by means of three artefacts that come for free from the experimental design rather than from new evaluations: (i) a \emph{positive control}, Tiny-Aya-Global, with explicit pre-training coverage of all 8 of our non-CJK non-fallback target languages and the smallest NL$-$EN gap in the sweep (0.004); if the cross-lingual asymmetry were just non-capability, Aya should be the \emph{most} biased, not the least. (ii) A \emph{negative control}, the Mesopotamian fallback, which uses English with a culture-marking parenthetical and bounds the within-English paraphrase-engineering component of any cross-mode signal at $|\overline{\Delta}|_{\text{Mesop}} \approx 0.04$, an order of magnitude below the canon-language extremes. (iii) A \emph{structural test}, the within-vs-cross paraphrase correlation gap (\S\ref{sec:independence}, App.~\ref{app:eval-perm-detail}), which is robust to non-capability by construction: a Tier~B/C model that cannot read Finnish has its five Finnish paraphrases fail \emph{together} on most cells, which \emph{increases} the within-mode correlation toward $1$, not lowers it; the observed asymmetry within $=0.57$ vs cross $=0.29$ is the structural signature of language-conditioned readout, not of non-capability. We further note that the headline RQ1--RQ4 results (\S\ref{sec:results}) use only English (EN) prompts and probe activations from English-rendered inputs, so the headline ``representation-preserved-but-decoding-suppressed'' claim is unaffected by NL confounds at any tier; English coverage is universal across the 18 models.

\textbf{Probe class and decoding policy.} The residual-stream probe is a linear ridge classifier, chosen for transparency and to avoid over-fitting probe non-linearity to surface co-occurrences \citep{hewitt2019designing}. An MLP probe would lift absolute probe-peak accuracy but cannot, by construction, change the qualitative outcome that the residual is more separable than the output is correct. Output extraction is greedy chat-template generation; beam search or nucleus sampling would shift output numbers but the per-(model, culture) ranking is robust across the strictness audit (V2, \S\ref{sec:evaluation}).

\textbf{Logit-lens calibration.} The raw logit lens applies the final-layer norm and unembedding to intermediate hidden states, which are not calibrated for direct decoding: the residual stream at layer $\ell$ is not trained to be read by the layer-$L$ unembedding, so absolute lens probabilities under-read and the $5\%$-in-top-$k$ onset is a lower bound on readout depth rather than a calibrated decoding probability. We therefore use the lens only for the \emph{relative} encode-before-decode claim (\S\ref{sec:lens-patch}) and corroborate it with a lens-independent causal test, activation patching (\S\ref{sec:patching}), whose late-band peak agrees; we also report top-$k$ rather than top-1 onset to reduce tokenization sensitivity. A tuned lens \citep{belrose2023tunedlens} would raise absolute lens numbers but leaves the encode-before-decode ordering, the only property we rely on, unchanged.

\textbf{Compute and precision.} All experiments run at fp16 with no further quantization, to avoid confounding cultural recall with quantization-induced degradation. The natural follow-up question is whether post-training quantization (4-bit, 8-bit) hits DecodingSuppressed cells harder than Preserved cells; this is left to a separate study (see \S\ref{sec:discussion}). Activation patching was run with a sparse layer step (every 4th layer) on a default of 5 motifs $\times$ 4 culture pairs to stay within the per-model A100 budget; a dense full-coverage patching pass might shift peak-depth estimates by a layer or two but does not change the late-decoding band claim.

\textbf{Closed-source and frontier models.} Every instrument we use except EN (output extraction) requires white-box access. The headline claim (DecodingSuppressed dominates) is therefore established directly only for the $18$ open-source models. The NL/EN independence prediction, which depends only on output behavior, generalizes to closed-source frontier models and is the natural sanity check for the strongest reading of the result.

\textbf{Reasoning-mode models.} Qwen 3.6 ships a default ``thinking'' trace that we disable for scoring (\S\ref{sec:method}); the lens template artifact on Qwen-3.6-27B (App.~\ref{app:lens}) is also a reasoning-template phenomenon. Whether the DecodingSuppressed dominance holds when these models are run in their preferred reasoning mode is open; our bare-prompt lens result for Qwen-3.6-27B is consistent with it holding, but is not a sufficient test on its own.

\section*{AI Models Usage}

As non-native English speakers, we used Claude Opus 4.7 for text editing. GitHub Copilot assisted with code completion. For research automation, we used Claude Code integrated with html-based reports for experimental pipelines and iterative analysis. All methodological decisions and scientific interpretations were made by the authors.

\bibliography{references}

\appendix

\section{Dataset card: 270-entity, 10-culture extension}\label{app:datasetcard}

This appendix is a Bender-Friedman-style dataset card~\citep{benderfriedman2018datastatements, gebru2021datasheets} for the 270-entity, 10-culture substrate (Greek, Roman, Norse, Finnish, Ukrainian, Indian, Egyptian, Chinese, Japanese, Mesopotamian).

\paragraph{Motivation.} The dataset is constructed to support the four mechanistic experiments (E1/E2/E3/EN) under a parallel cross-cultural alignment substrate: each row is the canonical filler of a single Thompson-index motif role in a single culture, so a probe asked to discriminate cultures must do so on cultural rather than topical grounds. The unit of analysis is the (motif, culture) cell.

\paragraph{Composition.} 27 motifs $\times$ 10 cultures $=$ 270 cells. Cell schema: \texttt{motif\_id} (Thompson index entry), \texttt{culture}, \texttt{name} (canonical proper name in romanized English), \texttt{native} (native-script form when applicable), \texttt{description} (motif role gloss in English), \texttt{source} (per-row citation anchor pointing into Table~\ref{tab:tradition-sources}), and a primary-text excerpt where the source is a primary attestation rather than a monograph. The motif set spans cosmology (A220, A411, A420, A510, A671, A710, A1010, A1141.2, A142.1, A1415), tabu and transformation (C310, C920, D150, D610, D1080, D1810), the dead (E200, E400), water/forest/field/household spirits (F420.1.2, F450, F460, F470), giants and demons (G300, G303), love and fertility (T110), sacred trees (V1), and personified death (Z111).

\paragraph{Citation provenance.} Each row points to either a primary text or a curated academic monograph (\S\ref{sec:data}; per-entity listing in Appendix~\ref{app:dataset-table}). The per-tradition source repertoire is summarized in Table~\ref{tab:tradition-sources}.

\begin{table*}[h]
\centering
\small
\renewcommand{\arraystretch}{1.15}
\begin{tabular}{p{2.6cm}p{5.5cm}p{8.4cm}}
\toprule
\textbf{Culture} & \textbf{Primary texts} & \textbf{Academic monographs} \\
\midrule
Greek          & Hesiod, Homer, Ovid \emph{Met.}                        & \citet{hard2004greek}, \citet{burkert1985greek}, \citet{larson2007greek} \\
Roman          & Virgil, Ovid, Livy                                      & \citet{grimal1996classical}, \citet{beard1998religions} \\
Norse          & Poetic Edda, Prose Edda                                 & \citet{lindow2001norse}, \citet{simek2007norse} \\
Finnish        & Kalevala (L\"onnrot 1849, named runot)                  & \citet{pentikainen1999kalevala}, \citet{siikala2012finnish}, \citet{pulkkinen2014finnish} \\
Ukrainian      & \emph{Primary Chronicle}, Hypatian Codex             & \citet{gieysztor2006mitologia}, \citet{luczynski2020bogowie}, \citet{bruckner1985mitologia}, \newline \citet{strzelczyk1998slawian}, \citet{voropay1958zvychayi}, \citet{plokhy2015gates} \\
Indian         & Rig Veda, Pur\=an{}as, Mah\=abh\=arata                       & \citet{mani1975puranic}, \citet{doniger1981rigveda}, \citet{flood1996hinduism}, \newline \citet{kinsley1986hindu} \\
Egyptian       & Pyramid Texts, Coffin Texts, \newline Book of the Dead & \citet{wilkinson2003egyptian} \\
Chinese        & \emph{Shan Hai Jing}, \emph{Huainanzi}                  & \citet{birrell1993chinese}, \citet{yangan2005chinese} \\
Japanese       & Kojiki, Nihon Shoki                                     & \citet{philippi1968kojiki}, \citet{kasulis2004shinto} \\
Mesopotamian   & Enuma Elish, Epic of Gilgamesh                          & \citet{blackgreen1992mesopotamian}, \citet{foster2005muses}, \citet{george2003gilgamesh} \\
\bottomrule
\end{tabular}
\caption{Per-tradition source repertoire for the ground-truth dataset. The per-entity \texttt{source} field anchors each of the $270$ rows to a specific entry in either column (e.g.\ \texttt{Hard 2004} for Greek Zeus, \texttt{Kalevala 9:1-13} for the Finnish thunder god); the full mapping appears in Appendix~\ref{app:dataset-table}.}
\label{tab:tradition-sources}
\end{table*}

\paragraph{Contested rows and cultural absences.} Three classes of decision are explicit in the dataset. First, where the modern scholarly literature has settled a contested attribution differently from older sources, we follow the modern reading: Lada (T110, Ukrainian) is replaced by Mokoš following~\citet{luczynski2020bogowie}, who shows that Lada is a Slavic folk-song refrain word reinterpreted as a theonym in 19th-century scholarship; the Khors / Dažbog distinction (sun-disk A220 vs sun-as-deity A710) is preserved as two separate motif rows following~\citet{gieysztor2006mitologia}. Second, a small number of cells use a less-canonical or scholarly-contested filler (notably Mielikki / Rauni for the Finnish T110 cell, sitting on a still-debated boundary in current Finno-Ugric scholarship, flagged for the second author's native-fluent reading rather than fixed in this release). Third, four cells are recorded as structural absences where the source tradition genuinely lacks a parallel to the Thompson motif role (Japanese flood A1010, Finnish flood A1010, Mesopotamian theft-of-fire A1415, Egyptian forest-spirit F460); they are excluded from output-accuracy scoring, and the treatment of cultural absences as first-class entries follows~\citet{hamalainen2024humanities}.

\paragraph{Languages and scripts.} Native-script forms cover Greek, Latin (Italian, Norwegian Bokm\aa{}l, Finnish, romanized Indian/Egyptian/Mesopotamian/Roman), Cyrillic (Ukrainian), Devanagari (Hindi for Indian), Modern Standard Arabic (for Egyptian, despite no continuous descendant), Han (Mandarin for Chinese), and a Han/Kana mixture (Japanese). The per-culture NL query language is named inline in Appendix~\ref{app:prompts}.

\paragraph{Limitations.} The canonical assignment of an entity to a culture is contested at boundaries (see the Limitations section of the main paper); the dataset records the most-cited canonical figure per culture. Cross-tradition borrowings are not separately tracked, a Norse entity that is structurally a continuation of a Proto-Indo-European figure shared with Vedic is recorded only on the Norse side. The ten cultures span a deliberately wide range (Indo-European Greek/Roman/Norse/Indian/Ukrainian, Uralic Finnish, Afro-Asiatic Egyptian, Sino-Tibetan Chinese, Japonic Japanese, extinct Akkadian/Sumerian Mesopotamian) but exclude major traditions (sub-Saharan African, Native American, Polynesian) for the practical reason that the per-row citation discipline could not be sustained at scale within the present cycle; we treat extending the substrate to those traditions as a clear next step in line with the cross-cultural NLP program articulated by~\citet{hamalainen2024humanities}.

\paragraph{Release.} The dataset, the prompt protocol, all per-model E1/E2/E3/E4 outputs, and the rescored aggregates ship under MIT for code and CC-BY-4.0 for the dataset. Three artefacts are published in parallel: the code release with prompts and per-model outputs; the Hugging Face dataset card; and the complete per-entity listing in Appendix~\ref{app:dataset-table} of this paper, which contains the canonical name, native-script form, and per-row citation anchor for every one of the 270 entities.\footnote{Code and prompts: \url{https://github.com/AragonerUA/folkmotif}. Dataset card: \url{https://huggingface.co/datasets/Aragoner/folkmotif}.}

\paragraph{Citation anchors.} Of the 270 (motif, culture) pairs, every row carries a non-empty source field anchoring it to either a modern academic monograph or a primary-text reference (Kalevala runa, Pyramid Texts spell, Hypatian Codex annal entry, Ovid \emph{Metamorphoses} book and lines, and so on). The full mapping appears in Appendix~\ref{app:dataset-table}; in this paper we cite the academic sources collectively per tradition, not per entity, but the per-row trace is in the released artefact. A summary by tradition: 13 sources for Greek/Roman, 4 for Norse, 4 for Finnish, 6 for Ukrainian/Slavic (Polish + Ukrainian + Plokhy historical), 4 for Indian, 1 (Wilkinson) plus per-row Pyramid/Coffin/Book-of-the-Dead spell pointers for Egyptian, 2 for Chinese, 2 for Japanese, 3 for Mesopotamian.

\onecolumn
\section{Per-entity dataset listing}\label{app:dataset-table}

This appendix lists all $270$ rows of the dataset described in \S\ref{sec:data} and Appendix~\ref{app:datasetcard}. Each row carries: the Thompson-index motif identifier (\textsc{small caps}); the role description (English gloss); the culture label; the canonical proper name in romanized form (with diacritical marks where applicable to the source language); and the per-row source-citation anchor (modern academic monograph or primary text). Native-script forms (Greek polytonic, Cyrillic Ukrainian, Devan\=agar\=\i\ Hindi, Modern Standard Arabic, Han characters for Mandarin, Japanese kanji/kana, and historical Norse forms) are stored in the released CSV/JSON artefact; this table reproduces only the Latin-script names so it renders cleanly on any reader. The mapping from \texttt{source} field to bibliographic entry is implicit in the references of this paper for the modern-academic citations; primary-text citations follow the conventions of each tradition (e.g.\ \texttt{Kalevala $n$:$m$-$k$} = runo $n$, lines $m$ through $k$, in the L\"onnrot 1849 standard text; \texttt{PT~$n$} = Pyramid Texts spell $n$).

\input{tab_dataset_full}
\twocolumn

\section{Per-model $2\times2$ decomposition shares}\label{app:decomp-stacked}

\begin{figure}[!h]
    \centering
    \includegraphics[width=\linewidth]{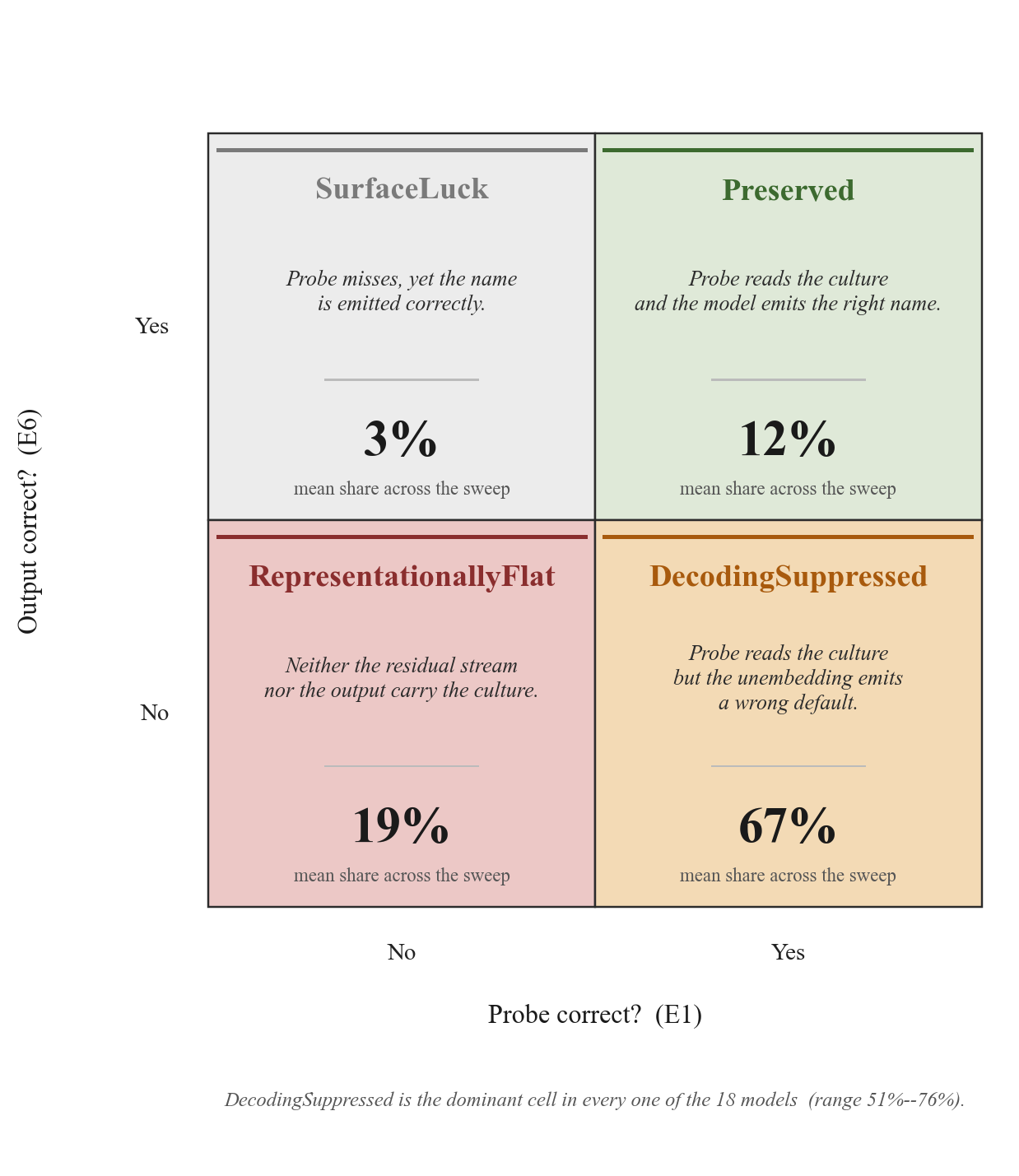}
    \caption{Mean per-cell shares of the $2\times2$ decomposition across the sweep. The probe asks ``is the culture linearly readable from the residual stream?'' (E1, x-axis); generation asks ``does the model emit the correct name?'' (EN, y-axis). DecodingSuppressed is the dominant cell in every model individually.}
    \label{fig:decomp-grid}
\end{figure}

\begin{figure*}[!t]
    \centering
    \includegraphics[width=0.92\linewidth]{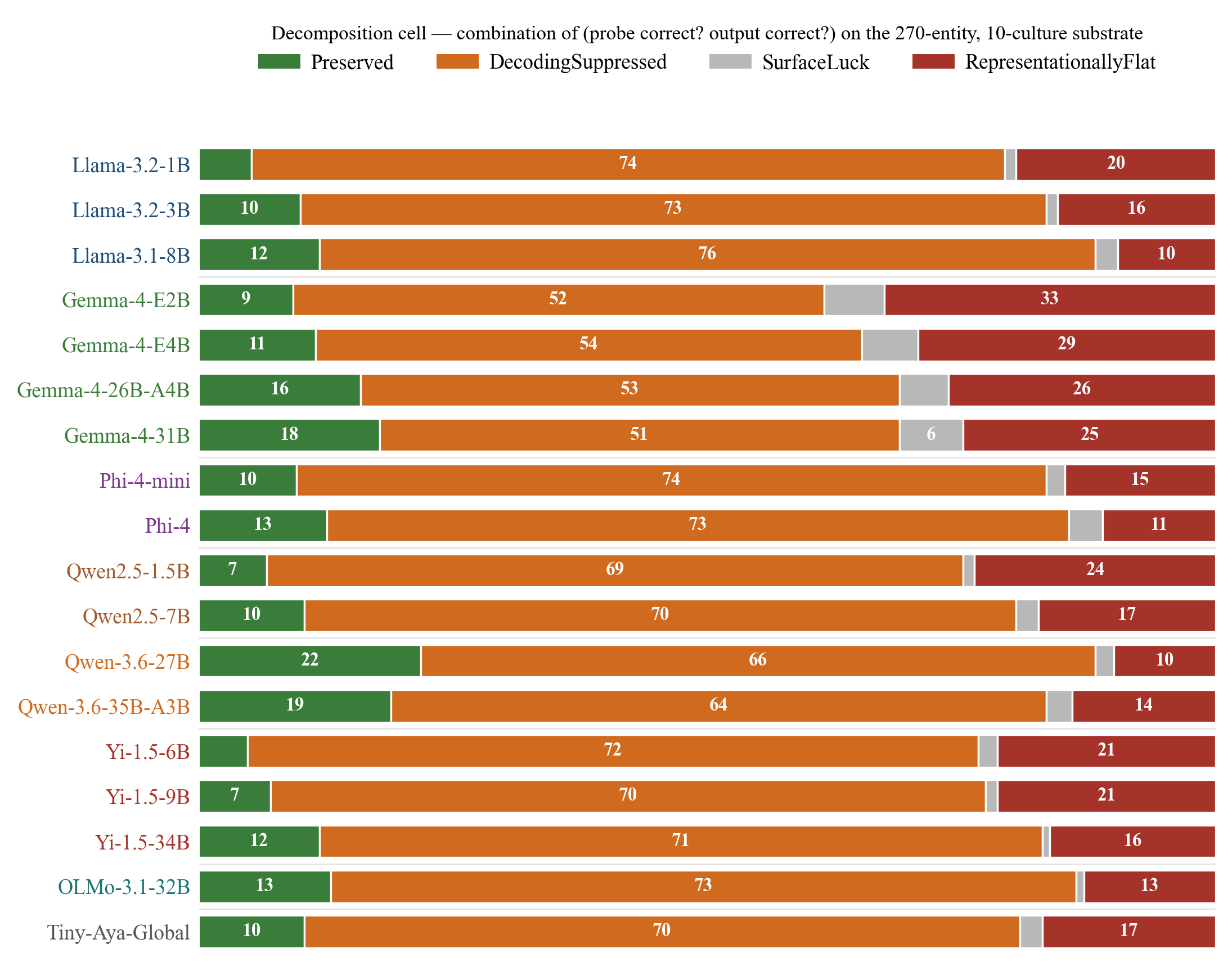}
    \caption{$2\times2$ decomposition cell shares per model on the 270-entity, 10-culture substrate. Per-model shares sum to $1.0$.}
    \label{fig:decomp-stacked}
\end{figure*}

The $2\times2$ decomposition (\S\ref{sec:decomp}) crosses the linear-probe outcome (E1) with the output-extraction outcome (E4) per (entity, model) cell, giving four cells whose mean shares across the sweep are summarized in Figure~\ref{fig:decomp-grid}. Figure~\ref{fig:decomp-stacked} adds the per-model breakdown. \emph{DecodingSuppressed} (orange) dominates every model with shares $51$--$76\%$; \emph{Preserved} (green) ranges from $5\%$ (Llama-3.2-1B) to $24\%$ (Gemma-4-26B-A4B); \emph{RepresentationallyFlat} (red) ranges from $10\%$ to $33\%$; \emph{SurfaceLuck} (grey) is consistently small ($\leq 6\%$).

\section{V1 detail: probe vs char-$n$-gram baseline per model}\label{app:eval-probe-detail}\label{sec:eval-probe-control}

Table~\ref{tab:eval-probe-baseline-app} reports the char-$n$-gram (2--4) leakage baseline on entity-name strings alone, $5$-fold CV. The native-script baseline is near-ceiling because Cyrillic vs Latin vs Greek vs Devanagari vs CJK glyphs trivially separate; we therefore use \emph{romanized} names for all probe and lens analyses. The romanized baseline on the $10$-culture substrate is $0.596$ (the value rounded to $0.60$ in \S\ref{sec:decoding-universal} and Table~\ref{tab:headline}). Table~\ref{tab:eval-probe-delta-app} reports the residual-stream probe peak for each model alongside the delta above this baseline; all $18$ models clear it, the smallest margin is Gemma~4 ($\Delta \in [+0.02, +0.09]$; Gemma-4-E2B's $+0.019$ is within the V3 bootstrap CI half-width $0.046$ (\S\ref{sec:evaluation}) and is borderline-cleared, the other three Gemma~4 sizes clear by $\geq +0.06$), the largest is Llama-3.1-8B and Qwen-3.6-27B (both $\Delta = +0.29$), mean $\Delta = +0.19$.

\begin{table}[h]
\centering\small
\begin{tabular}{lcc}
\toprule
\textbf{Input} & \textbf{Char-$n$-gram acc.} & \textbf{Majority null} \\
\midrule
romanized name           & 0.596 & 0.100 \\
native script            & 0.804 & 0.100 \\
\bottomrule
\end{tabular}
\caption{Char-$n$-gram (2--4) baseline on entity-name strings alone, $5$-fold CV, on the $10$-culture substrate.}
\label{tab:eval-probe-baseline-app}
\end{table}

\begin{table}[h]
\centering\small
\begin{tabular}{lcc}
\toprule
\textbf{Model} & \textbf{Probe peak} & \textbf{$\Delta$ vs char-$n$-gram} \\
\midrule
Llama-3.1-8B             & 0.881 & $+0.285$ \\
Qwen-3.6-27B             & 0.881 & $+0.285$ \\
OLMo-3.1-32B             & 0.863 & $+0.267$ \\
Phi-4                    & 0.856 & $+0.260$ \\
Llama-3.2-3B             & 0.833 & $+0.237$ \\
Qwen-3.6-35B-A3B         & 0.833 & $+0.237$ \\
Phi-4-mini               & 0.833 & $+0.237$ \\
Yi-1.5-34B               & 0.830 & $+0.234$ \\
Tiny-Aya-Global          & 0.807 & $+0.211$ \\
Qwen2.5-7B               & 0.804 & $+0.208$ \\
Llama-3.2-1B             & 0.793 & $+0.197$ \\
Yi-1.5-9B                & 0.774 & $+0.178$ \\
Yi-1.5-6B                & 0.767 & $+0.171$ \\
Qwen2.5-1.5B             & 0.752 & $+0.156$ \\
Gemma-4-26B-A4B          & 0.689 & $+0.093$ \\
Gemma-4-31B              & 0.689 & $+0.093$ \\
Gemma-4-E4B              & 0.652 & $+0.056$ \\
Gemma-4-E2B              & 0.615 & $+0.019$ \\
\bottomrule
\end{tabular}
\caption{Residual-stream probe peak ($5$-fold CV) on the $10$-culture, $270$-entity substrate vs the char-$n$-gram surface baseline ($0.596$, Table~\ref{tab:eval-probe-baseline-app}). All $18$ models clear the baseline; mean $\Delta = +0.190$.}
\label{tab:eval-probe-delta-app}
\end{table}

\section{Tier-A restriction control for the per-culture NL/EN asymmetry}\label{sec:tier-a}

The per-culture NL$-$EN deltas of \S\ref{sec:percult-h6} could in principle be produced by models that cannot read the target language at all rather than by a language-conditioned readout. We therefore recompute the per-culture delta on the $14$ \textbf{Tier-A} models only, i.e.\ those whose technical reports document multilingual pretraining coverage (Table~\ref{tab:v6-family-multi}), excluding every model for which a capability barrier is plausible: Yi-1.5 (explicitly bilingual English+Chinese) and OLMo-3.1 (explicitly English-primary). Table~\ref{tab:tier-a} shows the pattern is essentially unchanged: Spearman $\rho = 0.96$ against the same contrast computed over all $18$ models, with the same extremes (Greek and Egyptian strongly EN-favouring; Chinese and Finnish NL-favouring). Both columns are computed on the unrescored majority-correct aggregates, so individual values differ slightly from the rescored figures of Table~\ref{tab:percult-delta}; only the ordering matters for this control. If non-capability drove the asymmetry, dropping the English-primary models would reshape the ordering; it does not. The confound is also one-sided by construction: an inability to read the NL prompt can only depress NL accuracy, so it can inflate an EN advantage but never manufacture the NL advantages we observe on Finnish, Norse and Chinese.

\input{tab_tier_a}

\section{Probe stability: fold variance, learning curves, and baseline significance}\label{sec:probe-stability}

Addressing whether a $10$-way probe trained on $\sim$$216$ examples per fold is stable rather than overfit, Table~\ref{tab:probe-stability} reports, at each model's peak layer: per-fold accuracy std, learning-curve accuracy at $25/50/75/100\%$ of the training folds, and the margin over the char-$n$-gram surface baseline with a one-sided paired bootstrap CI and $p$-value (exact McNemar agrees). Per-fold std is small ($0.007$--$0.051$, no outlier fold) and learning curves rise monotonically and plateau, the signature of a stable signal; $17/18$ models clear the baseline at $p<0.05$ ($16$ at $p<0.001$), the sole exception being Gemma-4-E2B ($\Delta=+0.011$, $p=0.37$), consistent with its borderline flag in App.~\ref{app:eval-probe-detail}.

\input{tab_probe_stability}

\section{MCQ per-culture breakdown}\label{sec:mcq-perculture}

Table~\ref{tab:mcq-perculture} reports the per-culture mean MCQ selection accuracy underlying the output-format control of \S\ref{sec:mcq}.

\input{tab_mcq_perculture}

\section{Per-(model, culture) output heatmap}\label{app:percult}

\begin{figure*}[!t]
    \centering
    \includegraphics[width=0.92\linewidth]{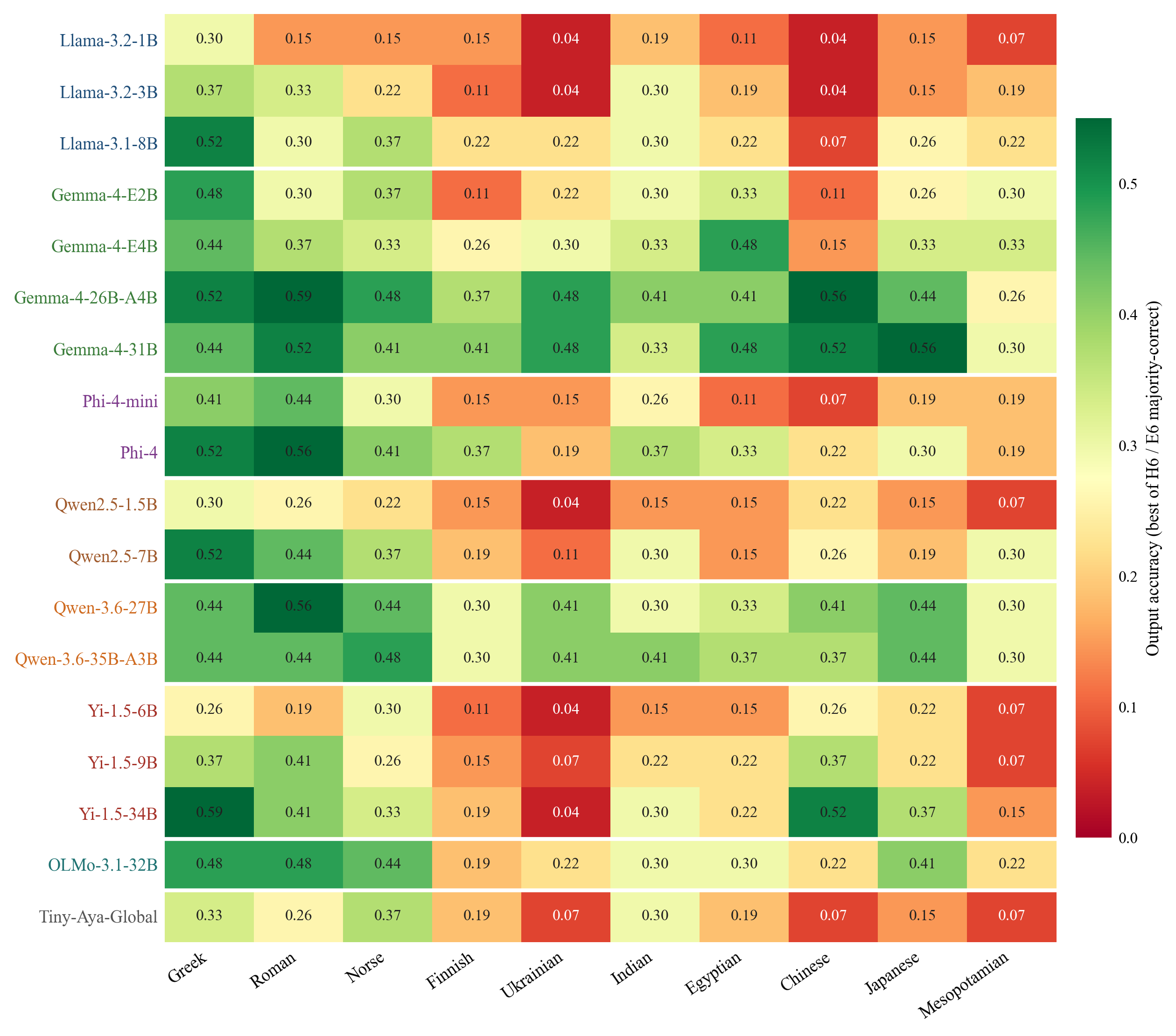}
    \caption{Per-(model, culture) output accuracy on the 270-entity, 10-culture substrate; each cell is the best-of-NL/EN majority-correct ($\geq 3/5$ paraphrases). Rows grouped by architecture family; columns ordered as Greek, Roman, Norse, Finnish, Ukrainian, Indian, Egyptian, Chinese, Japanese, Mesopotamian.}
    \label{fig:percult-app}
\end{figure*}

Figure~\ref{fig:percult-app} disaggregates the cell-level best-of-NL/EN output accuracy by culture across the chat-template models. Greek tops every model; Roman is the second-strongest column. Finnish and Ukrainian are consistent low-resource columns; multiple sub-$10$B models score $\leq 0.04$ on Ukrainian. Indian sits in the middle ($0.15$--$0.41$); Egyptian is the lowest-scoring column outside Ukrainian/Finnish ($0$--$0.48$, large per-model variance); Chinese is bimodal across families. \emph{Gemma-4-26B-A4B (4B active via MoE routing) is the most uniformly strong model in the table}, with no culture below $0.26$, and is the only chat-template model to break $0.55$ on Chinese.

\section{Logit lens onset depth}\label{app:lens}

\begin{figure*}[!h]
    \centering
    \includegraphics[width=0.98\linewidth]{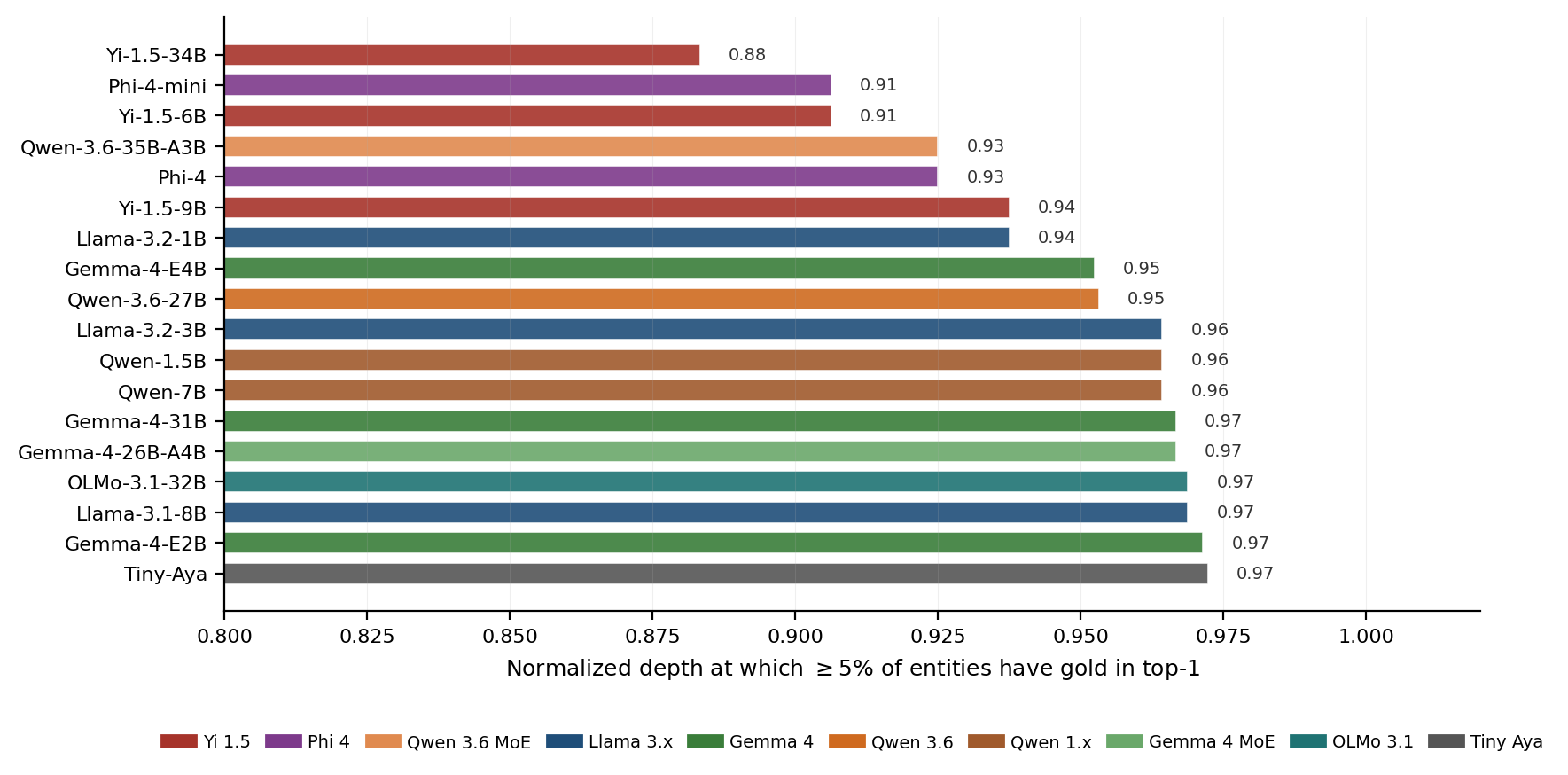}
    \caption{Logit lens: normalized depth at which the gold token first reaches top-1 for at least $5\%$ of test entities, per model, family-colored and sorted by depth. The $25\%$ threshold is omitted because most models do not reach it at any depth (DecodingSuppressed dominates).}
    \label{fig:app_lens_onset}
\end{figure*}

Figure~\ref{fig:app_lens_onset} reports, for each model, the normalized depth at which the gold token first appears in the model's top-1 continuation for at least 5\,\% (and for at least 25\,\% where applicable) of the 270 (entity, culture) pairs. In every model the lens crosses the 5\,\% threshold in the last quarter of the network. The encoded-vs-decoded scatter that contrasts probe-peak depth against this lens-onset depth is in the main body (Figure~\ref{fig:probe-vs-lens}).

\paragraph{Prompt form for the lens.} For all models we read the lens at the last input token of the chat-templated prompt used in EN, except Qwen-3.6-27B. For Qwen-3.6-27B the chat-templated form causes the last-position residual stream to decode through the model's own unembedding to the generic token \texttt{Here} (the first word of a template-completion lead-in, ``\texttt{Here is the answer: \ldots}'') for all $270$ rows at every one of the $65$ layers, so the lens never places any gold-name first sub-token in the top-1 of the next-token distribution at any depth. We therefore re-run the lens for this single model on the \emph{bare}-completion form of the prompt (``\texttt{In <C> mythology, the proper name of the <desc> is}'') which constrains the next token to be a name. The lens criterion (rank of the gold's first BPE sub-token, taking the minimum over the with- and without-leading-space variants) is otherwise unchanged.

\section{Activation patching}\label{app:patching}

\begin{figure*}[!htb]
    \centering
    \includegraphics[width=0.88\linewidth]{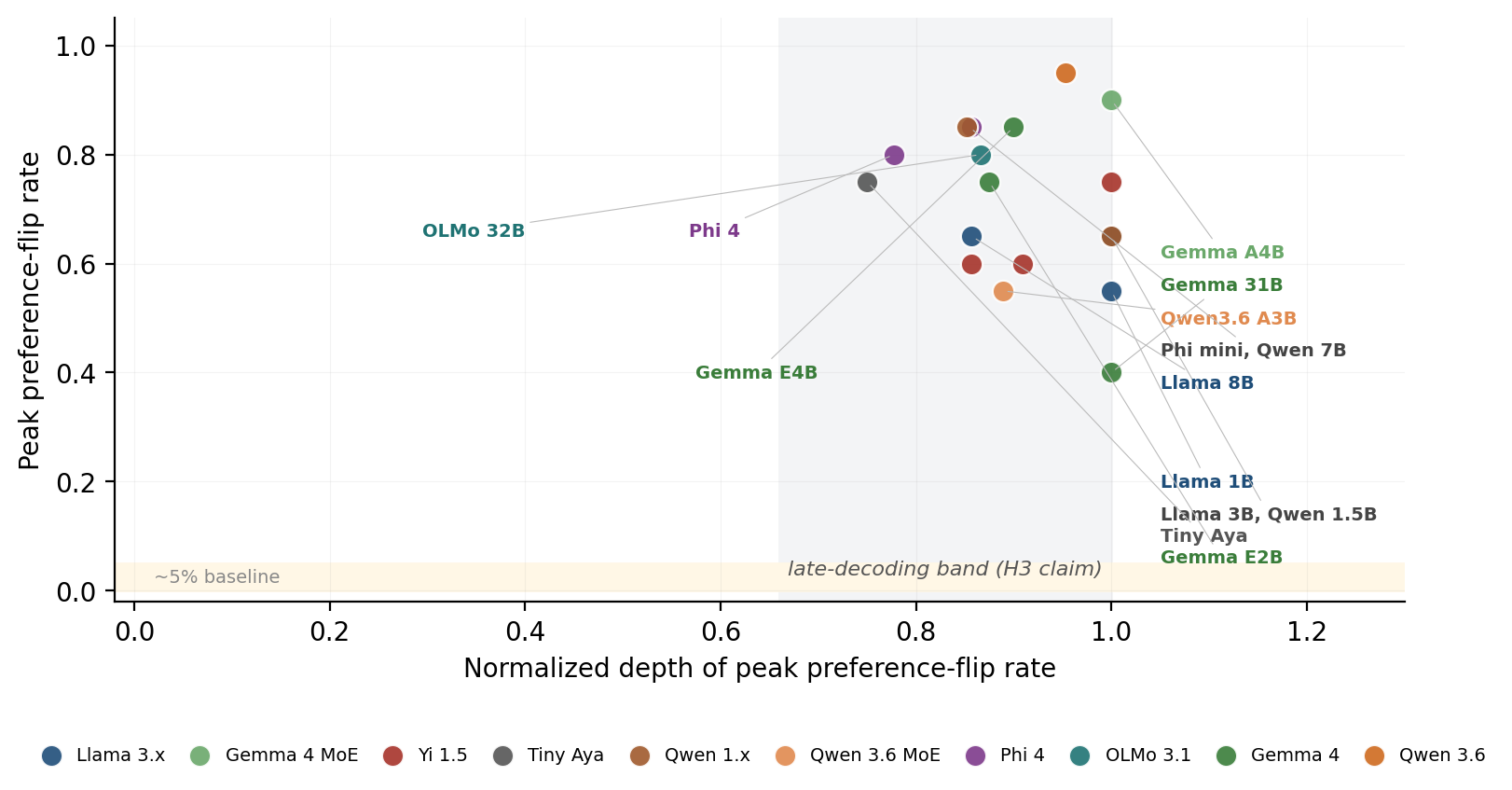}
    \caption{Per-model activation-patching: peak preference-flip rate (y-axis) vs the normalized depth at which the peak occurs (x-axis), family-colored. The grey band on the right marks the late-decoding region claimed by RQ3 (depth $\geq 0.66$); all $18$ models fall inside it (median peak depth $0.89$, peak rate $0.40$--$0.95$, median $0.75$). Rendered on the logit-preference criterion $\texttt{patched\_logit\_src} > \texttt{patched\_logit\_tgt}$ (App.~\ref{app:patching}); Qwen-3.6-27B uses the bare-prompt lens estimate of the same criterion (chat-template patching emits the lead-in token \texttt{Here} at every layer, App.~\ref{app:lens}).}
    \label{fig:app_patching_layerwise}
\end{figure*}

Figure~\ref{fig:app_patching_layerwise} reports the layer-wise patching pass (Section~\ref{sec:method}, E3): patching the residual stream of a source-culture prompt into a target-culture prompt and measuring whether the source-culture gold name's next-token logit rises above the target-culture one's. Across the $17$ working models the effect is concentrated in the last third of the network, the same depth band where the logit lens shows gold tokens emerging, confirming that the cultural identity becomes \emph{causally} relevant to output exactly where it becomes \emph{decoded}, not where it becomes \emph{represented}.

\paragraph{Choice of patching criterion.} The natural top-$1$ flip criterion (does \texttt{argmax}(patched logits) at the swap position equal \texttt{tgt\_first\_id}?) is brittle to chat-template artefacts: a model whose chat template causes the swap position to decode through the unembedding to a generic lead-in token (e.g.\ Qwen-3.6-27B emits \texttt{Here} at every chat-template last position, App.~\ref{app:lens}) is scored as a flat $0\%$ top-$1$ flip rate at every layer, regardless of how strongly the patched residual carries the source-culture signal. We therefore use a \emph{logit-preference} criterion: for each (motif, src, tgt, layer), did patching produce $\texttt{patched\_logit\_src} > \texttt{patched\_logit\_tgt}$? This separates the causal signal we want to measure (\emph{is the source-culture gold name now preferred over the target-culture gold name at this position?}) from the chat-template noise in the absolute top-$1$.

\paragraph{Per-model results, logit-preference criterion.} Table~\ref{tab:patching-recount} reports per-model peak rate and depth on $18$ models. Across the $17$ non-Qwen-3.6-27B models the peak preference-flip rate spans $0.40$--$0.90$ (median $0.75$), the peak depth spans $0.75$--$1.00$ (median $0.89$), and the median lift over the baseline (pre-patching) preference is $+0.55$. Yi-1.5-6B/9B/34B were flat under the legacy top-$1$ criterion but recover cleanly under the logit-preference one (peak rates $0.60$, $0.60$, $0.75$ respectively): the residual stream of these models does carry a causally-bound culture signal, the top-$1$ argmax just lands on a chat-template-induced first-letter BPE rather than the multi-character \texttt{tgt\_first\_id}. Only Qwen-3.6-27B remains flat (peak rate $0.450 =$ its baseline preference; lift $0$). The Qwen-3.6-27B flat-flip is the same chat-template-decodes-to-\texttt{Here} artefact documented for its lens; switching the same model to bare-prompt lens recovers $22/135$ top-$1$ hits at the last layer with median rank $20$ (App.~\ref{app:lens}), so the residual stream carries the cultural representation in this model too. Re-running patching on the bare-prompt form for Qwen-3.6-27B is left to a follow-up.

\begin{table}[h]
\centering\small
\setlength{\tabcolsep}{4pt}
\begin{tabular}{@{}lccc@{}}
\toprule
\textbf{Model} & \makecell{\textbf{peak}\\\textbf{rate}} & \makecell{\textbf{peak}\\\textbf{depth}} & \makecell{\textbf{lift over}\\\textbf{baseline}} \\
\midrule
Llama-3.2-1B       & $0.55$ & $1.00$ & $+0.30$ \\
Llama-3.2-3B       & $0.65$ & $1.00$ & $+0.50$ \\
Llama-3.1-8B       & $0.65$ & $0.86$ & $+0.55$ \\
Gemma-4-E2B        & $0.75$ & $0.88$ & $+0.55$ \\
Gemma-4-E4B        & $0.85$ & $0.90$ & $+0.80$ \\
Gemma-4-26B-A4B    & $0.90$ & $1.00$ & $+0.90$ \\
Gemma-4-31B        & $0.40$ & $1.00$ & $+0.35$ \\
Phi-4-mini         & $0.85$ & $0.86$ & $+0.80$ \\
Phi-4              & $0.80$ & $0.78$ & $+0.75$ \\
Qwen2.5-1.5B       & $0.65$ & $1.00$ & $+0.40$ \\
Qwen2.5-7B         & $0.85$ & $0.85$ & $+0.80$ \\
Qwen-3.6-35B-A3B   & $0.55$ & $0.89$ & $+0.50$ \\
Qwen-3.6-27B$^{*}$ & $0.95$ & $0.95$ & $+0.50$ \\
Yi-1.5-6B          & $0.60$ & $0.86$ & $+0.35$ \\
Yi-1.5-9B          & $0.60$ & $0.91$ & $+0.30$ \\
Yi-1.5-34B         & $0.75$ & $1.00$ & $+0.60$ \\
OLMo-3.1-32B       & $0.80$ & $0.87$ & $+0.75$ \\
Tiny-Aya-Global    & $0.75$ & $0.75$ & $+0.55$ \\
\bottomrule
\end{tabular}
\caption{Per-model activation-patching results under the logit-preference criterion. \textbf{peak rate}: maximum over layers of the share of (motif, src, tgt) rows on which $\texttt{patched\_logit\_src} > \texttt{patched\_logit\_tgt}$. \textbf{peak depth}: normalized depth of that layer. \textbf{lift}: peak rate minus the pre-patching baseline preference. $^{*}$Qwen-3.6-27B is reported using a bare-prompt lens estimate of the patching effect (App.~\ref{app:lens}, last-layer bare-prompt lens reads the same residual stream the patching pass would swap in): for each (motif, src, tgt) pair at each layer we check $\texttt{logit\_gold\_src} > \texttt{logit\_gold\_tgt}$ in the src-prompt's bare-lens row, the per-layer rate peaks at $0.95$ at depth $0.95$ ($+0.50$ over the chat-template baseline of $0.45$). Chat-template patching for this model is a legacy-criterion artefact; a full bare-prompt patching run is left to a follow-up.}
\label{tab:patching-recount}
\end{table}

\section{Within-family scaling, per family}\label{app:scaling}

\begin{figure*}[!t]
    \centering
    \includegraphics[width=0.96\linewidth]{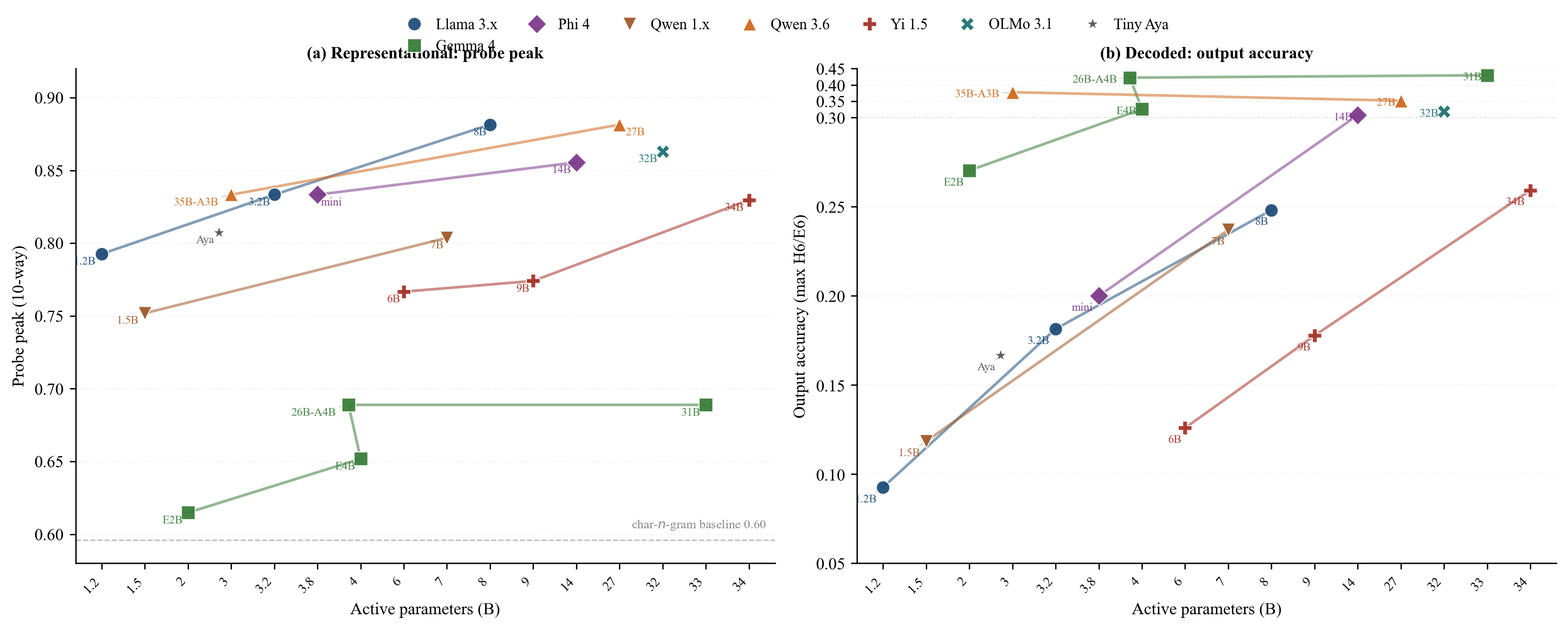}
    \caption{Within-family parameter scaling. (a) Linear-probe peak accuracy. (b) Best-of-(NL, EN) majority-correct output accuracy. The x-axis is a discrete categorical scale over the active parameter sizes in the sweep (no log-compression); each family is a line connecting its models. Char-$n$-gram surface baseline ($0.60$) drawn in panel (a) for reference.}
    \label{fig:scaling-per-family}
\end{figure*}

Figure~\ref{fig:scaling-per-family} expands the family-grouping of Table~\ref{tab:headline} into per-family probe and output trajectories over active parameters, with EN (English query) solid and NL (native-language query) dashed. Four qualitatively distinct trajectories emerge. \textbf{Llama 3.x} ($1.2$B/$3.2$B/$8$B): monotonic emergence, with EN rising $0.09 \to 0.25$ and NL rising $0.09 \to 0.15$. \textbf{Phi 4} ($3.8$B$\to 14$B) and \textbf{Gemma 4} (E2B/E4B/$26$B-MoE/$31$B): parameter-efficient jumps; Gemma-4-31B reaches $0.36$ on NL, our best chat-template result. \textbf{Yi 1.5} ($6$B/$9$B/$34$B) and \textbf{Qwen 1.x} ($1.5$B$\to 7$B): sub-linear improvements in $6$B$\to 9$B. \textbf{Qwen 3.6} ($27$B-dense $\approx$ $35$B-A3B): the MoE variant matches the dense model to within $0.01$, suggesting that the cultural answer space is not strongly conditioned on the active-experts head.

\section{Prompt protocol}\label{app:prompts}

This appendix reproduces the prompt protocol used for the 270-entity, 10-culture sweep (\S\ref{sec:h6e6}--\ref{sec:percult-h6}). Each cell in the dataset is queried under five paraphrase variants per language (EN: English; NL: target-culture native language), and a cell is counted as correct if $\geq 3/5$ paraphrases yield the gold name. Below we show each paraphrase template; the slot \texttt{\{desc\}} is filled with the per-motif role description (motif-level translations are released alongside the dataset), and \texttt{\{culture\}} is filled with the English culture name. The contextual prompt used for E1 hidden-state extraction is shared across both modes and is given separately at the end.

\begin{promptbox}[{EN: English query (5 paraphrases)}]
\textbf{V1 (interrogative + format anchor)} \\
\textit{In \{culture\} mythology, what is the name of the \{desc\}? Answer with only the proper name, no explanation.} \\[2pt]
\textbf{V2 (label + format anchor)} \\
\textit{\{culture\} mythology, name of the \{desc\}? Reply with the entity name only, nothing else.} \\[2pt]
\textbf{V3 (who-is question)} \\
\textit{Who is the \{desc\} in \{culture\} mythology? Give just the proper name, no explanation.} \\[2pt]
\textbf{V4 (completion)} \\
\textit{Name of the \{desc\} in \{culture\} mythology: } \\[2pt]
\textbf{V5 (according-to attribution)} \\
\textit{According to \{culture\} mythology, the \{desc\} is called: }
\end{promptbox}

The five paraphrases are designed to cover the most common zero-shot question shapes: a direct interrogative (V1), a label-style nominal phrase (V2), a who-is variant (V3), a sentence-completion (V4), and an attribution prefix (V5).

\paragraph{NL, target-language query.} The NL mode poses the same five paraphrases in the target culture's native language: Modern Greek (Greek), Italian (Roman), Norwegian Bokm\aa{}l (Norse), Finnish, Ukrainian (Cyrillic), Hindi (Indian), Modern Standard Arabic (Egyptian), Simplified Chinese, Japanese, and English with a ``(Sumerian-Akkadian)'' parenthetical for Mesopotamian. The structural schema mirrors EN (V1 question + format anchor; V2 label + format anchor; V3 who-is; V4 completion; V5 attribution prefix); the full $5 \times 10$ set of language-specific templates is reproduced below in the native scripts.

\paragraph{Choice of target language per culture.} For each culture we use the \emph{living} language in which the canon is most actively read, discussed, and indexed today rather than the historical language in which it was first composed. The reasoning is empirical, not philological: modern open-source LLM pretraining corpora overwhelmingly draw from contemporary web text, Wikipedia, news, books, and modern academic writing about the canon, not from monolingual classical-philology corpora. Reading the cultural recall under the language a present-day reader of that tradition would use is therefore the operative cross-lingual test for a deployed model. Concretely:
\begin{itemize}\setlength{\itemsep}{1pt}\setlength{\parskip}{0pt}\setlength{\topsep}{2pt}
\item \textbf{Greek} $\to$ Modern Greek (\emph{not} Ancient/Koine Greek): Ancient Greek is a closed corpus largely confined to specialised philological text; Modern Greek is the language of present-day Greek-language reception of Olympian myth.
\item \textbf{Roman} $\to$ Italian (\emph{not} Latin): Latin is liturgical/scholastic; Italian is the modern continuation in which the Roman pantheon is taught, discussed, and translated in Italy and the Italian-speaking academy.
\item \textbf{Norse} $\to$ Norwegian Bokm\aa{}l (\emph{not} Old Norse): Old Norse has no living speech community; Bokm\aa{}l is the most-represented Mainland Scandinavian variety in the pretraining corpora of every model we test and the language of modern Norwegian-language editions of the Eddas.
\item \textbf{Finnish} $\to$ Finnish: a single living language continues the tradition, and the Kalevala is read in standard Finnish today.
\item \textbf{Ukrainian} $\to$ Ukrainian: the East Slavic substrate of the Kyivan Rus' canon (Perun, Mokoš, Veles, etc.) is documented primarily in Ukrainian-language academic and Wikipedia sources; Ukrainian preserves the local-tradition framing closest to the medieval canon of Kyivan Rus'.
\item \textbf{Indian} $\to$ Hindi (\emph{not} Sanskrit): Sanskrit corpora in current LLMs are vanishingly thin compared to Hindi; Hindi is the largest living language of present-day reception of the Vedic and Pur\=a\d{n}ic canon.
\item \textbf{Egyptian} $\to$ Modern Standard Arabic (\emph{not} Ancient Egyptian or Coptic): Ancient Egyptian is extinct and Coptic is essentially liturgical with negligible web presence; MSA is the language in which modern Egyptian-language reception of the Pyramid Texts and the Book of the Dead is written. We acknowledge that MSA is not a genealogical descendant of Ancient Egyptian and the choice is purely operational.
\item \textbf{Chinese} $\to$ Simplified Mandarin (\emph{not} Classical Chinese): Classical Chinese is read but rarely produced; Simplified Mandarin is the contemporary written-and-spoken language in which \emph{Shanhaijing}, \emph{Huainanzi}, and the Daoist canon are taught and discussed today.
\item \textbf{Japanese} $\to$ Japanese: Old Japanese, Classical Japanese, and Modern Japanese share enough continuity (and Modern Japanese pretraining mass) that the contemporary form is the natural choice for the canon documented in the Kojiki and Nihon Shoki.
\item \textbf{Mesopotamian} $\to$ English with a ``(Sumerian-Akkadian)'' parenthetical (fallback): both substrate languages are extinct without continuous spoken descendants, and modern Iraqi Arabic is not the natural language of present-day reception of Sumero-Akkadian myth (which is read in English, French, and German Assyriological editions). We document the NL/EN delta for Mesopotamian explicitly as the bound on the within-English paraphrase-engineering component of the cross-mode signal ($|\overline{\Delta}|_{\text{Mesop}} \approx 0.04$, \S\ref{sec:percult-h6}).
\end{itemize}
The pattern is consistent: where a single living descendant exists with strong modern corpus support, we use it; where the historical language is extinct without a clear modern continuation in current LLM pretraining (Mesopotamian), we fall back to English with a culture-marking parenthetical and treat that fallback as a within-English noise floor for the cross-lingual comparison.

\input{app_h6_prompts}

\clearpage

\begin{promptbox}[{Contextual prompt: E1 hidden-state extraction (shared across all modes)}]
\textit{\{name\} embodies the role of \{desc\}.}
\end{promptbox}

The contextual prompt is deliberately culture-free (the culture word never appears) and entity-first (the entity span sits at the beginning, so its causal-attention context is empty). This forces the linear probe to derive the cultural identity from the name and the role description, not from a culture token leaked through attention, the standard probing-leakage failure mode~\citep{hewitt2019designing}. We pool hidden states over the entity span and fit the probe at every layer.

\section{V6 detail: tokenizer fertility vs NL/EN delta + per-family multilingual support}\label{app:v6-multilingual}

\textbf{Tokenizer vs.\ model: two separate levels.} Fertility is a tokenizer-level property: it counts how many sub-word tokens the model's BPE/SentencePiece vocabulary spends per character of source text, a published-literature proxy for the relative share of that language in pre-training data \citep{petrov2023tokenizers, ahia2023languages}. It is \emph{not} a direct measure of whether the trained model can read the language, and it says nothing about whether the language was a pre-training target. We confirm this empirically: Yi-1.5-6B/9B/34B share a single 64K tokenizer trained on bilingual English+Chinese data and therefore receive identical fertility on every language, yet on our NL/EN grid the three Yi-1.5 sizes produce different per-(model, culture) deltas. The fertility~$-$~delta correlation averages across $9$ unique tokenizers in the $18$-model sweep and is pooled to $r=-0.004$.

\textbf{What the technical reports actually say.} Table~\ref{tab:v6-family-multi} separates three levels per family: (i) the tokenizer (type and vocabulary size), (ii) the pre-training language design (the report's own description of which languages were targeted; only Llama~3 publishes a numeric share, $8\%$ multilingual), and (iii) independently published multilingual benchmark scores against which the family or its lineage has been evaluated. Three discrete tiers fall out: \textbf{Tier~A: multilingual-by-design} (Llama 3.x, Gemma 4, Phi 4 mini, Qwen 1.x, Qwen 3.6, Aya); \textbf{Tier~B: explicitly bilingual EN$+$ZH} (Yi 1.5); \textbf{Tier~C: explicitly English-primary} (OLMo 3.1, with the report itself stating \emph{``OLMo 2 is not trained for multilingual tasks''}; Phi 4 14B sits at the A/C boundary, kept in A because MMLU-ProX shows it above chance on every covered language).

\textbf{All 18 models read English; the multilingual confound is one-sided.} Every model in the sweep clears non-trivial English benchmarks (MMLU, BBH at $\gg$ chance) and English is in the pre-training target of every family. The headline \emph{representation-vs-decoding} claim (RQ1--RQ4, \S\ref{sec:results}) is based on probe-vs-EN comparisons \emph{within model on English prompts} and is therefore unaffected by Tier~B/C status. The NL/EN contrast (RQ5) is where the language-proficiency confound can in principle bite: NL generations from a Tier~B/C model on a language the model was not pretrained on may be confounded with non-capability. \textbf{Three controls hold the cross-lingual claim in place even so.} (a) \emph{Positive control}: Tiny-Aya-Global \citep{ustun2024aya23, aryabumi2024aya101} explicitly targets all 8 of our non-CJK non-fallback languages and produces $NL \approx EN$ (gap $0.004$), the smallest in the sweep; if NL$<$EN were just non-capability, Aya should be the \emph{most} biased, not the least. (b) \emph{Negative control}: the Mesopotamian fallback poses NL as English with an explicit ``(Sumerian-Akkadian)'' parenthetical, bounding the within-English paraphrase-engineering component of any cross-mode signal at $|\overline{\Delta}|_{\text{Mesop}} \approx 0.04$, well below the canon-language extremes ($\geq 0.20$). (c) \emph{Structural test}: the within$-$cross paraphrase-correctness correlation gap (\S\ref{sec:independence}) is robust to language-proficiency by construction: a Tier~B model that cannot read Finnish has all five Finnish paraphrases fail \emph{together} on most cells, which \emph{maximizes} within-mode correlation (pushing it toward $1$), not lowers it; the observed asymmetry within $=0.57$ vs cross $=0.29$ is the structural signature of language-conditioned readout, not of non-capability.

\textbf{Tokenizer fertility on the per-cell delta.} Pooled across the $n=120$ (model, language) pairs in our 12-tokenizer $\times$ 10-language grid, per-(model, language) tokenizer fertility (tokens per character on $100$ Belebele \citep{bandarkar2024belebele} FLORES-200 \citep{costajussa2022flores} passages) does not correlate with the per-(model, culture) EN$-$NL delta: Pearson $r=-0.004$ ($p=0.97$); Spearman $\rho=+0.025$ ($p=0.78$). Per-culture correlations are uniformly small ($|\rho|<0.5$, $p>0.1$) with one exception (Japanese, $\rho=-0.80$, $p=0.002$), which goes in the \emph{opposite} direction to the language-proficiency hypothesis: models that tokenize Japanese more efficiently exhibit a \emph{stronger} English advantage on Japanese cultural questions, not weaker.

\section{NL vs EN aggregate, per-model wins, and NL leaderboard}\label{app:h6e6-extra}

\begin{figure*}[!t]
    \centering
    \includegraphics[width=\linewidth]{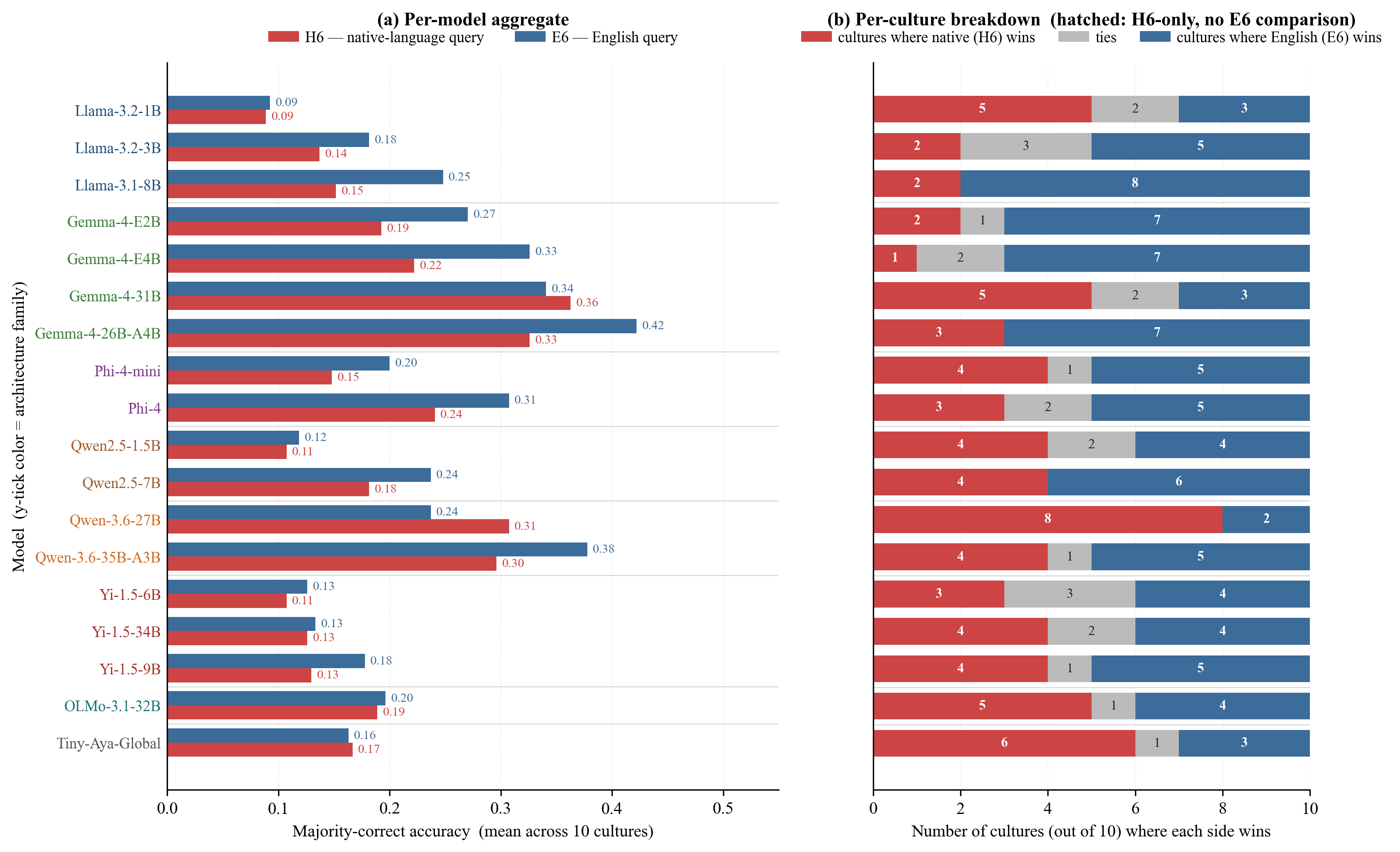}
    \caption{Cross-lingual querying: NL (target-language) vs EN (English). (a) Per-model aggregate accuracy across all 10 cultures. (b) Per-model count of cultures in which each side wins, out of 10, for chat-template models with both modes; NL-only rows are shown as a hatched ``no EN comparison available'' bar.}
    \label{fig:e6vsh6}
\end{figure*}

Figure~\ref{fig:e6vsh6} shows the NL vs EN picture at two levels. Panel (a) is per-model aggregate accuracy across all $10$ cultures: most chat-template models with both modes prefer EN in the aggregate (English mean $0.23$, native mean $0.19$, $+0.04$ on average), with three exceptions (Gemma-4-31B, Qwen-3.6-27B, Tiny-Aya-Global) where native-language querying narrowly wins. Panel (b) re-disaggregates the same data as the count of cultures in which each side wins, out of $10$, for the chat-template models with both modes. The aggregation paradox is symptomatic: every chat-template model has \emph{some} cultures where native-language querying outperforms English, and Tiny-Aya-Global (the most multilingual model in our sweep) tops the per-culture count with $6/10$ cultures where native wins despite a near-zero aggregate mean. The NL leaderboard reads Gemma-4-31B ($0.36$), Gemma-4-26B-A4B ($0.33$), Qwen-3.6-27B ($0.31$), Llama-3.1-8B ($0.25$): the top models are not strong because they recover one culture especially well, but because they spread retrieval more evenly across the $10$ cultures than smaller models do (the top row of the heatmap in Appendix~\ref{app:percult} shows Gemma-4-26B-A4B and Gemma-4-31B with no culture below $0.26$). \emph{The leaders lose the Greek default less, not the Norse advantage more.}

\section{V4 detail: per-model permutation $p$-values}\label{app:eval-perm-detail}\label{sec:eval-permutation}

Table~\ref{tab:eval-perm} reports, per chat-template model with both modes, the observed within$-$cross independence gap, the mean$\pm$std of the permuted-null distribution (5{,}000 shuffles of the 10 paraphrase-column labels), and the one-sided $p$-value. The gap is significant at $p \leq 0.005$ on every model.

\section{Cross-lingual independence and bilingual lift per model}\label{app:independence}

Table~\ref{tab:independence} reports the within-mode and cross-mode paraphrase correlations, the within$-$cross gap, the best-single-mode majority-correct rate, the NL$\cup$EN bilingual disjunction, and the per-model lift, for every chat-template model with both modes.

\paragraph{Same-language control for the bilingual lift.} The bilingual lift of \S\ref{sec:bilingual-honest} could in principle come from simply asking ten questions instead of five. To separate language-switching from paraphrase count we use the $12$ chat-template models for which we collected a second batch of $5$ same-language English paraphrases (App.~\ref{app:prompts}) and compare the bilingual ensemble $\text{NL5} \cup \text{EN5}$ against the same-language $\text{EN10}$ ensemble. The two ensembles \emph{dissociate by metric}. Under any-of-$10$ cell recovery, $\text{NL5} \cup \text{EN5}$ beats $\text{EN10}$ by $+0.053$ absolute on every model (mean $0.50$ vs $0.45$); under the per-mode majority union ($\geq 3/5$ in NL OR $\geq 3/5$ in EN) the same direction holds with $+0.052$ ($0.28$ vs $0.23$). Conversely, under a strict majority-of-$10$ threshold ($\geq 6/10$ paraphrases correct), $\text{EN10}$ wins by $0.038$ ($0.13$ vs $0.17$): same-language doubling builds strict consensus that cross-language ensembling does not. The language-switching contribution is therefore to \emph{breadth} of recovery, not to strict consensus.

\section{Per-model probe trajectories}\label{app:probe}

\begin{figure*}[!t]
    \centering
    \includegraphics[width=0.92\linewidth]{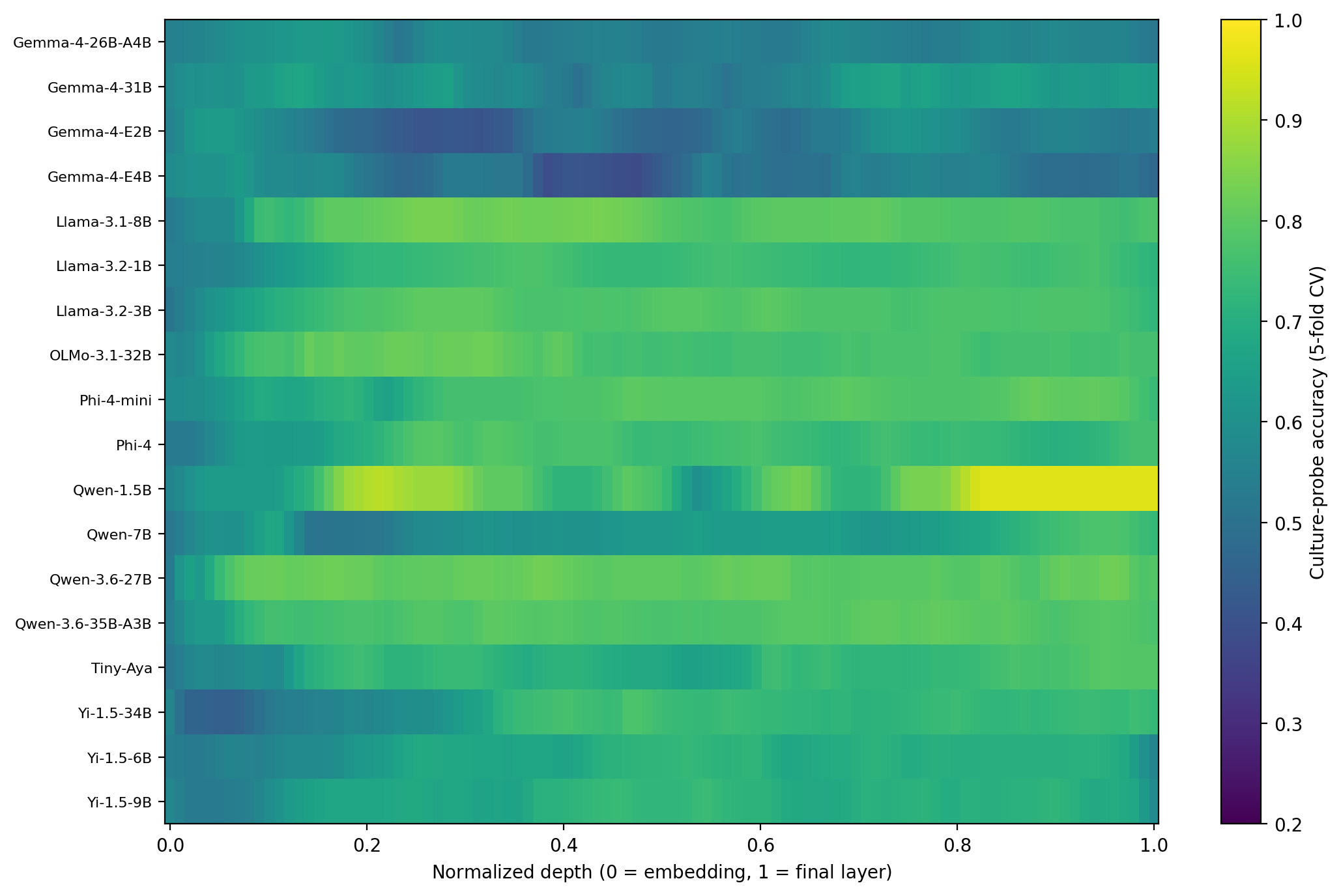}
    \caption{Layer-wise culture-probe accuracy across all models. Each row is one model; the depth axis is normalized to $[0,1]$. The probe rises early and stays high, with the peak typically before mid-network.}
    \label{fig:app_probe_layerwise}
\end{figure*}

Figure~\ref{fig:app_probe_layerwise} shows the layer-wise culture-probe accuracy across all models, with the depth axis normalized to $[0,1]$ so that models of different depths are directly comparable. Two cross-model regularities are visible: (i) for almost every model the probe accuracy starts well above the chance ceiling at the embedding ($\geq 0.5$), reaches its peak in the first half of the network, and remains roughly flat thereafter; (ii) the larger models in each family generally peak slightly later, consistent with the within-family scaling result in the main text.

\section{Per-motif Preserved share}\label{app:permotif}

\begin{figure*}[!t]
    \centering
    \includegraphics[width=0.75\linewidth]{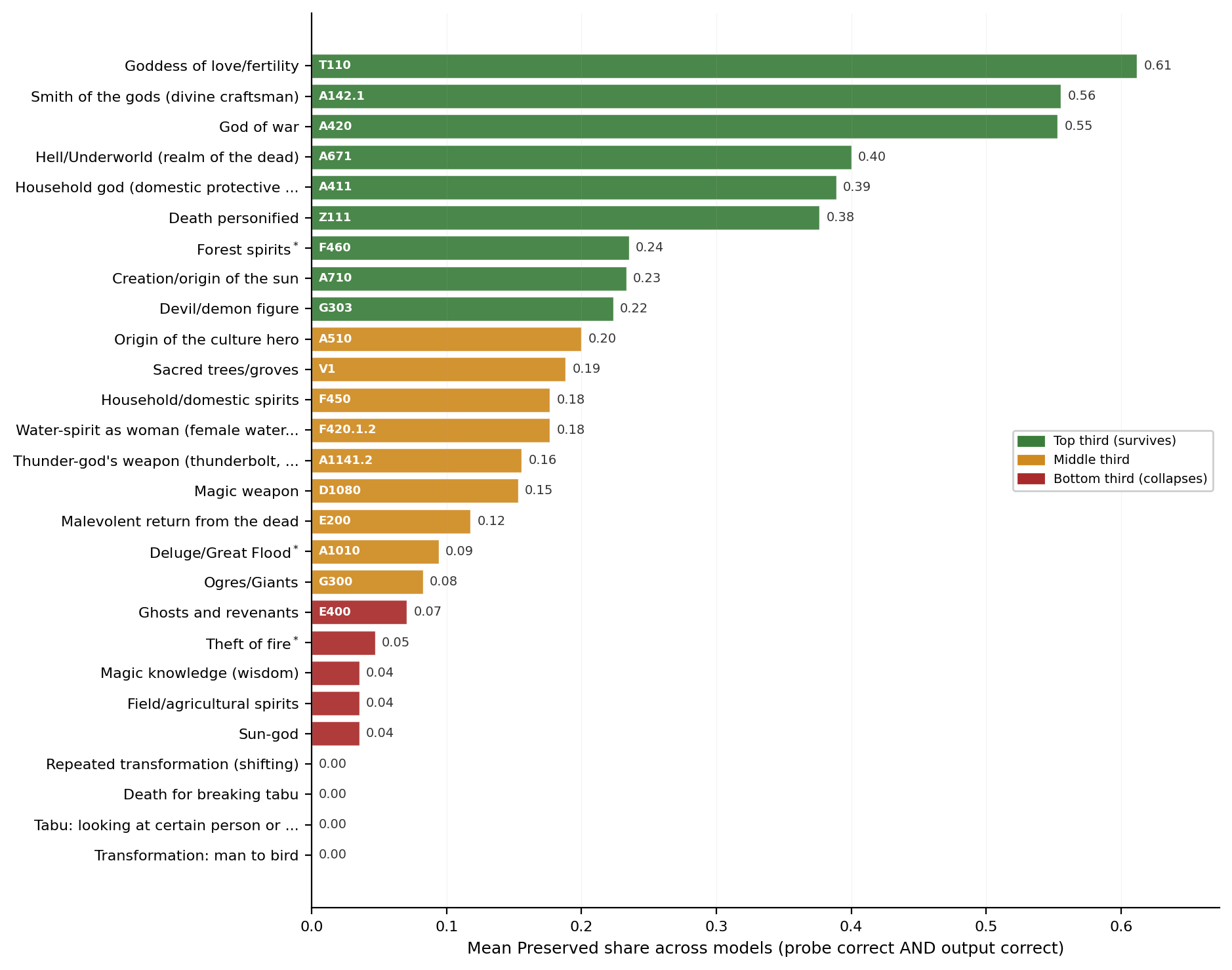}
    \caption{Per-motif Preserved share, sorted descending. Each row is one Thompson-index motif (role description on the left, motif ID inside the bar); the bar shows the mean across models of the share of (motif, culture) cells in the Preserved decomposition cell (both probe and output correct). Color tier: top third (green) survives flattening; middle third (amber); bottom third (red) collapses. Asterisked motifs (A1010, A1415, F460) are the four structural absences where not all $10$ cultures provide a canonical filler (\S\ref{sec:data}).}
    \label{fig:app_permotif}
\end{figure*}

Figure~\ref{fig:app_permotif} ranks the 27 Thompson motifs by the share of models for which the (motif, culture) pair lands in the Preserved cell of the decomposition (probe and output both correct). The top motifs (e.g. T110 ``the world tree'', A420 ``the god of the sea'') survive flattening across most models; the bottom motifs (D610 ``repeated transformations'', C310 ``tabu against looking'', D150 ``transformation: man to bird'') are nearly never recovered. The split largely tracks how strongly each motif is associated with a single named entity in popular reference works, rather than how culturally important the motif is in the source tradition.

\clearpage

\clearpage

\clearpage
\input{tab_v6_family_multi}

\clearpage
\input{tab_eval_perm}
\input{tab_independence}

\clearpage
\begin{figure*}[!t]
    \centering
    \includegraphics[width=\linewidth]{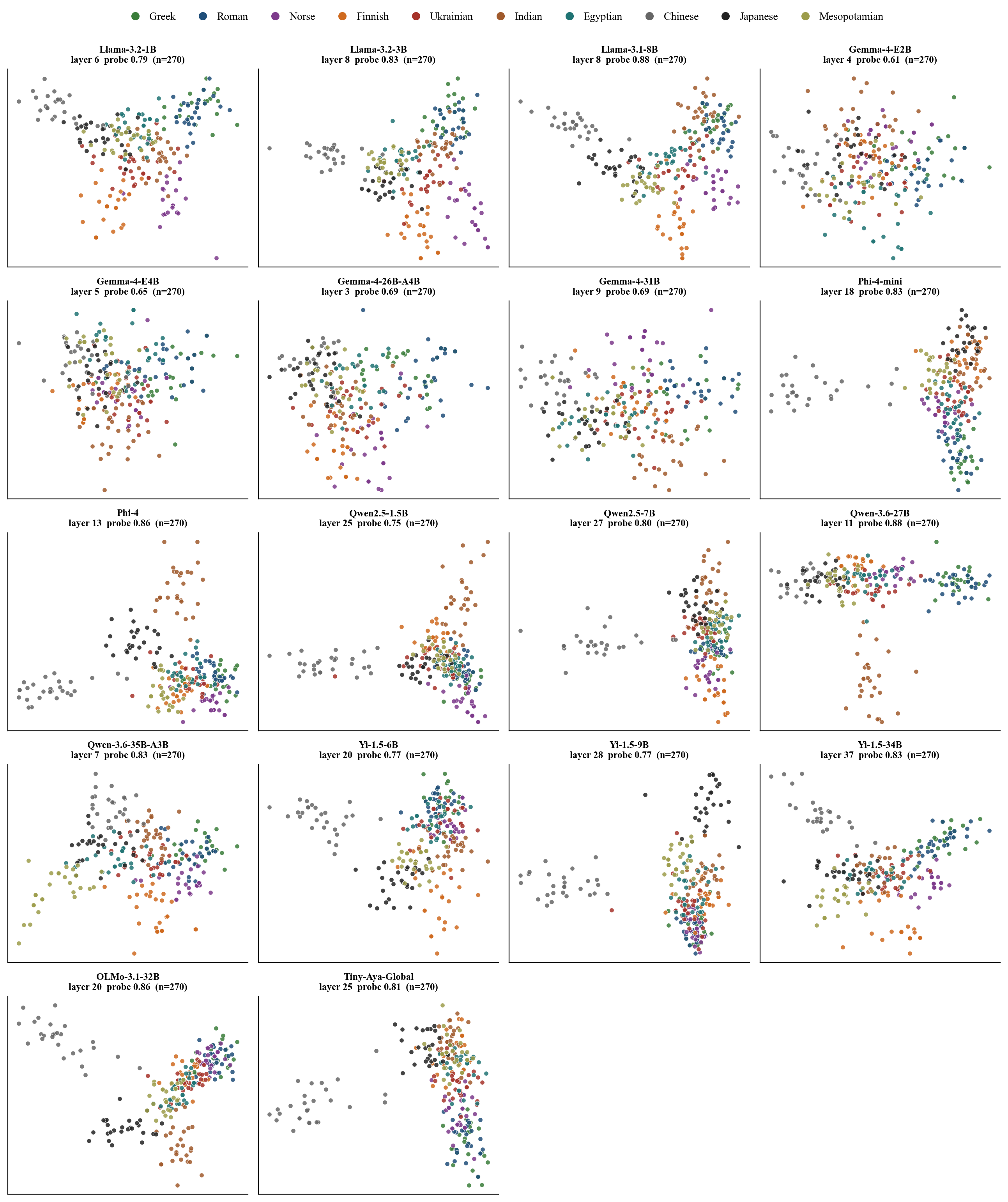}
    \caption{Probe-peak-layer latent-space projection (LDA on culture labels, $50$-dim PCA preprocessing) for every model in our $18$-model sweep. Each panel is one model; panel title gives the peak layer index, the probe peak accuracy, and the $n$ entities. Points are the $270$ (motif, culture) entities, colored by culture (legend at top). The cultural-cluster geometry is reproducible across all model families and scales: the residual stream encodes a clean, linearly-separable culture signal at mid-network in every model in our sweep.}
    \label{fig:latent-all}
\end{figure*}

\onecolumn

\section{Per-model latent-space visualization}\label{app:latent-all}

Figure~\ref{fig:latent-all} shows the peak-layer LDA projection for every model in our sweep ($n=18$). Each panel uses the same projection pipeline: $50$-dimensional PCA preprocessing of the hidden states at the model's own probe-peak layer, followed by Linear Discriminant Analysis on the $10$ culture labels to extract the top-$2$ class-separating directions. The peak-layer index ranges from $3$ (Gemma-4-26B-A4B, very early) to $37$ (Yi-1.5-34B, late mid-network), with most chat-template models peaking between layer $7$ and layer $20$. Probe peak accuracies are annotated per panel and span $0.62$ (Gemma-4-E2B) to $0.88$ (Llama-3.1-8B, Qwen-3.6-27B). Cultures form distinguishable clusters in every model: the residual stream encodes a clean, linearly-separable culture signal at mid-network in every model in our sweep. Panels are family-grouped: rows 1--2 contain Llama 3.x and Gemma 4, row 3 starts with Phi 4 then Qwen 1.x and 3.6, row 4 has Yi 1.5 and OLMo 3.1, and the last panel is Tiny-Aya-Global.

\input{tab_eval_scoring}

\clearpage

\section{Full per-model results}\label{app:fullresults}

Table~\ref{tab:app_full_results} reports the full per-model results across the 18-model sweep on the 270-entity, 10-culture substrate: residual-stream depth, probe accuracy at layer~0, probe peak accuracy and peak layer index, the label-shuffled null ceiling, and per-mode (NL, EN) majority-correct output accuracy with the best-of-(NL,~EN) aggregate. This is the source-of-truth artefact for all per-model numbers cited throughout the paper; the per-(model, culture) output breakdown is in Figure~\ref{fig:percult-app} (App.~\ref{app:percult}).

\input{tab_app_full_results}

\end{document}

%% file: tab_models.tex
\begin{table}[!h]
\centering
\small
\setlength{\tabcolsep}{4pt}
\begin{tabular}{llrr}
\toprule
Family & Model & Params (B) & Layers \\
\midrule
Llama 3.x    & Llama-3.2-1B-Instruct  &  1.2  & 16 \\
Llama 3.x    & Llama-3.2-3B-Instruct  &  3.2  & 28 \\
Llama 3.x    & Llama-3.1-8B-Instruct  &  8.0  & 32 \\
Gemma 4      & Gemma-4-E2B-it         &  2.0  & 35 \\
Gemma 4      & Gemma-4-E4B-it         &  4.0  & 42 \\
Gemma 4      & Gemma-4-31B-it         & 31.0  & 60 \\
\makecell[tl]{Gemma 4\\MoE}  & Gemma-4-26B-A4B-it     &  4.0$^a$ & 30 \\
Phi 4        & Phi-4-mini-instruct    &  3.8  & 32 \\
Phi 4        & Phi-4                  & 14.0  & 40 \\
Qwen 1.x     & Qwen2.5-1.5B-Instruct  &  1.5  & 28 \\
Qwen 1.x     & Qwen2.5-7B-Instruct    &  7.0  & 28 \\
Qwen 3.6     & Qwen3.6-27B            & 27.0  & 64 \\
\makecell[tl]{Qwen 3.6\\MoE} & Qwen3.6-35B-A3B        &  3.0$^a$ & 40 \\
Yi 1.5       & Yi-1.5-6B-Chat         &  6.0  & 32 \\
Yi 1.5       & Yi-1.5-9B-Chat         &  9.0  & 48 \\
Yi 1.5       & Yi-1.5-34B-Chat        & 34.0  & 60 \\
OLMo 3.1     & Olmo-3.1-32B-Instruct  & 32.0  & 64 \\
Aya          & tiny-aya-global        &  3.0  & 36 \\
\bottomrule
\end{tabular}
\caption{18 open-source LLMs in the sweep. $^a$Active params per token for MoE (total $26$B / $35$B); layers exclude embedding.}
\label{tab:models}
\end{table}

%% file: tab_headline.tex
\begin{table}[t]
\centering\footnotesize
\setlength{\tabcolsep}{2.5pt}
\renewcommand{\arraystretch}{0.95}
\begin{tabular}{@{}lcccccc@{}}
\toprule
\textbf{Model} & \textbf{Par.} & \textbf{Pr.} & \textbf{MCQ} & \textbf{EN} & \textbf{Best} & \textbf{Gap} \\
\midrule
Llama-3.2-1B & 1.2 & .79 & .20$^\dagger$ & .09 & .09 & .70 \\
Llama-3.2-3B & 3.2 & .83 & .60 & .18 & .18 & .65 \\
Llama-3.1-8B & 8.0 & .88 & .74 & .25 & .25 & .63 \\
\midrule
Gemma-4-E2B & 2.0 & .61 & .41 & .27 & .27 & .34 \\
Gemma-4-E4B & 4.0 & .65 & .73 & .33 & .33 & .32 \\
Gemma-4-26B-A4B & 4.0$^*$ & .69 & .96 & .42 & .42 & .27 \\
Gemma-4-31B & 33.0 & .69 & .95 & .43 & .43 & \textbf{.26} \\
\midrule
Phi-4-mini & 3.8 & .83 & .52 & .20 & .20 & .63 \\
Phi-4 & 14.0 & .86 & .86 & .31 & .31 & .55 \\
\midrule
Qwen2.5-1.5B & 1.5 & .75 & .40 & .12 & .12 & .63 \\
Qwen2.5-7B & 7.0 & .80 & .67 & .24 & .24 & .56 \\
\midrule
Qwen-3.6-35B-A3B & 3.0$^*$ & .83 & .93 & .38 & .38 & .45 \\
Qwen-3.6-27B & 27.0 & .88 & .95 & .35 & .35 & .53 \\
\midrule
Yi-1.5-6B & 6.0 & .77 & .42 & .13 & .13 & .64 \\
Yi-1.5-9B & 9.0 & .77 & .70 & .18 & .18 & .59 \\
Yi-1.5-34B & 34.0 & .83 & .83 & .26 & .26 & .57 \\
\midrule
OLMo-3.1-32B & 32.0 & .86 & .76 & .32 & .32 & .54 \\
\midrule
Tiny-Aya-Global & 3.0 & .81 & .45 & .16 & .17 & .64 \\
\midrule
\textbf{Mean} (18) & & \textbf{.79} & \textbf{.67} & \textbf{.26} & .26 & .53 \\
\bottomrule
\end{tabular}
\caption{Decoding flattening and the output-format control across the 18-model sweep. \textbf{Pr.} = linear-probe peak (10-way, chance $.10$, surface baseline $.60$). \textbf{MCQ} = selecting the gold entity among the same motif's $10$ parallel fillers (same chance as the probe), scored by restricted first-token log-probability over $5$ paraphrases $\times$ $3$ option orders (\S\ref{sec:mcq}). \textbf{EN} = English free-generation majority-correct; \textbf{Best} = best-of-(NL, EN); \textbf{Gap} = Pr.\ $-$ Best, bold marks the smallest. The chain Pr.\ $\rightarrow$ MCQ $\rightarrow$ EN localizes the loss at generation: selection recovers most of what the probe reads, generation does not. $^*$Active params for MoE. $^\dagger$Format-compliance floor ($4\%$ letters).}
\label{tab:headline}\label{tab:mcq}
\end{table}

%% file: tab_dataset_full.tex
\begingroup\footnotesize\setlength{\tabcolsep}{4pt}\renewcommand{\arraystretch}{1.06}
\begin{longtable}{@{}l p{3.0cm} l p{2.8cm} p{7.0cm}@{}}
\caption{Full per-entity dataset listing for the 270-entity, 10-culture Thompson-motif substrate. Native-script forms are in the released CSV/JSON artefact. Superscripts mark contested or structural-absence rows (footnoted after the table). \label{tab:dataset-full}} \\
\toprule
\textbf{Motif} & \textbf{Role} & \textbf{Culture} & \textbf{Name} & \textbf{Source} \\
\midrule \endfirsthead
\multicolumn{5}{l}{\small\itshape (continued)} \\ \toprule
\textbf{Motif} & \textbf{Role} & \textbf{Culture} & \textbf{Name} & \textbf{Source} \\ \midrule \endhead
\midrule \multicolumn{5}{r}{\small\itshape continued on next page} \\ \endfoot
\bottomrule \endlastfoot
\textsc{A1010} & Deluge/Great Flood & Greek & Deucalion & \texttt{Hard 2004} \\
 &  & Roman & Deucalion & \texttt{Grimal 1996} \\
 &  & Norse & Bergelmir & \texttt{Lindow 2001} \\
 &  & Finnish & (structural absence)\textsuperscript{13} & \texttt{(no canonical Finnish flood-survivor narrative)} \\
 &  & Ukrainian & Noi & \texttt{Voropay 1958 *Zvychayi nashoho narodu*; Hnatyuk *Etnohrafichnyi zbirnyk* (Christianized flood-hero in Ukrainian folk tradition)} \\
 &  & Indian & Matsya & \texttt{Matsya Purana 1-2; Bhagavata Purana 8.24; Mahabharata 3.187 (flood narrative)} \\
 &  & Egyptian & Hathor (Sekhmet) & \texttt{*The Book of the Heavenly Cow* (New Kingdom funerary text)} \\
 &  & Chinese & Yu the Great & \texttt{*Shujing* (Book of Documents, Yugong chapter); *Mengzi*; *Shanhaijing*} \\
 &  & Japanese & (structural absence)\textsuperscript{6} & \texttt{(no canonical Japanese deluge narrative)} \\
 &  & Mesopotamian & Atrahasis (Utnapishtim) & \texttt{*Atrahasis Epic*; *Epic of Gilgamesh* tablet XI} \\
\addlinespace[2pt]
\textsc{A1141.2} & Thunder-god's weapon (thunderbolt, hammer, axe) & Greek & Zeus & \texttt{Hard 2004} \\
 &  & Roman & Jupiter & \texttt{Grimal 1996} \\
 &  & Norse & Thor & \texttt{Lindow 2001} \\
 &  & Finnish & Ukko & \texttt{Pentik\"ainen 1999} \\
 &  & Ukrainian & Perun & \texttt{Br\"uckner 1918/1985 *Mitologia s\l{}owia\'nska*; Gieysztor 2006 *Mitologia S\l{}owian* (Warsaw UP)} \\
 &  & Indian & Vajra & \texttt{Rig Veda 1.32 (Indra slays Vritra with vajra); Mahabharata} \\
 &  & Egyptian & Set & \texttt{Pyramid Texts; Coffin Texts; Book of the Dead (storm and disorder god)} \\
 &  & Chinese & Lei Gong (Duke of Thunder) & \texttt{*Shanhaijing* (Classic of Mountains and Seas, c. 4 c. BCE); folk iconography} \\
 &  & Japanese & Takemikazuchi & \texttt{*Kojiki* (712 CE) Kamiumi episode; *Nihon Shoki* (720 CE)} \\
 &  & Mesopotamian & Adad & \texttt{*Enuma Elish*; storm-god hymns; royal inscriptions throughout Mesopotamian history} \\
\addlinespace[2pt]
\textsc{A1415} & Theft of fire & Greek & Prometheus & \texttt{Hard 2004} \\
 &  & Roman & Prometheus & \texttt{Grimal 1996} \\
 &  & Norse & Loki & \texttt{Lindow 2001} \\
 &  & Finnish & V\"ain\"am\"oinen & \texttt{Kalevala runo 47} \\
 &  & Ukrainian & Svarog\textsuperscript{7} & \texttt{Gieysztor 2006 *Mitologia S\l{}owian* (Warsaw UP)} \\
 &  & Indian & M\=atari\'svan & \texttt{Rig Veda 1.31, 3.5, 3.9, 10.46 (M\=atari\'svan brings fire from heaven to the Bhrigus)} \\
 &  & Egyptian & Prometheus-equivalent\textsuperscript{8} & \texttt{(no Egyptian fire-thief narrative)} \\
 &  & Chinese & Suiren-shi & \texttt{*Hanshu* (Han history); *Han Feizi*; folk tradition} \\
 &  & Japanese & Kagutsuchi\textsuperscript{9} & \texttt{*Kojiki*; *Nihon Shoki*} \\
 &  & Mesopotamian & (structural absence)\textsuperscript{10} & \texttt{(no Mesopotamian theft-of-fire narrative)} \\
\addlinespace[2pt]
\textsc{A142.1} & Smith of the gods (divine craftsman) & Greek & Hephaestus & \texttt{Hard 2004} \\
 &  & Roman & Vulcan & \texttt{Grimal 1996} \\
 &  & Norse & Brokkr & \texttt{Lindow 2001} \\
 &  & Finnish & Ilmarinen & \texttt{Pentik\"ainen 1999} \\
 &  & Ukrainian & Svarog & \texttt{Gieysztor 2006 *Mitologia S\l{}owian* (Warsaw UP)} \\
 &  & Indian & Tvashtar & \texttt{Rig Veda 1.32, 10.81 (forges Indra's vajra; 'all-fashioner')} \\
 &  & Egyptian & Ptah & \texttt{Memphite Theology (Shabaka Stone, c. 700 BCE); Pyramid Texts} \\
 &  & Chinese & Lu Ban & \texttt{*L\"ushi Chunqiu*; *Mozi*; folk patron of craftsmen since Spring \& Autumn period} \\
 &  & Japanese & Amatsumara & \texttt{*Kojiki* (712 CE)} \\
 &  & Mesopotamian & Ea (Enki) & \texttt{Sumerian *Enki and the World Order*; *Atrahasis*; *Enuma Elish*} \\
\addlinespace[2pt]
\textsc{A220} & Sun-god & Greek & Apollo & \texttt{Hard 2004} \\
 &  & Roman & Apollo & \texttt{Grimal 1996} \\
 &  & Norse & Baldr & \texttt{Lindow 2001} \\
 &  & Finnish & P\"aiv\"a & \texttt{Pentik\"ainen 1999} \\
 &  & Ukrainian & Khors\textsuperscript{1} & \texttt{Gieysztor 2006 *Mitologia S\l{}owian* (Warsaw UP)} \\
 &  & Indian & Surya & \texttt{Rig Veda 1.50 (Surya hymn); Surya Upanishad; Mahabharata} \\
 &  & Egyptian & Ra & \texttt{Pyramid Texts; Litany of Ra; Book of the Dead} \\
 &  & Chinese & Xihe & \texttt{*Shanhaijing*; *Huainanzi* (2 c. BCE)} \\
 &  & Japanese & Amaterasu & \texttt{*Kojiki*; *Nihon Shoki*; *Engishiki*} \\
 &  & Mesopotamian & Utu (Shamash) & \texttt{Sumerian and Akkadian sun-god hymns; *Code of Hammurabi* prologue} \\
\addlinespace[2pt]
\textsc{A411} & Household god (domestic protective deity) & Greek & Hestia & \texttt{Hard 2004} \\
 &  & Roman & Lares & \texttt{Grimal 1996} \\
 &  & Norse & Tomte & \texttt{Lindow 2001} \\
 &  & Finnish & Tonttu & \texttt{Pentik\"ainen 1999} \\
 &  & Ukrainian & Domovyk & \texttt{Voropay 1958 *Zvychayi nashoho narodu* (Customs of our people, Ukrainian diaspora ethnography); Br\"uckner 1918/1985} \\
 &  & Indian & V\=asto\d{s}pati & \texttt{Rig Veda 7.54-55 (V\=asto\d{s}pati hymn --- protector of dwellings)} \\
 &  & Egyptian & Bes & \texttt{Egyptian household amulets, popular religion (Middle Kingdom onwards)} \\
 &  & Chinese & Zao Jun (Stove God) & \texttt{*Liji* (Book of Rites, 1 c. BCE); folk household tradition continuous since Han dynasty} \\
 &  & Japanese & Kamado-no-Kami & \texttt{*Engishiki* (10 c. CE); folk tradition} \\
 &  & Mesopotamian & Lamassu & \texttt{Mesopotamian protective-amulet inscriptions; royal palace gateway figures (Assyrian)} \\
\addlinespace[2pt]
\textsc{A420} & God of war & Greek & Ares & \texttt{Hard 2004} \\
 &  & Roman & Mars & \texttt{Grimal 1996} \\
 &  & Norse & Tyr & \texttt{Lindow 2001} \\
 &  & Finnish & Turisas & \texttt{Finnish mythology} \\
 &  & Ukrainian & Perun & \texttt{Br\"uckner 1918/1985; Gieysztor 2006 *Mitologia S\l{}owian* (Warsaw UP)} \\
 &  & Indian & Skanda & \texttt{Mahabharata Vana Parva (Skanda's birth); Skanda Purana} \\
 &  & Egyptian & Montu & \texttt{Pyramid Texts; royal Theban war-cult (Middle Kingdom)} \\
 &  & Chinese & Guandi (Guan Yu) & \texttt{*Sanguozhi* (3 c. CE historical); *Sanguo Yanyi* (14 c. novel); state cult deification (Ming-Qing)} \\
 &  & Japanese & Hachiman & \texttt{*Shoku Nihongi* (797 CE); Iwashimizu Hachiman shrine cult} \\
 &  & Mesopotamian & Ninurta & \texttt{*Anzu Epic*; *Lugal-e* (Sumerian); royal inscriptions} \\
\addlinespace[2pt]
\textsc{A510} & Origin of the culture hero & Greek & Heracles & \texttt{Hard 2004} \\
 &  & Roman & Romulus & \texttt{Grimal 1996} \\
 &  & Norse & Sigurd & \texttt{Lindow 2001} \\
 &  & Finnish & V\"ain\"am\"oinen & \texttt{Pentik\"ainen 1999} \\
 &  & Ukrainian & Kyi & \texttt{Primary Chronicle s.a. ~6th c. (Cross \& Sherbowitz-Wetzor 1953); Plokhy 2017} \\
 &  & Indian & Manu & \texttt{Manusmriti; Matsya Purana 1-2; Mahabharata 3.187 (flood-survivor --- progenitor of humanity)} \\
 &  & Egyptian & Osiris & \texttt{Pyramid Texts (Heliopolitan Osiris cycle); Plutarch *De Iside et Osiride* (1c. CE)} \\
 &  & Chinese & Fuxi & \texttt{*Shiji* (Sima Qian, 1 c. BCE); *Yijing* attribution; *Huainanzi*} \\
 &  & Japanese & Ninigi-no-Mikoto & \texttt{*Kojiki*; *Nihon Shoki*} \\
 &  & Mesopotamian & Gilgamesh & \texttt{*Epic of Gilgamesh* (Old Babylonian + Standard Babylonian versions); Sumerian Bilgames poems} \\
\addlinespace[2pt]
\textsc{A671} & Hell/Underworld (realm of the dead) & Greek & Hades & \texttt{Hard 2004} \\
 &  & Roman & Pluto & \texttt{Grimal 1996} \\
 &  & Norse & Hel & \texttt{Lindow 2001} \\
 &  & Finnish & Tuoni & \texttt{Pentik\"ainen 1999 (Tuoni, lord of Tuonela)} \\
 &  & Ukrainian & Veles & \texttt{Gieysztor 2006 *Mitologia S\l{}owian* (Warsaw UP); Strzelczyk 1998 *Mity, podania i wierzenia dawnych S\l{}owian* (Pozna\'n) (Veles as ruler of Nav)} \\
 &  & Indian & Yama & \texttt{Rig Veda 10.14 (Yama-Yami hymn); Garuda Purana (preta-kalpa, narakas)} \\
 &  & Egyptian & Osiris & \texttt{Pyramid Texts; Coffin Texts; Book of the Dead} \\
 &  & Chinese & Yanluo Wang (Yama-King) & \texttt{Buddhist sutras (transmission of Yama as Yanluo); Daoist *Diyu* texts; folk *Yu Li Bao Chao*} \\
 &  & Japanese & Izanami & \texttt{*Kojiki* book 1 (Izanami becomes ruler of Yomi after death); *Nihon Shoki*} \\
 &  & Mesopotamian & Ereshkigal & \texttt{*Descent of Inanna*; *Nergal and Ereshkigal*; *Epic of Gilgamesh* tablet XII} \\
\addlinespace[2pt]
\textsc{A710} & Creation/origin of the sun & Greek & Helios & \texttt{Hard 2004} \\
 &  & Roman & Sol & \texttt{Grimal 1996} \\
 &  & Norse & S\'ol & \texttt{Lindow 2001} \\
 &  & Finnish & P\"aiv\"at\"ar & \texttt{Pentik\"ainen 1999} \\
 &  & Ukrainian & Dazhboh & \texttt{Gieysztor 2006 *Mitologia S\l{}owian* (Warsaw UP); Strzelczyk 1998} \\
 &  & Indian & Vivasvant & \texttt{Rig Veda 10.17.1-2 (origin of Vivasvant); 1.115 (his radiant chariot)} \\
 &  & Egyptian & Atum & \texttt{Heliopolitan creation account; Pyramid Texts; Coffin Texts; Book of the Dead} \\
 &  & Chinese & Pangu & \texttt{*Sanwu Liji* (3 c. CE, Xu Zheng); *Wuyun Linian Ji*} \\
 &  & Japanese & Amaterasu (cave myth) & \texttt{*Kojiki*; *Nihon Shoki*} \\
 &  & Mesopotamian & Utu's daily journey & \texttt{Sumerian Utu hymns; Akkadian Shamash hymns} \\
\addlinespace[2pt]
\textsc{C310} & Tabu: looking at certain person or thing & Greek & Orpheus & \texttt{Hard 2004} \\
 &  & Roman & Orpheus & \texttt{Grimal 1996} \\
 &  & Norse & Baldr & \texttt{Lindow 2001} \\
 &  & Finnish & Lemmink\"ainen & \texttt{Kalevala} \\
 &  & Ukrainian & Kupala\textsuperscript{12} & \texttt{Strzelczyk 1998 *Mity, podania i wierzenia dawnych S\l{}owian*; Voropay 1958} \\
 &  & Indian & Ahalya & \texttt{Ramayana Bala Kanda 47-48 (Ahalya's curse and stoning)} \\
 &  & Egyptian & Hathor (mirror-tabu) & \texttt{Egyptian wisdom literature; *The Tale of the Two Brothers* (Papyrus d'Orbiney, c. 1185 BCE)} \\
 &  & Chinese & Niulang and Zhin\"u & \texttt{Han dynasty *Gushi Shijiu Shou* (Nineteen Old Poems); Tang/Song *Qixi* festival traditions} \\
 &  & Japanese & Izanagi at Yomi & \texttt{*Kojiki* book 1; *Nihon Shoki*} \\
 &  & Mesopotamian & Adapa & \texttt{*Adapa Epic* (Akkadian, c. 14 c. BCE)} \\
\addlinespace[2pt]
\textsc{C920} & Death for breaking tabu & Greek & Semele & \texttt{Hard 2004} \\
 &  & Roman & Semele & \texttt{Grimal 1996} \\
 &  & Norse & Baldr & \texttt{Lindow 2001} \\
 &  & Finnish & Kullervo & \texttt{Kalevala runo 31-36} \\
 &  & Ukrainian & Marena & \texttt{Br\"uckner 1918/1985 *Mitologia s\l{}owia\'nska*; Strzelczyk 1998} \\
 &  & Indian & Daksha & \texttt{Linga Purana; Vayu Purana 30; Bhagavata Purana 4.2-7} \\
 &  & Egyptian & Bata & \texttt{*The Tale of the Two Brothers* (Papyrus d'Orbiney, c. 1185 BCE)} \\
 &  & Chinese & Houyi & \texttt{*Huainanzi* (2 c. BCE); *Shanhaijing*} \\
 &  & Japanese & Izanami & \texttt{*Kojiki*; *Nihon Shoki*} \\
 &  & Mesopotamian & Etana & \texttt{*Etana Epic* (Old Babylonian + Middle Assyrian)} \\
\addlinespace[2pt]
\textsc{D1080} & Magic weapon & Greek & Aegis & \texttt{Hard 2004} \\
 &  & Roman & Ancile & \texttt{Grimal 1996} \\
 &  & Norse & Gram & \texttt{Lindow 2001} \\
 &  & Finnish & Sampo & \texttt{Pentik\"ainen 1999} \\
 &  & Ukrainian & Mech-kladenets\textsuperscript{5} & \texttt{Strzelczyk 1998 *Mity, podania i wierzenia dawnych S\l{}owian* (Pozna\'n); Hnatyuk *Etnohrafichnyi zbirnyk*} \\
 &  & Indian & Sudarshana Chakra & \texttt{Mahabharata; Bhagavata Purana; Vishnu Purana} \\
 &  & Egyptian & Was scepter & \texttt{Pyramid Texts; royal regalia from earliest dynasties} \\
 &  & Chinese & Ruyi Jingu Bang & \texttt{*Xiyou Ji* (Journey to the West)} \\
 &  & Japanese & Kusanagi-no-Tsurugi & \texttt{*Kojiki*; *Nihon Shoki*; one of three Imperial Regalia} \\
 &  & Mesopotamian & Sharur & \texttt{*Lugal-e* (Sumerian); *Anzu Epic*} \\
\addlinespace[2pt]
\textsc{D150} & Transformation: man to bird & Greek & Ceyx & \texttt{Ovid Met. 11} \\
 &  & Roman & Picus & \texttt{Grimal 1996} \\
 &  & Norse & Odin & \texttt{Lindow 2001} \\
 &  & Finnish & Lemmink\"ainen & \texttt{Kalevala runo 12} \\
 &  & Ukrainian & Finist & \texttt{Voropay 1958 *Zvychayi nashoho narodu*; Strzelczyk 1998} \\
 &  & Indian & Garuda & \texttt{Mahabharata Adi Parva 16-34 (Garuda's birth and quest for amrita); Bhagavata Purana} \\
 &  & Egyptian & Horus & \texttt{Pyramid Texts; Coffin Texts; Book of the Dead} \\
 &  & Chinese & Jingwei & \texttt{*Shanhaijing* (Beishan Jing); *Shuyi Ji*} \\
 &  & Japanese & Yatagarasu & \texttt{*Kojiki*; *Nihon Shoki*} \\
 &  & Mesopotamian & Etana & \texttt{*Etana Epic*} \\
\addlinespace[2pt]
\textsc{D1810} & Magic knowledge (wisdom) & Greek & Athena & \texttt{Hard 2004} \\
 &  & Roman & Minerva & \texttt{Grimal 1996} \\
 &  & Norse & Odin & \texttt{Lindow 2001} \\
 &  & Finnish & V\"ain\"am\"oinen & \texttt{Pentik\"ainen 1999} \\
 &  & Ukrainian & Veles & \texttt{Gieysztor 2006 *Mitologia S\l{}owian* (Warsaw UP); \L{}uczy\'nski 2020 *Bogowie dawnych S\l{}owian*} \\
 &  & Indian & Brihaspati & \texttt{Rig Veda 4.50 (Brahmanaspati hymn); Mahabharata; Brahmanas} \\
 &  & Egyptian & Thoth & \texttt{Pyramid Texts; Book of the Dead; Hermopolitan theology} \\
 &  & Chinese & Wenchang Wang & \texttt{Daoist *Yu Li Bao Chao*; popular literacy-cult since Tang dynasty} \\
 &  & Japanese & Omoikane & \texttt{*Kojiki*; *Nihon Shoki*} \\
 &  & Mesopotamian & Ea (Enki) & \texttt{Sumerian and Akkadian wisdom literature; *Atrahasis*; *Enuma Elish*} \\
\addlinespace[2pt]
\textsc{D610} & Repeated transformation (shifting) & Greek & Proteus & \texttt{Hard 2004} \\
 &  & Roman & Vertumnus & \texttt{Grimal 1996} \\
 &  & Norse & Loki & \texttt{Lindow 2001} \\
 &  & Finnish & Joukahainen & \texttt{Kalevala runo 3} \\
 &  & Ukrainian & Vodianyk & \texttt{Voropay 1958 *Zvychayi nashoho narodu*; Br\"uckner 1918/1985} \\
 &  & Indian & Vishnu & \texttt{Bhagavata Purana 1.3 (ten avatars listed); Garuda Purana} \\
 &  & Egyptian & Khepri & \texttt{Pyramid Texts; Book of the Dead} \\
 &  & Chinese & Sun Wukong & \texttt{*Xiyou Ji* (Journey to the West, 16 c. CE, Wu Cheng'en)} \\
 &  & Japanese & Kitsune & \texttt{*Konjaku Monogatari* (12 c.); *Kojiki* references; folk *kitsune-tsuki*} \\
 &  & Mesopotamian & Inanna (descent and ascent) & \texttt{*Descent of Inanna* (Sumerian); *Descent of Ishtar* (Akkadian)} \\
\addlinespace[2pt]
\textsc{E200} & Malevolent return from the dead & Greek & Lamia & \texttt{Hard 2004 *Routledge Handbook of Greek Mythology*, pp. 116-117 (Lamia myth); cf. Aristophanes *Wasps* 1035 and scholia on Aristophanes *Frogs* 285} \\
 &  & Roman & Lemures & \texttt{Grimal 1996} \\
 &  & Norse & Draugr & \texttt{Lindow 2001} \\
 &  & Finnish & Kalma & \texttt{Pentik\"ainen 1999} \\
 &  & Ukrainian & Upyr & \texttt{Br\"uckner 1918/1985; Voropay 1958 *Zvychayi nashoho narodu*} \\
 &  & Indian & Vetala & \texttt{Vetala Panchavimshati (Twenty-five tales of Vetala); Kathasaritsagara} \\
 &  & Egyptian & Apophis (Apep) & \texttt{Pyramid Texts; Book of the Dead; Coffin Texts; *Book of Apophis* (Bremner-Rhind Papyrus)} \\
 &  & Chinese & Jiangshi & \texttt{Qing dynasty *Zibuyu* (Yuan Mei, 18 c.); folk hopping-vampire tradition} \\
 &  & Japanese & Onry\=o & \texttt{*Konjaku Monogatari*; Heian-period y\=urei tradition; Edo-period *Yotsuya Kaidan*} \\
 &  & Mesopotamian & Lamashtu & \texttt{Akkadian incantation series *Lamashtu*} \\
\addlinespace[2pt]
\textsc{E400} & Ghosts and revenants & Greek & Eidolon\textsuperscript{4} & \texttt{Greek belief} \\
 &  & Roman & Manes & \texttt{Grimal 1996} \\
 &  & Norse & Haugbui & \texttt{Lindow 2001} \\
 &  & Finnish & Kalma & \texttt{Finnish belief} \\
 &  & Ukrainian & Mertviaky & \texttt{Voropay 1958 *Zvychayi nashoho narodu*; Br\"uckner 1918/1985} \\
 &  & Indian & Preta & \texttt{Garuda Purana (preta-kalpa); Mahabharata Anushasana Parva; Manusmriti 3.226-249} \\
 &  & Egyptian & Akh & \texttt{Pyramid Texts; Coffin Texts} \\
 &  & Chinese & Gui & \texttt{*Shijing*; *Liji*; *Yu Li Bao Chao*} \\
 &  & Japanese & Y\=urei & \texttt{*Genji Monogatari* (11 c.); Edo-period kaidan} \\
 &  & Mesopotamian & Etemmu & \texttt{Akkadian funerary incantations; *Epic of Gilgamesh* tablet XII} \\
\addlinespace[2pt]
\textsc{F420.1.2} & Water-spirit as woman (female water entity) & Greek & Naiad & \texttt{Hard 2004} \\
 &  & Roman & Nympha & \texttt{Grimal 1996} \\
 &  & Norse & Nixie & \texttt{Lindow 2001} \\
 &  & Finnish & N\"akki & \texttt{Pentik\"ainen 1999} \\
 &  & Ukrainian & Rusalka & \texttt{Voropay 1958 *Zvychayi nashoho narodu*; Strzelczyk 1998} \\
 &  & Indian & Apsaras & \texttt{Rig Veda 10.95 (Urvashi-Pururavas); Mahabharata; Ramayana} \\
 &  & Egyptian & Anuket & \texttt{Pyramid Texts; Aswan / First-Cataract cult} \\
 &  & Chinese & Long N\"u (Dragon Princess) & \texttt{*Liu Yi Zhuan* (Tang chuanqi, 9 c.); Tang Buddhist Longwang traditions} \\
 &  & Japanese & Mizuchi & \texttt{*Nihon Shoki* (720 CE); folk tradition} \\
 &  & Mesopotamian & Tiamat & \texttt{*Enuma Elish*} \\
\addlinespace[2pt]
\textsc{F450} & Household/domestic spirits & Greek & Agathos Daimon & \texttt{Greek belief} \\
 &  & Roman & Penates & \texttt{Grimal 1996} \\
 &  & Norse & Tomte & \texttt{Lindow 2001} \\
 &  & Finnish & Tonttu & \texttt{Finnish belief} \\
 &  & Ukrainian & Domovyk & \texttt{Voropay 1958 *Zvychayi nashoho narodu*; Br\"uckner 1918/1985} \\
 &  & Indian & Grihya devatas & \texttt{Grihya Sutras (Ashvalayana, Apastamba); Manusmriti 3.80-90 (panchayajna ritual)} \\
 &  & Egyptian & Bes & \texttt{Egyptian household amulet tradition} \\
 &  & Chinese & Zao Jun (Stove God) & \texttt{*Liji*; folk household tradition} \\
 &  & Japanese & Zashiki-warashi & \texttt{T\=ono region folk tradition; Yanagita Kunio *T\=ono Monogatari* (1910)} \\
 &  & Mesopotamian & Lamassu & \texttt{Akkadian incantations; royal-palace gateway figures} \\
\addlinespace[2pt]
\textsc{F460} & Forest spirits & Greek & Dryad & \texttt{Hard 2004} \\
 &  & Roman & Silvanus & \texttt{Grimal 1996} \\
 &  & Norse & Skogsfru & \texttt{Scandinavian belief} \\
 &  & Finnish & Mets\"anhaltija & \texttt{Finnish belief} \\
 &  & Ukrainian & Lisovyk & \texttt{Voropay 1958 *Zvychayi nashoho narodu*; Br\"uckner 1918/1985} \\
 &  & Indian & Yaksha & \texttt{Mahabharata Vana Parva 313 (Yaksha-Prashna); Atharvaveda; Buddhist Jatakas} \\
 &  & Egyptian & (structural absence)\textsuperscript{11} & \texttt{(no canonical Egyptian forest-spirit)} \\
 &  & Chinese & Shanshen (Mountain God) & \texttt{*Shanhaijing*; folk tradition; Daoist mountain cults} \\
 &  & Japanese & Kodama & \texttt{*Wakan Sansai Zue* (1712); folk tradition} \\
 &  & Mesopotamian & Humbaba (Huwawa) & \texttt{*Epic of Gilgamesh* tablets II-V; Sumerian *Bilgames and Huwawa*} \\
\addlinespace[2pt]
\textsc{F470} & Field/agricultural spirits & Greek & Demeter & \texttt{Hard 2004} \\
 &  & Roman & Ceres & \texttt{Grimal 1996} \\
 &  & Norse & Freyr & \texttt{Lindow 2001} \\
 &  & Finnish & Pellon-Pekko & \texttt{Finnish belief} \\
 &  & Ukrainian & Polevyk & \texttt{Voropay 1958 *Zvychayi nashoho narodu*; Br\"uckner 1918/1985} \\
 &  & Indian & Bh\=umi Devi & \texttt{Atharvaveda 12.1 (P\d{r}thiv\={\i} S\=ukta --- hymn to earth/soil); Vishnu Purana} \\
 &  & Egyptian & Min & \texttt{Pyramid Texts; Min temple at Coptos (predynastic origin)} \\
 &  & Chinese & Tudi Gong (Earth God) & \texttt{*Shijing* references to *she*; folk tradition continuous since pre-Han period} \\
 &  & Japanese & Inari & \texttt{*Engishiki* (10 c.); Fushimi Inari shrine cult since 711 CE} \\
 &  & Mesopotamian & Ninhursag & \texttt{Sumerian *Enki and Ninhursag*; *Atrahasis*} \\
\addlinespace[2pt]
\textsc{G300} & Ogres/Giants & Greek & Cyclops & \texttt{Hard 2004, 269-270} \\
 &  & Roman & Cacus & \texttt{Grimal 1996} \\
 &  & Norse & J\"otunn & \texttt{Lindow 2001} \\
 &  & Finnish & Hiisi & \texttt{Pentik\"ainen 1999} \\
 &  & Ukrainian & Chudo-Yudo & \texttt{Hnatyuk *Etnohrafichnyi zbirnyk* (Lviv, early 20c.); Br\"uckner 1918/1985} \\
 &  & Indian & Rakshasa & \texttt{Ramayana (Ravana, Kumbhakarna, Ravana's army); Mahabharata (Hidimba)} \\
 &  & Egyptian & Apophis (Apep) & \texttt{*Book of Apophis*; Pyramid Texts; Coffin Texts} \\
 &  & Chinese & Niu Mowang (Bull Demon King) & \texttt{*Xiyou Ji* (Journey to the West)} \\
 &  & Japanese & Oni & \texttt{*Konjaku Monogatari*; *Otogiz\=oshi*; folk tradition} \\
 &  & Mesopotamian & Humbaba (Huwawa) & \texttt{*Epic of Gilgamesh*} \\
\addlinespace[2pt]
\textsc{G303} & Devil/demon figure & Greek & Typhon & \texttt{Hard 2004} \\
 &  & Roman & Orcus & \texttt{Grimal 1996} \\
 &  & Norse & Loki & \texttt{Lindow 2001} \\
 &  & Finnish & Piru & \texttt{Finnish belief} \\
 &  & Ukrainian & Chort & \texttt{Voropay 1958 *Zvychayi nashoho narodu*; Br\"uckner 1918/1985} \\
 &  & Indian & Mahishasura & \texttt{Devi Mahatmya 2-3 (Markandeya Purana 81-93)} \\
 &  & Egyptian & Set (Seth) & \texttt{Pyramid Texts; Book of the Dead; Plutarch *De Iside et Osiride*} \\
 &  & Chinese & Yaoguai & \texttt{*Shanhaijing*; *Soushen Ji* (Gan Bao, 4 c. CE); *Xiyou Ji*} \\
 &  & Japanese & Akuma & \texttt{Buddhist transmission texts; *Konjaku Monogatari*; folk tradition} \\
 &  & Mesopotamian & Pazuzu & \texttt{Akkadian incantations; bronze-amulet figures (Neo-Assyrian and later)} \\
\addlinespace[2pt]
\textsc{T110} & Goddess of love/fertility & Greek & Aphrodite & \texttt{Hard 2004} \\
 &  & Roman & Venus & \texttt{Grimal 1996} \\
 &  & Norse & Freyja & \texttt{Lindow 2001} \\
 &  & Finnish & Rauni\textsuperscript{3} & \texttt{Mikael Agricola 1551 (Tavastian gods list, preface to Psalter); Haavio 1959 *Karjalan jumalat*} \\
 &  & Ukrainian & Mokosh\textsuperscript{2} & \texttt{Gieysztor 2006 [pages pending]} \\
 &  & Indian & Lakshmi & \texttt{Sri Sukta (Rig Veda khila); Atharvaveda; Vishnu Purana 1.8-9} \\
 &  & Egyptian & Hathor & \texttt{Pyramid Texts; Coffin Texts; Dendera temple inscriptions} \\
 &  & Chinese & Yue Lao (Old Man under the Moon) & \texttt{*Sui Tang Jiahua* (Tang anecdotes); *Yu Li Bao Chao*} \\
 &  & Japanese & Konohanasakuya-hime & \texttt{*Kojiki*; *Nihon Shoki*} \\
 &  & Mesopotamian & Inanna (Ishtar) & \texttt{Sumerian *Inanna* hymns; Akkadian *Ishtar* hymns; *Epic of Gilgamesh* tablet VI} \\
\addlinespace[2pt]
\textsc{V1} & Sacred trees/groves & Greek & Dodona Oak & \texttt{Hard 2004} \\
 &  & Roman & Lucus & \texttt{Roman religion} \\
 &  & Norse & Yggdrasil & \texttt{Lindow 2001} \\
 &  & Finnish & Maailmanpuu & \texttt{Pentik\"ainen 1999 (Finnish *maailmanpuu* / cosmological world-pillar)} \\
 &  & Ukrainian & Perunov dub & \texttt{Constantine Porphyrogenitus *De Administrando Imperio* 9.70-80 (Moravcsik \& Jenkins 1967); Gieysztor 2006 (Perun's sacred oak at Khortytsia)} \\
 &  & Indian & A\'svattha & \texttt{Bhagavad Gita 15.1 (urdhva-m\=ula a\'svattha --- cosmic tree); Rig Veda 1.135.8} \\
 &  & Egyptian & Ished tree (Sycamore of Hathor) & \texttt{Pyramid Texts; Book of the Dead spell 109; royal annal inscriptions} \\
 &  & Chinese & Fusang & \texttt{*Shanhaijing* (Hai Wai Dong Jing); *Huainanzi* Tianwen Xun} \\
 &  & Japanese & Sakaki & \texttt{*Engishiki*; *Kojiki* (Ame-no-Iwato cave episode); Shinto ritual continuous to present} \\
 &  & Mesopotamian & Huluppu tree & \texttt{Sumerian *Bilgames, Enkidu and the Netherworld* (Huluppu Tree prologue)} \\
\addlinespace[2pt]
\textsc{Z111} & Death personified & Greek & Thanatos & \texttt{Hard 2004} \\
 &  & Roman & Mors & \texttt{Grimal 1996} \\
 &  & Norse & Hel & \texttt{Lindow 2001} \\
 &  & Finnish & Kalma & \texttt{Pentik\"ainen 1999} \\
 &  & Ukrainian & Mara & \texttt{Br\"uckner 1918/1985 *Mitologia s\l{}owia\'nska*; Strzelczyk 1998 *Mity, podania i wierzenia dawnych S\l{}owian*} \\
 &  & Indian & Yama & \texttt{Rig Veda 10.14; Katha Upanishad 1 (Yama as death personified, instructs Nachiketa)} \\
 &  & Egyptian & Anubis & \texttt{Pyramid Texts; Book of the Dead (esp. weighing-of-the-heart)} \\
 &  & Chinese & Yanluo Wang & \texttt{Daoist underworld bureaucracy texts; *Yu Li Bao Chao*} \\
 &  & Japanese & Shinigami & \texttt{Edo period folk tradition; modern *kaidan* literature} \\
 &  & Mesopotamian & Mamitu (Namtar) & \texttt{*Atrahasis*; *Epic of Gilgamesh*; Akkadian death-personification incantations} \\
\end{longtable}\endgroup

\paragraph{Footnotes on contested or structural-absence attributions.}
Thirteen of the 270 cells carry a superscript marker in Table~\ref{tab:dataset-full}. The five cases with specific scholarly contention concern the choice of figure within a tradition; the eight structural-absence cases mark cells where the motif role has no canonical filler in the source tradition and are preserved as flagged rows rather than forced near-miss substitutions, in line with the cross-cultural NLP program articulated by~\citet{hamalainen2024humanities}.

\begin{enumerate}[noitemsep,leftmargin=*]
\item[1.] \textbf{Khors} (A220, sun-god). Solar identification is contested: \citet{luczynski2020bogowie} argues for lunar reading on Iranian-etymological grounds (\textit{xur\v{s}\=\i d}); \citet{gieysztor2006mitologia} retains the canonical solar-disk reading from the Hypatian Codex \textit{s.a.}~6622 (1114) interpolation of John Malalas. We follow Gieysztor and the Hypatian gloss because it is the earliest and only continuous medieval attestation of the entity in a solar context. We separately retain Da\v{z}bog as the sun-as-deity filler of A710 (creation/origin of the sun), which is the same Hypatian passage's other named figure.
\item[2.] \textbf{Mokoš} (T110, goddess of love/fertility), replaced our v1 entity Lada. Modern Slavicist consensus holds Lada to be a refrain word in folk wedding songs misinterpreted as a theonym from the 15th century onward; \citet{luczynski2020bogowie} traces the historiography of the misattribution. Mokoš is the only female deity in the canonical Vladimirian pantheon (Primary Chronicle 980) and is attested in the East Slavic continuum across the medieval--modern transition; we use her as the fertility-and-female-domain filler.
\item[3.] \textbf{Rauni} (T110, goddess of love/fertility), replaced our v1 entity Mielikki. Mielikki is a forest-goddess in the Kalevala, not a fertility figure; we corrected to Rauni following Mikael Agricola's 1551 Tavastian list of pre-Christian Finnish deities, where Rauni is glossed as Ukko's wife. Identity contested in modern Finnish folkloristics: \citet{siikala2012finnish} treats Rauni as a possible epithet of Ukko himself (cf.\ \textit{rauni}, ``rowan tree''); we adopt the Agricola canonical reading, pending native-speaker validation by the second author.
\item[4.] \textbf{Eidolon} (E400, ghosts/revenants). The plural \textit{eidola} is more standard in modern Greek-philological practice (cf.\ \textit{Odyssey}~11), but the singular \textit{eidolon} is what an LLM is most likely to emit if asked for a name; we use the singular for tokenizer-friendliness and to bias the test toward modern English-classical-tradition usage. Either form scores correctly under our substring/Levenshtein metric.
\item[5.] \textbf{Mech-kladenets} (D1080, magic weapon) is a generic East Slavic folktale name for a self-wielding sword (lit.\ ``treasure-sword''), not a specific named weapon as Aegis (Greek), Ancile (Roman), or Gram (Norse) are. Specific named alternatives exist (Dobrynya's saber; Volga's sword) but are tale-type-bound rather than canonically cultural; we keep the generic East Slavic term as a more honest representation of how the motif is realized in the Slavic continuum.
\item[6.] \textbf{Structural absence}: Japanese mythology lacks a canonical flood-myth parallel to Deucalion (Greek) or Bergelmir (Norse); the Kojiki and Nihon Shoki narrate cosmic events (\textit{Ame-no-iwato}, the world-tree, the heavenly retreat) without a primordial deluge. We mark this cell as a documented cultural absence rather than force a near-miss attribution.
\item[7.] \textbf{Structural absence}: Slavic mythology has no Prometheus-parallel theft-of-fire narrative. Svarog is the smith-deity associated with the heavenly fire (Hypatian Codex \textit{s.a.}~6622) but does not steal fire on humanity's behalf. The absence is itself culturally meaningful and is preserved as a flagged row rather than substituted with a near-miss.
\item[8.] \textbf{Structural absence}: Egyptian tradition has no Prometheus-equivalent. Fire is divine domain (especially Ra's solar fire and the Eye of Ra) but is never thieved on humanity's behalf in the Egyptian literary corpus.
\item[9.] \textbf{Partial-fit}: Kagutsuchi is the fire-deity born from Izanami (who dies giving birth to him in Kojiki book~1); this is a fire-origin narrative but not a theft narrative. We mark the row partial-fit rather than full-fit and let the row be evaluated as a structural near-miss.
\item[10.] \textbf{Structural absence}: Mesopotamian tradition has no theft-of-fire narrative. Fire is divine (Gibil, Nuska as fire-gods) but is not stolen on humanity's behalf.
\item[11.] \textbf{Structural absence}: Egypt's geography (desert plus Nile floodplain) lacks dense forest, hence has no canonical forest-spirit class comparable to Greek Dryad or Norse Skogsr\aa{}. The absence is geographical, not literary.
\item[12.] \textbf{Partial-fit}: Kupala is a midsummer ritual complex (with a ritual herb/effigy of the same name), not a personified figure comparable to Orpheus (looking-back tabu), Lemmink\"ainen (Tuonela tabu), or Baldr. We retain Kupala as a flagged row to mark the cultural realization of the looking-tabu motif as ritual-collective rather than personal-narrative.
\item[13.] \textbf{Structural absence}: Finnish folk tradition lacks a canonical flood-survivor figure comparable to Deucalion or Atrahasis; the Kalevala does not narrate a primordial deluge. The flood event itself (\textit{vedenpaisumus}) is named but no personified hero is documented in the canonical corpus.
\end{enumerate}

%% file: tab_tier_a.tex
\begin{table}[t]
\centering\small
\setlength{\tabcolsep}{6pt}
\begin{tabular}{@{}lcc@{}}
\toprule
\textbf{Culture} & \textbf{Tier-A (14)} & \textbf{Tier-B/C (4)} \\
\midrule
Greek & +0.18 & +0.21 \\
Egyptian & +0.18 & +0.09 \\
Japanese & +0.06 & -0.01 \\
Roman & +0.02 & +0.06 \\
Ukrainian & +0.02 & +0.00 \\
Indian & -0.01 & -0.01 \\
Mesopotamian & -0.05 & +0.01 \\
Finnish & -0.04 & -0.06 \\
Norse & -0.08 & -0.02 \\
Chinese & -0.07 & -0.16 \\
\bottomrule
\end{tabular}
\caption{Language-proficiency control (\S\ref{sec:tier-a}): per-culture EN$-$NL majority-accuracy delta (positive = English query wins) recomputed on the 14 Tier-A models with documented multilingual pretraining, vs the Tier-B/C models (Yi-1.5 EN+ZH only; OLMo-3.1 English-primary). The Tier-A ordering matches the same contrast computed over all 18 models (Spearman $\rho = 0.96$), so the per-culture asymmetry is not an artefact of language non-capability in English-primary models. Values are computed on the unrescored majority-correct aggregates and so differ slightly from the rescored figures in Table~\ref{tab:percult-delta}.}
\label{tab:tier-a}
\end{table}

%% file: tab_probe_stability.tex
\begin{table*}[t]
\centering\small
\setlength{\tabcolsep}{4pt}
\begin{tabular}{@{}llccccccc@{}}
\toprule
\textbf{Family} & \textbf{Model} & \textbf{fold\,std} & \textbf{LC.25} & \textbf{LC.50} & \textbf{LC.75} & \textbf{LC1.0} & \textbf{$\Delta$base [95\% CI]} & \textbf{$p$} \\
\midrule
\multirow{3}{*}{Llama 3.x} & Llama-3.2-1B & 0.027 & 0.62 & 0.70 & 0.75 & 0.79 & +0.189 [+0.13, +0.25] & $<$.001 \\
 & Llama-3.2-3B & 0.017 & 0.67 & 0.76 & 0.82 & 0.83 & +0.230 [+0.17, +0.29] & $<$.001 \\
 & Llama-3.1-8B & 0.019 & 0.75 & 0.82 & 0.85 & 0.88 & +0.278 [+0.22, +0.33] & $<$.001 \\
\midrule
\multirow{4}{*}{Gemma 4} & Gemma-4-E2B & 0.041 & 0.37 & 0.49 & 0.55 & 0.61 & +0.011 [-0.04, +0.06] & 0.368 \\
 & Gemma-4-E4B & 0.032 & 0.41 & 0.50 & 0.58 & 0.65 & +0.048 [-0.00, +0.10] & 0.040 \\
 & Gemma-4-26B-A4B & 0.046 & 0.45 & 0.56 & 0.61 & 0.69 & +0.085 [+0.03, +0.14] & $<$.001 \\
 & Gemma-4-31B & 0.032 & 0.46 & 0.57 & 0.62 & 0.69 & +0.085 [+0.04, +0.13] & $<$.001 \\
\midrule
\multirow{2}{*}{Phi 4} & Phi-4-mini & 0.020 & 0.71 & 0.77 & 0.80 & 0.83 & +0.230 [+0.18, +0.29] & $<$.001 \\
 & Phi-4 & 0.007 & 0.74 & 0.81 & 0.85 & 0.86 & +0.252 [+0.20, +0.31] & $<$.001 \\
\midrule
\multirow{2}{*}{Qwen 1.x} & Qwen2.5-1.5B & 0.015 & 0.62 & 0.71 & 0.74 & 0.75 & +0.148 [+0.10, +0.20] & $<$.001 \\
 & Qwen2.5-7B & 0.019 & 0.67 & 0.74 & 0.78 & 0.80 & +0.200 [+0.14, +0.26] & $<$.001 \\
\midrule
\multirow{2}{*}{Qwen 3.6} & Qwen-3.6-35B-A3B & 0.012 & 0.65 & 0.78 & 0.81 & 0.83 & +0.230 [+0.17, +0.29] & $<$.001 \\
 & Qwen-3.6-27B & 0.019 & 0.74 & 0.82 & 0.87 & 0.88 & +0.278 [+0.22, +0.33] & $<$.001 \\
\midrule
\multirow{3}{*}{Yi 1.5} & Yi-1.5-6B & 0.051 & 0.59 & 0.66 & 0.71 & 0.77 & +0.163 [+0.11, +0.21] & $<$.001 \\
 & Yi-1.5-9B & 0.040 & 0.61 & 0.72 & 0.74 & 0.77 & +0.170 [+0.12, +0.22] & $<$.001 \\
 & Yi-1.5-34B & 0.040 & 0.63 & 0.75 & 0.79 & 0.83 & +0.222 [+0.17, +0.27] & $<$.001 \\
\midrule
\multirow{1}{*}{OLMo 3.1} & OLMo-3.1-32B & 0.025 & 0.73 & 0.81 & 0.85 & 0.86 & +0.259 [+0.21, +0.31] & $<$.001 \\
\midrule
\multirow{1}{*}{Tiny Aya} & Tiny-Aya-Global & 0.022 & 0.64 & 0.73 & 0.76 & 0.81 & +0.204 [+0.15, +0.26] & $<$.001 \\
\bottomrule
\end{tabular}
\caption{Probe stability and surface-baseline significance at each model's peak layer (\S\ref{sec:probe-stability}); probe accuracy itself is in Table~\ref{tab:headline}. \textbf{fold\,std}: std across the 5 folds; \textbf{LC.$x$}: accuracy with $x$ of the training folds subsampled (learning curve). \textbf{$\Delta$base}: margin over the char-$n$-gram surface baseline with a one-sided paired bootstrap 95\% CI (10k resamples over entities), computed against a baseline refit on the identical folds ($0.604$), so it differs slightly from the $0.596$-based margins of Table~\ref{tab:eval-probe-delta-app}; \textbf{$p$}: bootstrap $p$-value (exact McNemar agrees). Learning curves rise monotonically and plateau and per-fold std is small, so the probes learn a stable signal, not fitted noise; 17/18 models clear the baseline at $p<0.05$ (Gemma-4-E2B does not: $\Delta{=}{+}0.011$, $p{=}0.37$).}
\label{tab:probe-stability}
\end{table*}

%% file: tab_mcq_perculture.tex
\begin{table}[t]
\centering\small
\begin{tabular}{@{}lc@{}}
\toprule
\textbf{Culture} & \textbf{MCQ (mean, 18 models)} \\
\midrule
Japanese & 0.78 \\
Indian & 0.74 \\
Greek & 0.73 \\
Egyptian & 0.68 \\
Norse & 0.67 \\
Ukrainian & 0.67 \\
Mesopotamian & 0.66 \\
Chinese & 0.61 \\
Roman & 0.59 \\
Finnish & 0.58 \\
\bottomrule
\end{tabular}
\caption{Per-culture MCQ selection accuracy (plurality vote), averaged over the 18 models. The Greco-Roman advantage of the free-generation setting persists under selection (Roman $0.60$, Finnish $0.58$ at the bottom vs Japanese $0.78$, Greek $0.74$): on Roman cells $49\%$ of errors land on the same-motif Greek counterpart (vs $\sim$$11\%$ under uniform error), so the collapse onto the dominant tradition is visible inside a selection task, with no generation involved.}
\label{tab:mcq-perculture}
\end{table}

%% file: app_h6_prompts.tex

\paragraph{Per-language NL templates.} The following ten panels reproduce the
NL prompt set in the native script of each target language. Greek, Hindi,
Arabic, Chinese, Japanese, and Ukrainian (Cyrillic) are rendered as figures
because the pdflatex pipeline used for this paper does not natively support
those scripts; the source-of-truth strings live in
\texttt{pipeline/prompts\_v3.py}. The Roman, Norwegian Bokm\aa{}l, Finnish,
and Mesopotamian (English-fallback) panels are rendered identically for
visual consistency. The structural schema (V1--V5) is the same as the EN
schema above; only the surface phrasing changes.

\begin{figure*}[!t]
\centering
\includegraphics[width=0.99\linewidth]{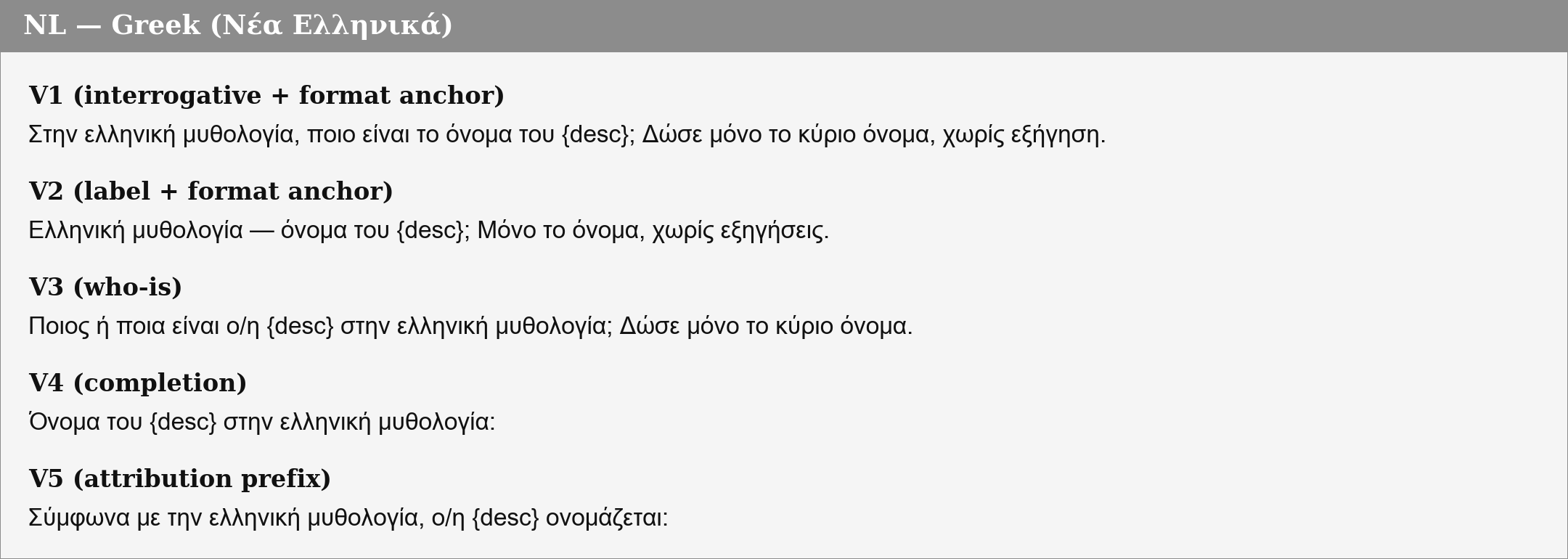}
\end{figure*}

\begin{figure*}[!t]
\centering
\includegraphics[width=0.99\linewidth]{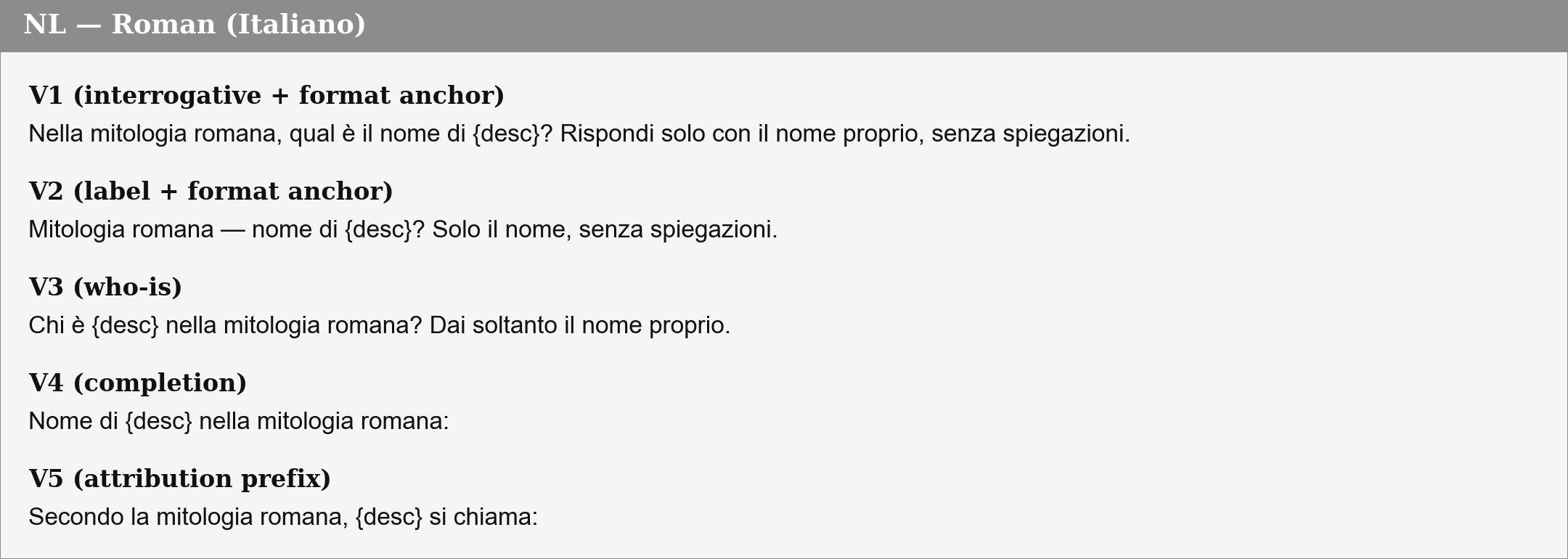}
\end{figure*}

\begin{figure*}[!t]
\centering
\includegraphics[width=0.99\linewidth]{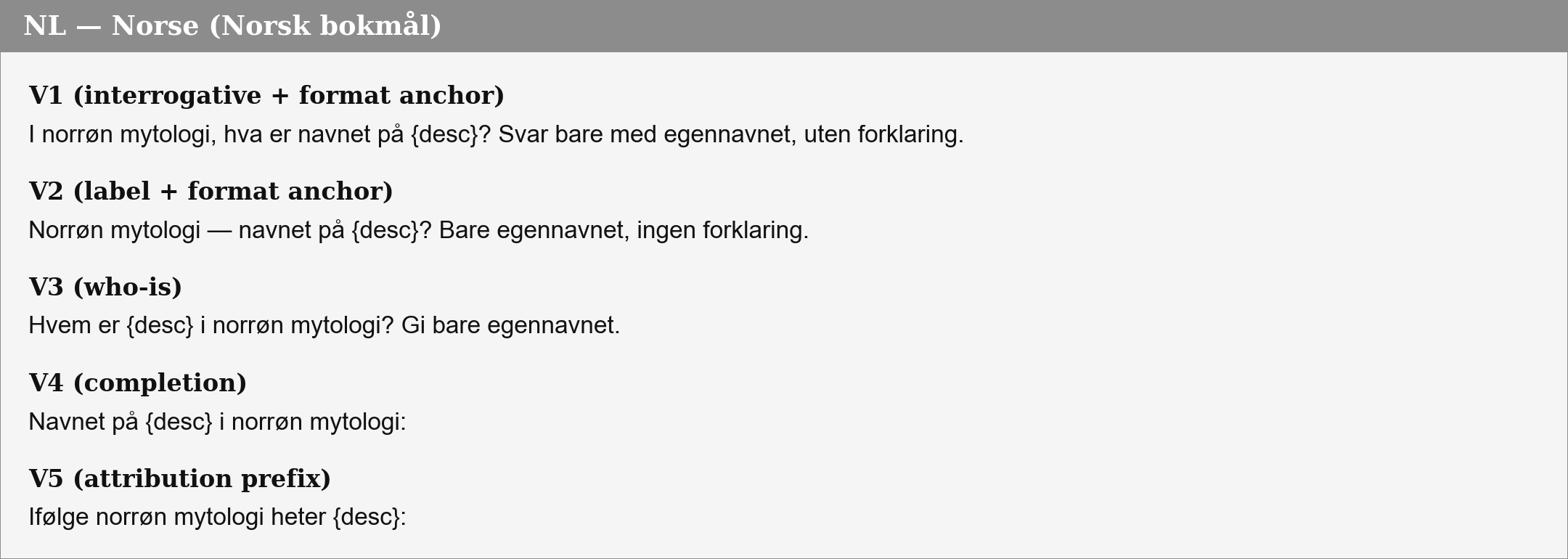}
\end{figure*}

\begin{figure*}[!t]
\centering
\includegraphics[width=0.99\linewidth]{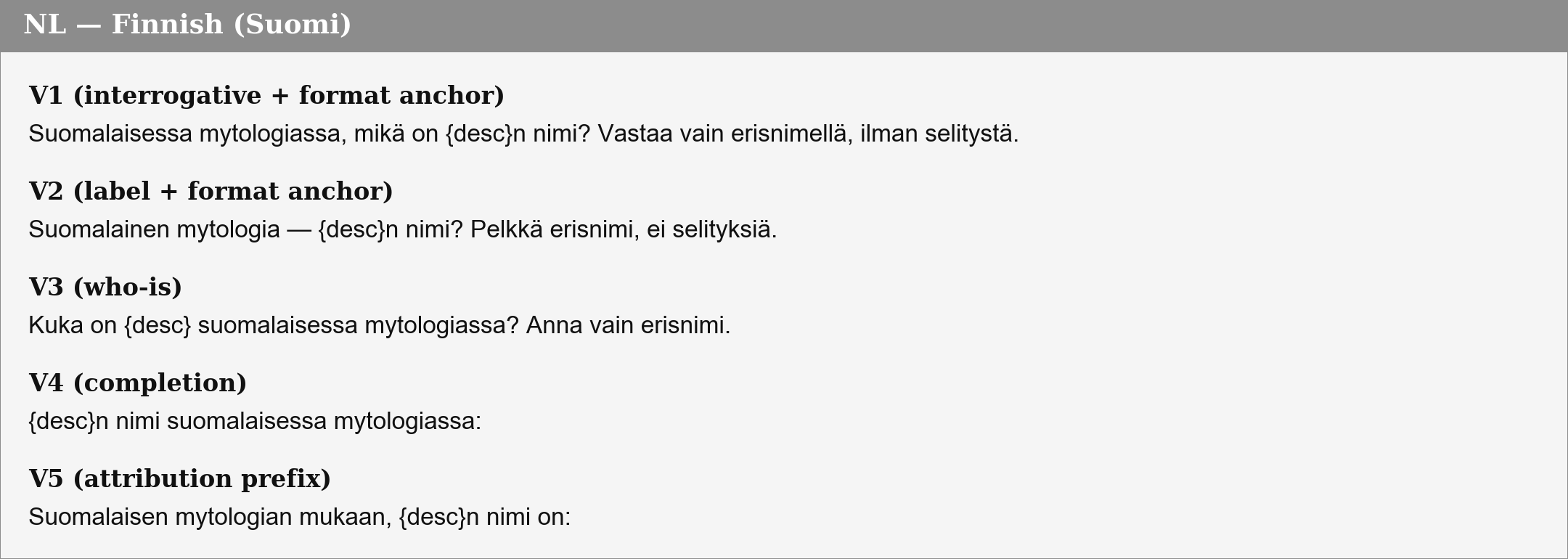}
\end{figure*}

\begin{figure*}[!t]
\centering
\includegraphics[width=0.99\linewidth]{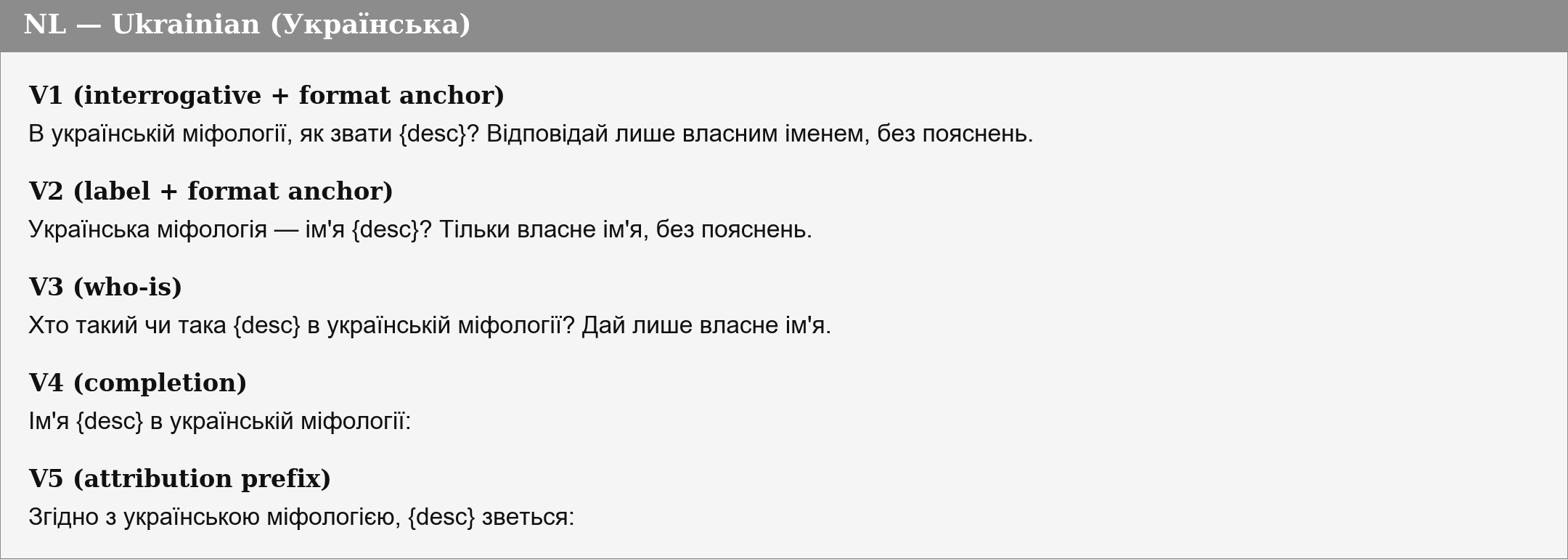}
\end{figure*}

\begin{figure*}[!t]
\centering
\includegraphics[width=0.99\linewidth]{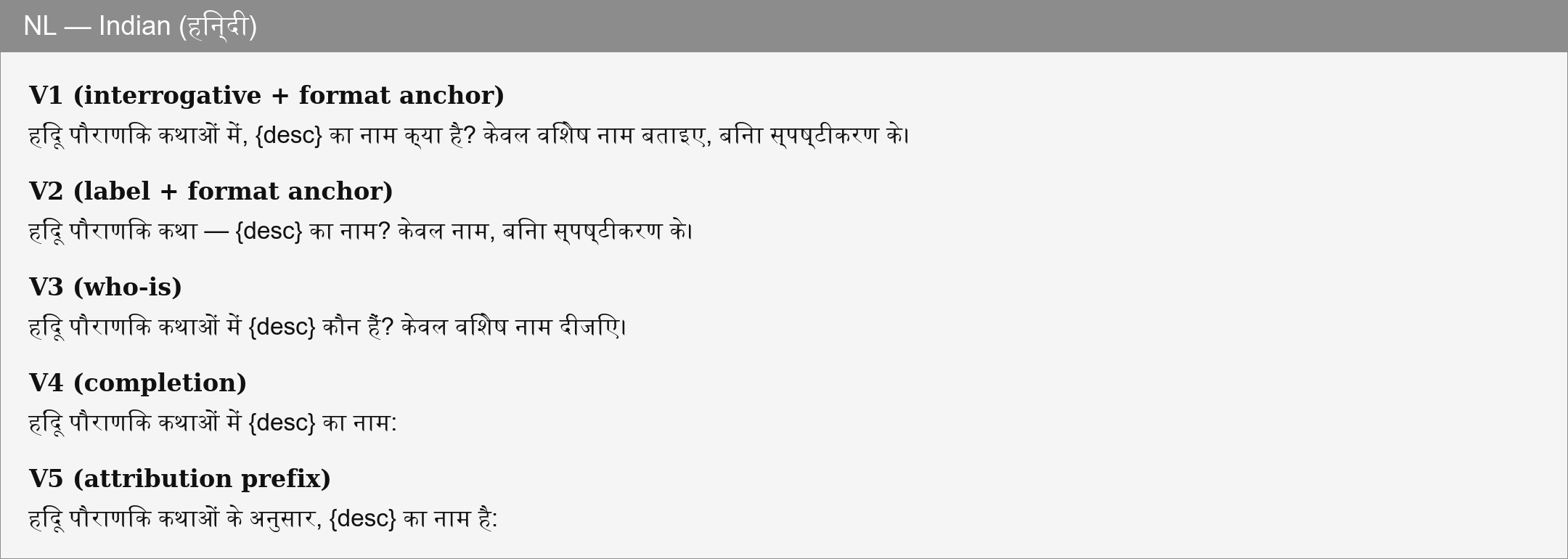}
\end{figure*}

\begin{figure*}[!t]
\centering
\includegraphics[width=0.99\linewidth]{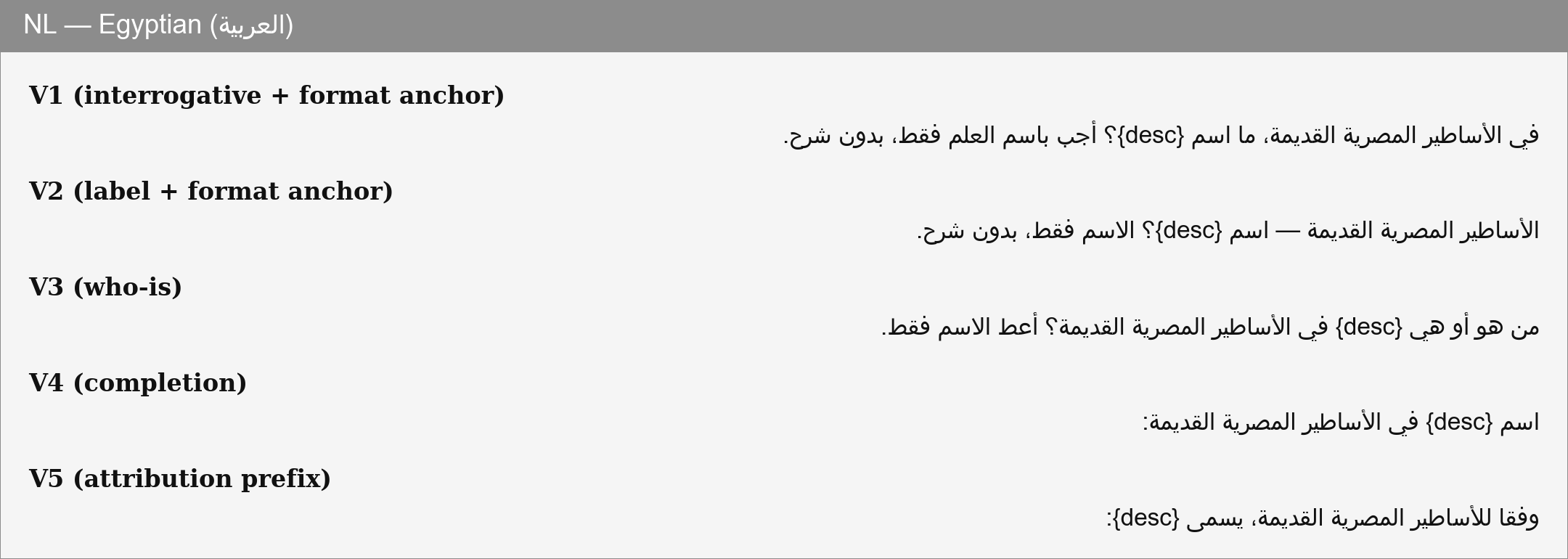}
\end{figure*}

\begin{figure*}[!t]
\centering
\includegraphics[width=0.99\linewidth]{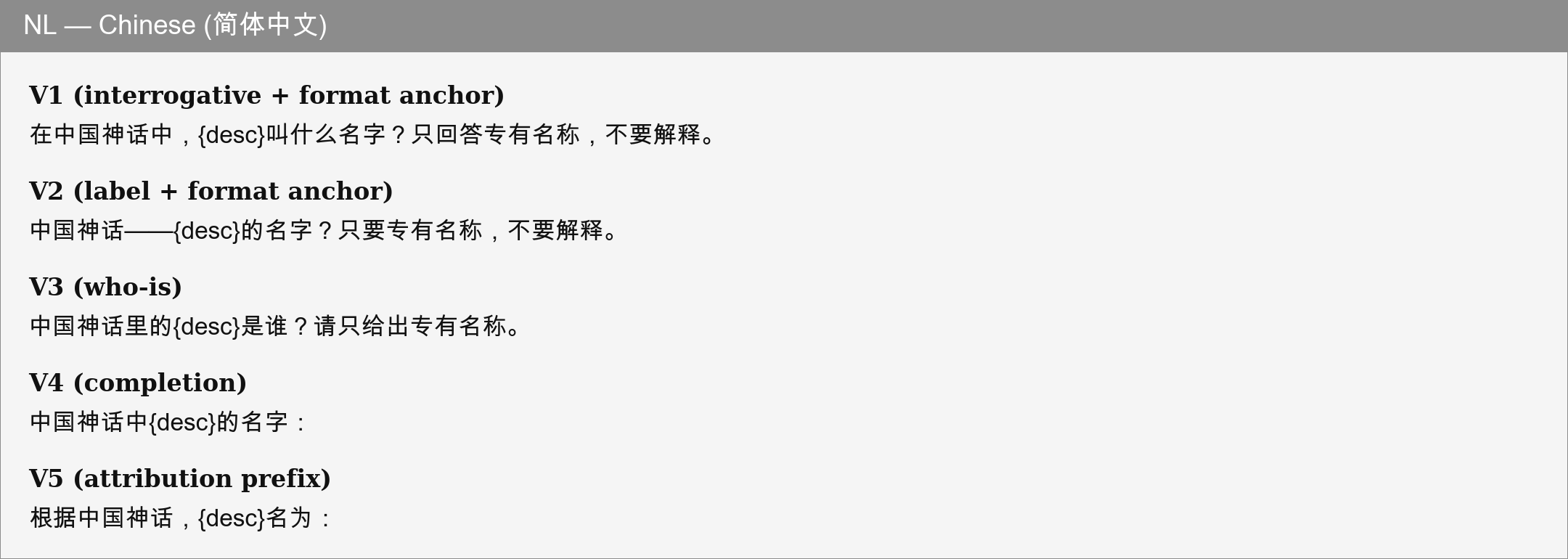}
\end{figure*}

\begin{figure*}[!t]
\centering
\includegraphics[width=0.99\linewidth]{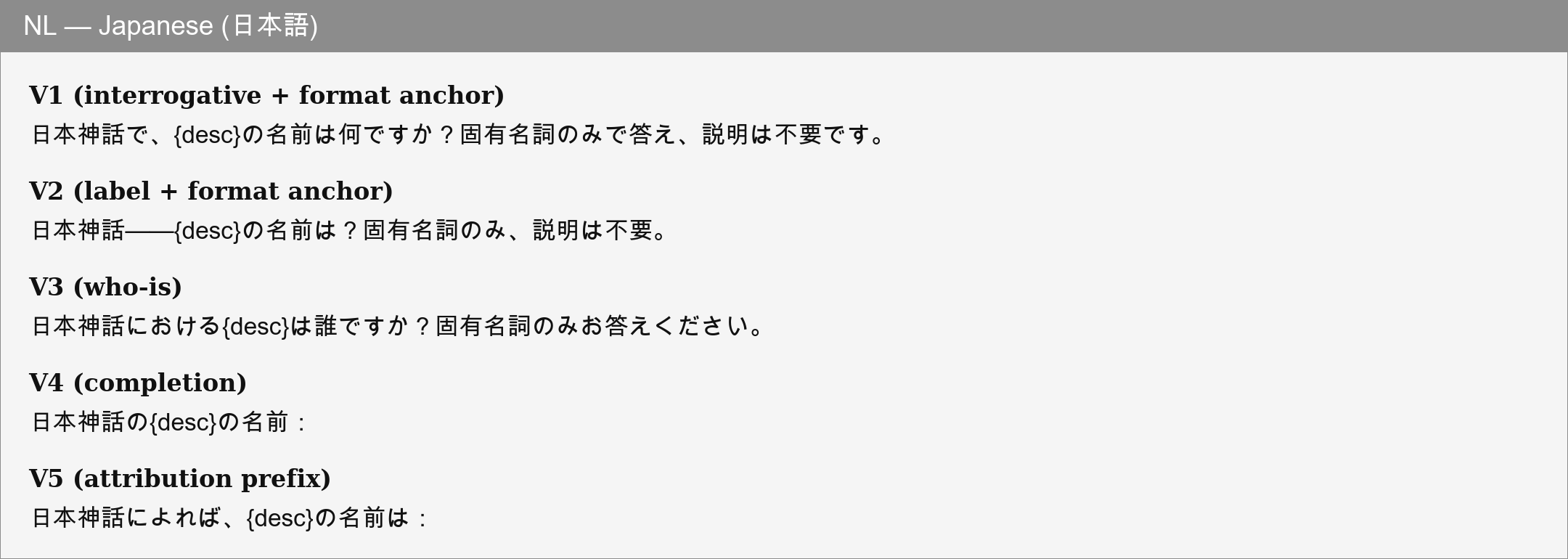}
\end{figure*}

\begin{figure*}[!t]
\centering
\includegraphics[width=0.99\linewidth]{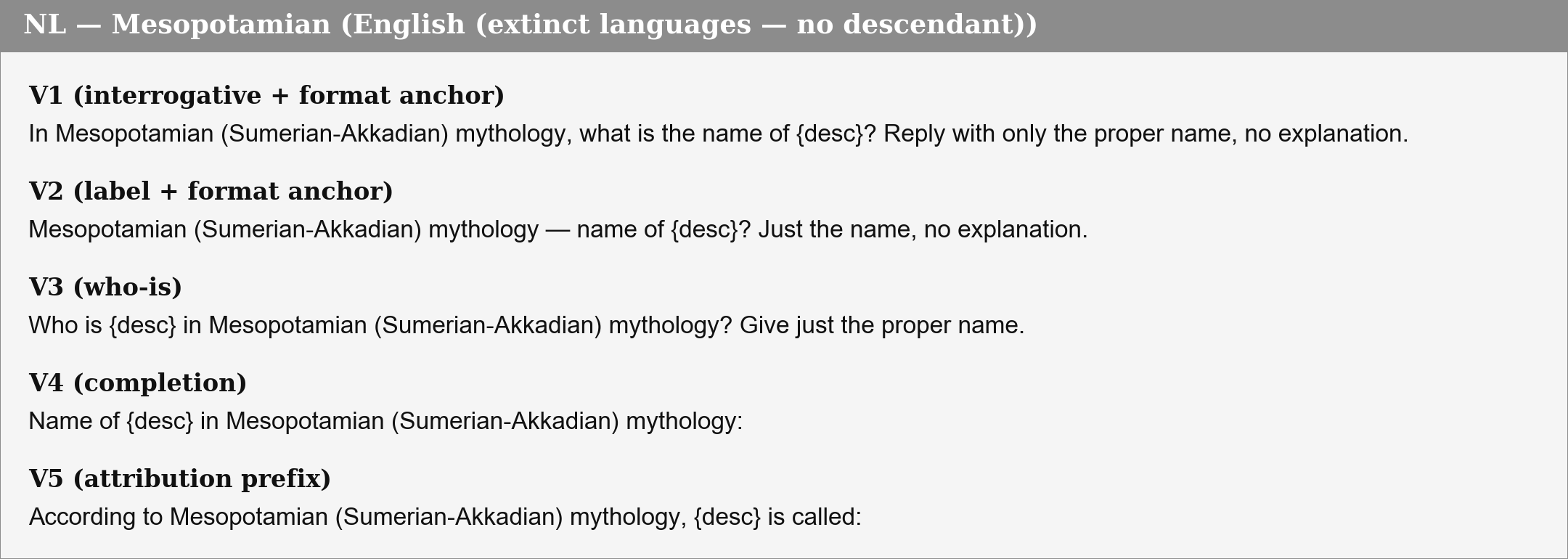}
\end{figure*}

\clearpage

%% file: tab_v6_family_multi.tex
\begin{table*}[t]
\centering
\small
\setlength{\tabcolsep}{4pt}
\renewcommand{\arraystretch}{1.18}
\begin{tabular}{@{}p{2.4cm} p{3.6cm} p{4.6cm} p{3.4cm} p{1.4cm}@{}}
\toprule
\textbf{Family} (models in our sweep) & \textbf{Tokenizer} (type, vocab) & \textbf{Pretraining language design} (from the technical report) & \textbf{Independent proxy evidence} & \textbf{Tier} \\
\midrule
Llama 3.x \newline (1.2B, 3.2B, 8B)               & tiktoken BPE, 128K (+28K non-EN tokens) \citep{dubey2024llama3}
                                                    & 8 supported langs: EN, DE, FR, IT, PT, HI, ES, TH; \textbf{8\% multilingual} of $\sim$15T tokens (the only family that publishes the share) \citep{dubey2024llama3}
                                                    & MMLU per-lang from the official card: IT $61.6$, HI $50.9$; BELEBELE coverage at 70B-scale clears chance for all 10 \citep{bandarkar2024belebele}
                                                    & \textbf{A} \\
Gemma 4 \newline (E2B, E4B, 26B-A4B, 31B)         & SentencePiece, 262K (Gemini-2.0 tokenizer) \citep{gemma3technical}
                                                    & ``140+ languages,'' no explicit list, no \% reported; report says ``more balanced for non-English'' \citep{gemma3technical}
                                                    & Global-MMLU-Lite (aggregate over 8 of our 10 langs): 27B $75.1$, 12B $69.5$, 4B $54.5$, 1B $34.2$ \citep{gemma3technical}; MMLU-ProX 27B $66.5$ on EN/IT/UK/HI/AR/ZH/JA \citep{xuan2025mmluprox}
                                                    & \textbf{A} \\
Phi 4 \newline (mini)                             & tiktoken BPE, 200K (Phi-4-mini, expanded for multi) \citep{abdin2024phi4}
                                                    & 22 supported langs including EN + 8 of our 10 (no EL, FI, NO, UK enumerated; Hindi, Arabic, Mandarin, Japanese present at the speech-modality level) \citep{abdin2024phi4}
                                                    & Multilingual-MMLU $49.3$, MGSM $63.9$ (aggregate, no per-lang) \citep{abdin2024phi4}; MMLU-ProX 14B per-lang \citep{xuan2025mmluprox}
                                                    & \textbf{A} \\
Phi 4 \newline (14B)                              & tiktoken BPE, 100K \citep{abdin2024phi4}
                                                    & \textbf{``Trained primarily on English''} (HuggingFace card); incidental multilingual data (DE, ES, FR, PT, IT, HI, JA) but no support claim \citep{abdin2024phi4}
                                                    & MMLU-ProX 14B: EN $71.5$, IT $60.2$, UK $61.3$, HI $47.8$, AR $56.8$, ZH $62.3$, JA $56.5$ \citep{xuan2025mmluprox} (above chance on all 7)
                                                    & \textbf{A}$^\dagger$ \\
Qwen 1.x \newline (1.5B, 7B; Qwen 2.5)            & byte-level BPE (BBPE), 151K \citep{yang2024qwen25}
                                                    & 29 supported langs (incl.\ all 10 of ours except NO and FI), 18T pretraining tokens; \% not reported \citep{yang2024qwen25}
                                                    & Multi-Understanding aggregate (BELEBELE, XCOPA, XWinograd, XStoryCloze, PAWS-X): 7B $79.3$, 1.5B $65.1$ \citep{yang2024qwen25}; okapi-MMLU 7B $66.98$
                                                    & \textbf{A} \\
Qwen 3.6 \newline (27B, 35B-A3B; Qwen 3)          & BBPE, 151K \citep{yang2025qwen3}
                                                    & \textbf{119 languages and dialects}, 36T tokens; explicit per-language list not enumerated in the report \citep{yang2025qwen3}
                                                    & MMMLU $81.5$--$83.8$, MGSM $79.1$--$83.1$, INCLUDE-44 $67.0$--$67.9$ for 30B/32B \citep{yang2025qwen3}
                                                    & \textbf{A} \\
Yi 1.5 \newline (6B, 9B, 34B)                     & BPE/SentencePiece, 64K \citep{young2024yi}
                                                    & \textbf{Bilingual EN + ZH only}; the report frames Yi as English\,+\,Chinese throughout, no other languages \citep{young2024yi}
                                                    & C-Eval, CMMLU, Gaokao-Bench (ZH); MMLU, BBH (EN). No XNLI / XCOPA / MGSM run \citep{young2024yi}
                                                    & \textbf{B} \\
OLMo 3.1 \newline (32B)                           & cl100k (tiktoken family), $\sim$100K \citep{olmo2025}
                                                    & \textbf{``OLMo 2 is not trained for multilingual tasks''} (explicit, OLMo 2 report); pretraining is Dolma/DCLM/Dolmino, English-primary by design \citep{olmo2025}
                                                    & No multilingual benchmark reported by the authors; English MMLU + OLMES only \citep{olmo2025}
                                                    & \textbf{C} \\
Tiny Aya \newline (Global, 3B; Aya 23 family)     & BPE, 256K \citep{aryabumi2024aya101}
                                                    & \textbf{23 supported languages enumerated}, incl.\ \emph{all 8} of our non-fallback non-CJK targets (EN, EL, IT, UK, HI, AR, ZH, JA), explicitly symmetric coverage \citep{ustun2024aya23, aryabumi2024aya101}
                                                    & XWinograd, XCOPA, XStoryCloze, M-MMLU, FLORES-200, MGSM run by the authors over all 23 langs \citep{aryabumi2024aya101}
                                                    & \textbf{A}$^\ddagger$ \\
\bottomrule
\end{tabular}
\caption{Per-family tokenizer + pretraining language design + independent multilingual benchmark evidence, and our \textbf{tier} classification: \textbf{A} = multilingual-by-design; \textbf{B} = explicitly bilingual EN+ZH only; \textbf{C} = English-primary, multilinguality explicitly disclaimed by the authors. ${}^\dagger$Phi-4 14B is the borderline case: classified A because the same family's mini variant is overtly multilingual and MMLU-ProX shows Phi-4 14B above chance on every covered language, but the card declares ``primarily English.'' ${}^\ddagger$Aya-Tiny is the family's \emph{positive control}: explicitly symmetric coverage of all 8 of our non-CJK non-fallback target languages. Tokenizer and pretraining design are distinct levels: even an English-primary model (OLMo) can have a multilingual-vocabulary tokenizer (cl100k), and conversely a multilingual tokenizer does not guarantee per-language pretraining coverage.}
\label{tab:v6-family-multi}
\end{table*}

%% file: tab_eval_perm.tex
\begin{table*}[t]
\centering
\small
\setlength{\tabcolsep}{4pt}
\begin{tabular}{lccc}
\toprule
\textbf{Model} & \textbf{Observed gap} & \textbf{Perm.\ null mean $\pm$ std} & \textbf{$p$ (one-sided)} \\
\midrule
Gemma-4-26B-A4B    & 0.254 & $-0.001 \pm 0.052$ & $0.0010$ \\
Tiny-Aya-Global    & 0.191 & $-0.001 \pm 0.045$ & $0.0010$ \\
Llama-3.2-3B       & 0.306 & $-0.000 \pm 0.054$ & $0.0010$ \\
Qwen-7B            & 0.225 & $+0.001 \pm 0.052$ & $0.0010$ \\
Phi-4              & 0.276 & $-0.002 \pm 0.047$ & $0.0010$ \\
Qwen-3.6-35B-A3B   & 0.217 & $-0.001 \pm 0.049$ & $0.0020$ \\
Gemma-4-31B        & 0.235 & $-0.000 \pm 0.054$ & $0.0025$ \\
Gemma-4-E4B        & 0.226 & $+0.002 \pm 0.051$ & $0.0025$ \\
OLMo-3.1-32B       & 0.208 & $-0.000 \pm 0.044$ & $0.0040$ \\
Gemma-4-E2B        & 0.189 & $-0.000 \pm 0.045$ & $0.0040$ \\
Yi-1.5-9B          & 0.359 & $+0.001 \pm 0.062$ & $0.0045$ \\
Qwen-1.5B          & 0.438 & $+0.002 \pm 0.081$ & $0.0055$ \\
Yi-1.5-6B          & 0.337 & $+0.000 \pm 0.058$ & $0.0060$ \\
Llama-3.2-1B       & 0.388 & $+0.001 \pm 0.070$ & $0.0070$ \\
Yi-1.5-34B         & 0.331 & $-0.000 \pm 0.057$ & $0.0070$ \\
Qwen-3.6-27B       & 0.218 & $-0.001 \pm 0.053$ & $0.0075$ \\
Llama-3.1-8B       & 0.277 & $+0.000 \pm 0.053$ & $0.0075$ \\
Phi-4-mini         & 0.274 & $+0.002 \pm 0.055$ & $0.0090$ \\
\bottomrule
\end{tabular}
\caption{Permutation test for the within-vs-cross paraphrase correlation gap (App.~\ref{app:eval-perm-detail}). For each model we shuffle the 10 paraphrase column labels (5 NL + 5 EN) and recompute the within$-$cross correlation gap 5{,}000 times. The observed gap exceeds the permutation null at $p \leq 0.01$ for every chat-template model in the bilingual sweep.}
\label{tab:eval-perm}
\end{table*}

%% file: tab_independence.tex
\begin{table*}[t]
\centering
\footnotesize
\setlength{\tabcolsep}{4pt}
\renewcommand{\arraystretch}{0.95}
\begin{tabular}{@{}llrcccccc@{}}
\toprule
\textbf{Family} & \textbf{Model} & \textbf{Par.} & $\overline{r}_{\text{within}}$ & $\overline{r}_{\text{cross}}$ & \textbf{Gap} & \textbf{Best-single} & $\textbf{NL}\cup\textbf{EN}$ & $+\Delta$ \\
\midrule
 \multirow{3}{*}{Llama 3.x} & Llama-3.2-1B & 1.2 & 0.56 & 0.17 & 0.39 & 0.09 & 0.16 & $+0.07$ \\
  & Llama-3.2-3B & 3.2 & 0.59 & 0.29 & 0.31 & 0.18 & 0.24 & $+0.06$ \\
  & Llama-3.1-8B & 8.0 & 0.58 & 0.30 & 0.28 & 0.25 & 0.30 & $+0.06$ \\
\midrule
 \multirow{4}{*}{Gemma 4} & Gemma-4-E2B & 2.0 & 0.56 & 0.37 & 0.19 & 0.27 & 0.34 & $+0.07$ \\
  & Gemma-4-E4B & 4.0 & 0.59 & 0.36 & 0.23 & 0.33 & 0.40 & $+0.07$ \\
  & Gemma-4-26B-A4B & 4.0 & 0.57 & 0.31 & 0.25 & 0.42 & 0.53 & $+0.10$ \\
  & Gemma-4-31B & 33 & 0.56 & 0.33 & 0.24 & 0.43 & 0.54 & $+0.11$ \\
\midrule
 \multirow{2}{*}{Phi 4} & Phi-4-mini & 3.8 & 0.54 & 0.26 & 0.27 & 0.20 & 0.27 & $+0.07$ \\
  & Phi-4 & 14 & 0.57 & 0.29 & 0.28 & 0.31 & 0.40 & $+0.10$ \\
\midrule
 \multirow{2}{*}{Qwen 1.x} & Qwen2.5-1.5B & 1.5 & 0.55 & 0.12 & 0.44 & 0.12 & 0.20 & $+0.09$ \\
  & Qwen2.5-7B & 7.0 & 0.52 & 0.30 & 0.23 & 0.24 & 0.32 & $+0.08$ \\
\midrule
 \multirow{2}{*}{Qwen 3.6} & Qwen-3.6-35B-A3B & 3.0 & 0.55 & 0.33 & 0.22 & 0.38 & 0.47 & $+0.10$ \\
  & Qwen-3.6-27B & 27 & 0.59 & 0.37 & 0.22 & 0.35 & 0.46 & $+0.11$ \\
\midrule
 \multirow{3}{*}{Yi 1.5} & Yi-1.5-6B & 6.0 & 0.54 & 0.20 & 0.34 & 0.13 & 0.19 & $+0.07$ \\
  & Yi-1.5-9B & 9.0 & 0.55 & 0.19 & 0.36 & 0.18 & 0.26 & $+0.09$ \\
  & Yi-1.5-34B & 34 & 0.62 & 0.29 & 0.33 & 0.26 & 0.35 & $+0.09$ \\
\midrule
 OLMo 3.1 & OLMo-3.1-32B & 32 & 0.55 & 0.35 & 0.21 & 0.32 & 0.37 & $+0.06$ \\
\midrule
 Tiny Aya & Tiny-Aya-Global & 3.0 & 0.58 & 0.39 & 0.19 & 0.17 & 0.24 & $+0.07$ \\
\midrule
 \textbf{Mean} & & & 0.57 & 0.29 & 0.27 & 0.26 & 0.34 & $+0.08$ \\
\bottomrule
\end{tabular}
\caption{Cross-lingual independence and bilingual-ensemble lift, per chat-template model with both NL and EN modes. $\overline{r}_{\text{within}}$: mean pairwise per-cell correctness correlation across the 5 paraphrases of one language, averaged over NL and EN; $\overline{r}_{\text{cross}}$: same across NL${\times}$EN pairs. \textbf{Gap} = within $-$ cross. \textbf{Best-single}: best of (NL, EN) majority-correct ($\geq 3/5$ paraphrases); $\textbf{NL}\cup\textbf{EN}$: bilingual disjunction over $5{+}5$ paraphrases; $+\Delta$ is the absolute lift over best-single. Cross-language queries are roughly $2{\times}$ more decoupled than within-language ones (mean gap $0.27$, $p\leq 0.01$ per model by permutation, App.~\ref{app:eval-perm-detail}); mean bilingual lift $+0.08$ absolute, $+36\%$ relative.}
\label{tab:independence}
\end{table*}

%% file: tab_eval_scoring.tex
\begin{table}[H]
\centering
\footnotesize
\setlength{\tabcolsep}{3pt}
\renewcommand{\arraystretch}{1.1}
\begin{tabular}{@{}lcccc@{}}
\toprule
\textbf{Metric} & \makecell{\textbf{Mean}\\\textbf{(all 36)}} & \makecell{\textbf{Mean}\\\textbf{(chat, 34)}} & \makecell{\textbf{Pearson}\\\textbf{vs paper$^\dagger$}} & \makecell{\textbf{Spearman}\\\textbf{vs paper$^\dagger$}} \\
\midrule
Substring                     & 0.180 & 0.178 & 1.00 & 1.00 \\
Levenshtein $\leq 0.2$        & 0.188 & 0.186 & 1.00 & 1.00 \\
\makecell[l]{Paper metric\\(substr.\ $\cup$ Lev.)} & \textbf{0.188} & \textbf{0.186} & 1.00 & 1.00 \\
First-word EM                 & 0.072 & 0.070 & 0.92 & 0.87 \\
\makecell[l]{Strict EM\\(normalized)}   & 0.061 & 0.059 & 0.87 & 0.80 \\
\bottomrule
\end{tabular}
\caption{Scoring strictness audit: per-row scoring of the 48\,568 generations (36 (model, mode) $\times$ 1\,350) under five metrics. The paper metric and substring/Levenshtein components rank models identically. Stricter EM-style metrics correlate well \emph{within chat-template models} (Spearman $\geq 0.80$) but lower across the full 36 cells. Headline rankings stable; absolute strict-EM levels $\sim$$0.13$ lower. $^\dagger$Chat-template subset (34 cells).}
\label{tab:eval-scoring}
\end{table}

\begin{table}[H]
\centering
\footnotesize
\setlength{\tabcolsep}{4pt}
\begin{tabular}{@{}lcc@{}}
\toprule
\textbf{Metric pair} & \textbf{Pearson} & \textbf{Spearman} \\
\midrule
paper $\leftrightarrow$ first-word EM   & 0.90 & 0.91 \\
paper $\leftrightarrow$ substring-only  & 0.97 & 0.96 \\
paper $\leftrightarrow$ strict EM       & 0.81 & 0.84 \\
\bottomrule
\end{tabular}
\caption{Per-model cross-metric correlation on chat-template models. Strict-EM rankings preserved at $r \geq 0.81$.}
\label{tab:eval-scoring-short}
\end{table}

%% file: tab_app_full_results.tex
\begin{table*}[!htbp]
\centering
\small
\setlength{\tabcolsep}{4pt}
\renewcommand{\arraystretch}{0.95}
\begin{tabular}{@{}llrrcccccccc@{}}
\toprule
\textbf{Family} & \textbf{Model} & \textbf{Par.} & \textbf{L} & \textbf{probe}$_0$ & \textbf{probe}$_\text{peak}$ & \textbf{peak L} & \textbf{null} & \textbf{NL} & \textbf{EN} & \textbf{best} \\
\midrule
 \multirow{3}{*}{Llama 3.x} & Llama-3.2-1B & 1.2 & 17 & 0.53 & 0.79 & 6 & 0.13 & 0.09 & 0.09 & 0.09 \\
  & Llama-3.2-3B & 3.2 & 29 & 0.56 & 0.83 & 8 & 0.15 & 0.14 & 0.18 & 0.18 \\
  & Llama-3.1-8B & 8.0 & 33 & 0.53 & 0.88 & 8 & 0.14 & 0.15 & 0.25 & 0.25 \\
\midrule
 \multirow{4}{*}{Gemma 4} & Gemma-4-E2B & 2.0 & 36 & 0.53 & 0.61 & 4 & 0.12 & 0.19 & 0.27 & 0.27 \\
  & Gemma-4-E4B & 4.0 & 43 & 0.54 & 0.65 & 5 & 0.14 & 0.22 & 0.33 & 0.33 \\
  & Gemma-4-26B-A4B & 4.0 & 31 & 0.53 & 0.69 & 3 & 0.14 & 0.33 & 0.42 & 0.42 \\
  & Gemma-4-31B & 33 & 61 & 0.51 & 0.69 & 9 & 0.13 & 0.36 & 0.43 & 0.43 \\
\midrule
 \multirow{2}{*}{Phi 4} & Phi-4-mini & 3.8 & 33 & 0.54 & 0.83 & 18 & 0.12 & 0.15 & 0.20 & 0.20 \\
  & Phi-4 & 14 & 41 & 0.55 & 0.86 & 13 & 0.13 & 0.24 & 0.31 & 0.31 \\
\midrule
 \multirow{2}{*}{Qwen 1.x} & Qwen2.5-1.5B & 1.5 & 29 & 0.53 & 0.75 & 25 & 0.14 & 0.11 & 0.12 & 0.12 \\
  & Qwen2.5-7B & 7.0 & 29 & 0.54 & 0.80 & 27 & 0.14 & 0.18 & 0.24 & 0.24 \\
\midrule
 \multirow{2}{*}{Qwen 3.6} & Qwen-3.6-35B-A3B & 3.0 & 41 & 0.52 & 0.83 & 7 & 0.13 & 0.30 & 0.38 & 0.38 \\
  & Qwen-3.6-27B & 27 & 65 & 0.54 & 0.88 & 11 & 0.15 & 0.31 & 0.35 & 0.35 \\
\midrule
 \multirow{3}{*}{Yi 1.5} & Yi-1.5-6B & 6.0 & 33 & 0.54 & 0.77 & 20 & 0.14 & 0.11 & 0.13 & 0.13 \\
  & Yi-1.5-9B & 9.0 & 49 & 0.53 & 0.77 & 28 & 0.14 & 0.13 & 0.18 & 0.18 \\
  & Yi-1.5-34B & 34 & 61 & 0.53 & 0.83 & 37 & 0.15 & 0.19 & 0.26 & 0.26 \\
\midrule
 OLMo 3.1 & OLMo-3.1-32B & 32 & 65 & 0.57 & 0.86 & 20 & 0.14 & 0.19 & 0.32 & 0.32 \\
\midrule
 Tiny Aya & Tiny-Aya-Global & 3.0 & 37 & 0.48 & 0.81 & 25 & 0.13 & 0.17 & 0.16 & 0.17 \\
\bottomrule
\end{tabular}
\caption{Full per-model results across the 18-model sweep on the 270-entity, 10-culture substrate. \textbf{L} = total residual-stream depth (incl.\ embedding); \textbf{probe}$_0$ = culture-probe accuracy at layer~0; \textbf{probe}$_\text{peak}$ = best across layers; \textbf{peak L} = layer index of the peak; \textbf{null} = best label-shuffled probe accuracy. \textbf{NL}, \textbf{EN}: per-mode majority-correct ($\geq 3/5$ paraphrases) output accuracy; \textbf{best}: best-of-(NL, EN). Per-(model, culture) breakdown is in Figure~\ref{fig:percult-app} (App.~\ref{app:percult}).}
\label{tab:app_full_results}
\end{table*}